\documentclass{article}
\pdfoutput=1

\usepackage{microtype}
\usepackage{graphicx}
\usepackage{subfigure}
\usepackage{booktabs} % for professional tables

\usepackage{url}            % simple URL typesetting
\usepackage{amsfonts}       % blackboard math symbols
\usepackage{nicefrac}       % compact symbols for 1/2, etc.
\usepackage{epsf}
\usepackage{epsfig}
\usepackage{amsmath}
\usepackage{amssymb}
\usepackage{amsthm}
\usepackage{mathtools}
\usepackage{multirow}
\usepackage{multicol}
\usepackage{xspace}
\usepackage{listings}
\usepackage{xcolor}
\usepackage{enumitem}

\usepackage{hyperref}

\usepackage[accepted]{icml2020}

\icmltitlerunning{Zeno++: Robust Fully Asynchronous SGD}

\newcounter{algline}
\newcommand{\algline}[1]{\refstepcounter{algline}\label{#1}}

\begin{document}

\def\Blue{\color{blue}}
\def\Purple{\color{purple}}

\def\A{{\bf A}}
\def\a{{\bf a}}
\def\B{{\bf B}}
\def\C{{\bf C}}
\def\c{{\bf c}}
\def\D{{\bf D}}
\def\d{{\bf d}}
\def\F{{\bf F}}
\def\e{{\bf e}}
\def\f{{\bf f}}
\def\G{{\bf G}}
\def\H{{\bf H}}
\def\I{{\bf I}}
\def\K{{\bf K}}
\def\M{{\bf M}}
\def\m{{\bf m}}
\def\N{{\bf N}}
\def\n{{\bf n}}
\def\Q{{\bf Q}}
\def\q{{\bf q}}
\def\S{{\bf S}}
\def\s{{\bf s}}
\def\T{{\bf T}}
\def\U{{\bf U}}
\def\u{{\bf u}}
\def\V{{\bf V}}
\def\v{{\bf v}}
\def\W{{\bf W}}
\def\w{{\bf w}}
\def\X{{\bf X}}
\def\x{{\bf x}}
\def\Y{{\bf Y}}
\def\y{{\bf y}}
\def\Z{{\bf Z}}
\def\z{{\bf z}}
\def\0{{\bf 0}}
\def\1{{\bf 1}}

\def\AM{{\mathcal A}}
\def\CM{{\mathcal C}}
\def\DM{{\mathcal D}}
\def\GM{{\mathcal G}}
\def\FM{{\mathcal F}}
\def\IM{{\mathcal I}}
\def\NM{{\mathcal N}}
\def\OM{{\mathcal O}}
\def\SM{{\mathcal S}}
\def\TM{{\mathcal T}}
\def\UM{{\mathcal U}}
\def\XM{{\mathcal X}}
\def\YM{{\mathcal Y}}
\def\RB{{\mathbb R}}

\def\TX{\tilde{\bf X}}
\def\tx{\tilde{\bf x}}
\def\ty{\tilde{\bf y}}
\def\TZ{\tilde{\bf Z}}
\def\tz{\tilde{\bf z}}
\def\hd{\hat{d}}
\def\HD{\hat{\bf D}}
\def\hx{\hat{\bf x}}
\def\TD{\tilde{\Delta}}

\def\alp{\mbox{\boldmath$\alpha$\unboldmath}}
\def\bet{\mbox{\boldmath$\beta$\unboldmath}}
\def\epsi{\mbox{\boldmath$\epsilon$\unboldmath}}
\def\etab{\mbox{\boldmath$\eta$\unboldmath}}
\def\ph{\mbox{\boldmath$\phi$\unboldmath}}
\def\pii{\mbox{\boldmath$\pi$\unboldmath}}
\def\Ph{\mbox{\boldmath$\Phi$\unboldmath}}
\def\Ps{\mbox{\boldmath$\Psi$\unboldmath}}
\def\tha{\mbox{\boldmath$\theta$\unboldmath}}
\def\Tha{\mbox{\boldmath$\Theta$\unboldmath}}
\def\muu{\mbox{\boldmath$\mu$\unboldmath}}
\def\Si{\mbox{\boldmath$\Sigma$\unboldmath}}
\def\si{\mbox{\boldmath$\sigma$\unboldmath}}
\def\Gam{\mbox{\boldmath$\Gamma$\unboldmath}}
\def\Lam{\mbox{\boldmath$\Lambda$\unboldmath}}
\def\De{\mbox{\boldmath$\Delta$\unboldmath}}
\def\Ome{\mbox{\boldmath$\Omega$\unboldmath}}
\def\TOme{\mbox{\boldmath$\hat{\Omega}$\unboldmath}}
\def\vps{\mbox{\boldmath$\varepsilon$\unboldmath}}
\newcommand{\ti}[1]{\tilde{#1}}
\def\Ncal{\mathcal{N}}
\def\argmax{\mathop{\rm argmax}}
\def\argmin{\mathop{\rm argmin}}
\providecommand{\abs}[1]{\lvert#1\rvert}
\providecommand{\norm}[2]{\lVert#1\rVert_{#2}}

\def\Zs{{\Z_{\mathrm{S}}}}
\def\Zl{{\Z_{\mathrm{L}}}}
\def\Yr{{\Y_{\mathrm{R}}}}
\def\Yg{{\Y_{\mathrm{G}}}}
\def\Yb{{\Y_{\mathrm{B}}}}
\def\Ar{{\A_{\mathrm{R}}}}
\def\Ag{{\A_{\mathrm{G}}}}
\def\Ab{{\A_{\mathrm{B}}}}
\def\As{{\A_{\mathrm{S}}}}
\def\Asr{{\A_{\mathrm{S}_{\mathrm{R}}}}}
\def\Asg{{\A_{\mathrm{S}_{\mathrm{G}}}}}
\def\Asb{{\A_{\mathrm{S}_{\mathrm{B}}}}}
\def\Or{{\Ome_{\mathrm{R}}}}
\def\Og{{\Ome_{\mathrm{G}}}}
\def\Ob{{\Ome_{\mathrm{B}}}}

\def\vec{\mathrm{vec}}
\def\fold{\mathrm{fold}}
\def\index{\mathrm{index}}
\def\sgn{\mathrm{sgn}}
\def\tr{\mathrm{tr}}
\def\rk{\mathrm{rank}}
\def\diag{\mathsf{diag}}
\def\const{\mathrm{Const}}
\def\dg{\mathsf{dg}}
\def\st{\mathsf{s.t.}}
\def\vect{\mathsf{vec}}
\def\MCAR{\mathrm{MCAR}}
\def\MSAR{\mathrm{MSAR}}
\def\etal{{\em et al.\/}\,}
\newcommand{\indep}{{\;\bot\!\!\!\!\!\!\bot\;}}

\def\Lsize{\hbox{\space \raise-2mm\hbox{$\textstyle \L \atop \scriptstyle {m\times 3n}$} \space}}
\def\Ssize{\hbox{\space \raise-2mm\hbox{$\textstyle \S \atop \scriptstyle {m\times 3n}$} \space}}
\def\Osize{\hbox{\space \raise-2mm\hbox{$\textstyle \Ome \atop \scriptstyle {m\times 3n}$} \space}}
\def\Tsize{\hbox{\space \raise-2mm\hbox{$\textstyle \T \atop \scriptstyle {3n\times n}$} \space}}
\def\Bsize{\hbox{\space \raise-2mm\hbox{$\textstyle \B \atop \scriptstyle {m\times n}$} \space}}

\newcommand{\twopartdef}[4]
{
	\left\{
		\begin{array}{ll}
			#1 & \mbox{if } #2 \\
			#3 & \mbox{if } #4
		\end{array}
	\right.
}

\newcommand{\tabincell}[2]{\begin{tabular}{@{}#1@{}}#2\end{tabular}}

\newtheorem{theorem}{Theorem}
\newtheorem{lemma}{Lemma}
\newtheorem{proposition}{Proposition}
\newtheorem{corollary}{Corollary}
\newtheorem{definition}{Definition}
\newtheorem{remark}{Remark}
\newtheorem{assumption}{Assumption}

\def\E{{\mathbb E}}
\def\R{{\mathbb R}}

\DeclarePairedDelimiter\ceil{\lceil}{\rceil}
\DeclarePairedDelimiter\floor{\lfloor}{\rfloor}

\newcommand{\ip}[2]{\left\langle #1, #2 \right \rangle}

\def\aggr{{\tt Aggr}}
\def\mean{{\tt Mean}}
\def\median{{\tt Median}}
\def\krum{{\tt Krum}}
\def\zeno{{\tt Zeno}}

\newcommand{\NewProcedure}[1]{\vspace{0.3cm}\STATE\hspace{-12pt} {\large\underline{\textbf{{#1}:}}}
\setcounter{ALC@line}{0}}
\newcommand{\NewThread}[1]{\vspace{0.2cm}\STATE\hspace{-\algorithmicindent} {\quad\large\underline{\textbf{{#1}:}}} 
\setcounter{ALC@line}{0}}

\setlength{\abovedisplayskip}{5pt}
\setlength{\belowdisplayskip}{0pt}
\setlength{\abovedisplayshortskip}{5pt}
\setlength{\belowdisplayshortskip}{0pt}

\twocolumn[
\icmltitle{Zeno++: Robust Fully Asynchronous SGD}




\begin{icmlauthorlist}
\icmlauthor{Cong Xie}{UIUC}
\icmlauthor{Oluwasanmi Koyejo}{UIUC}
\icmlauthor{Indranil Gupta}{UIUC}
\end{icmlauthorlist}

\icmlaffiliation{UIUC}{Department of Computer Science, University of Illinois, Urbana-Champaign, USA}

\icmlcorrespondingauthor{Cong Xie}{cx2@illinois.edu}

\icmlkeywords{SGD, robust}

\vskip 0.3in
]



\printAffiliationsAndNotice{}  % leave blank if no need to mention equal contribution

\begin{abstract}
We propose Zeno++, a new robust asynchronous Stochastic Gradient Descent~(SGD) procedure, intended to  tolerate Byzantine failures of  workers. In contrast to previous work, Zeno++ removes several unrealistic restrictions on worker-server communication, now allowing for fully asynchronous updates from anonymous workers, for arbitrarily stale worker updates, and for the possibility of an unbounded number of Byzantine workers. The key idea is to estimate the descent of the loss value after the candidate gradient is applied, where large descent values indicate that the update results in optimization progress. We prove the convergence of Zeno++ for non-convex problems under Byzantine failures. Experimental results show that Zeno++ outperforms existing Byzantine-tolerant asynchronous SGD algorithms.
\end{abstract}

\section{Introduction}

Synchronous training and asynchronous training are the two most common paradigms of distributed machine learning.
On the one hand, synchronous training requires 
the global updates at the server to be blocked until all the workers respond (after each period). In contrast, for asynchronous training, the server can update the global model immediately after each worker's response. 
Theoretical and experimental analysis~\citep{Dutta2018SlowAS} suggests that synchronous training is more stable and has lower noise, but is slower due to the global barrier across all the workers (after each period). Since asynchronous training is comparatively faster especially for  heterogeneous systems with stragglers, in this paper, we focus on asynchronous training. Doing so requires us to tackle challenges including instability and noisiness  that arise from asynchony. 

Concretely, we study the security of distributed asynchronous \underline{S}tochastic \underline{G}radient \underline{D}escent~(SGD) in a centralized worker-server architecture, also known as the \underline{P}arameter \underline{S}erver~(or PS) architecture. In the PS architecture, there are server nodes and worker nodes. 
Each worker pulls the global model from the servers, estimates the gradients using the local portion of the training data, then sends the gradient estimates to the servers. The servers update the model asynchronously, as soon as a new gradient is received from any worker.

The security of machine learning has gained increasing attention in recent years. In particular, tolerance to Byzantine failures~\citep{blanchard2017machine,Chen2017DistributedSM,yin2018byzantine,feng2014distributed,Su2016FaultTolerantMO,su2016defending,Xie2018ZenoBS,alistarh2018byzantine,Cao2018RobustDG,Xie2019FallOE} has become an important topic in the distributed machine learning literature. Byzantine nodes can behave arbitrarily, capturing behavior ranging from crash failures to malicious and arbitrary or compromised actions. 
In brief, the goal of Byzantine workers is to prevent convergence of the model training. By construction, Byzantine failures~\citep{lamport1982byzantine} assume the worst case, i.e., the Byzantine workers can behave arbitrarily. Such failures may be caused by a variety of reasons including but not limited to: hardware/software bugs, vulnerable communication channels, poisoned datasets, or malicious attackers. To make things worse, groups of Byzantine workers may collude, resulting in more harmful attacks. 

Byzantine failures have been well-studied for classical  distributed systems, especially for  consensus~\citep{lamport1982byzantine}. These results include well-known bounds on how many Byzantine workers can be tolerated. 
Yet, in distributed machine learning, Byzantine failures have a different impact than in classical consensus~\citep{damaskinos2018asynchronous}.
Unlike previous work, we consider Byzantine tolerance with minimal assumptions---an unbounded number of Byzantine workers, and  full asynchrony with unbounded delay.

Byzantine tolerance of asynchronous SGD is challenging due to several reasons:
\setitemize[0]{leftmargin=*,topsep=0pt}
\begin{itemize}
\item \textbf{Full asynchrony.} Lack of synchrony introduces  additional noise into the stochastic gradients. This makes it more difficult to distinguish gradients sent by Byzantine workers from gradients sent by  benign ones, especially as Byzantine behavior may exacerbate staleness. 
\item \textbf{Unpredictable successive updates.} The lack of synchronous scheduling means ``fast'' workers may send updates, to the server, more frequently than ``slower'' workers. Byzantine workers may exploit this by suffocating the server with their wrong gradients.
\item \textbf{Unbounded number of Byzantine workers.} While traditional Byzantine consensus assumes an upper bound on the number of malicious workers (typically one-third or half~\citep{lamport1982byzantine}), for fully asynchronous training, the assumption of a bounded number of Byzantine workers is impractical.  
In Byzantine-tolerant synchronous training~\citep{blanchard2017machine,Chen2017DistributedSM,yin2018byzantine,Xie2018ZenoBS,xie2019slsgd,xie2018phocas}, the servers can compare the candidate gradients with each other, and either utilize a majority  to filter out the harmful gradients, or use robust aggregation to bound the error. However, such strategies are infeasible in asynchronous training, since there may be nothing to compare against, or aggregate with. Aggregating the successive gradients is also meaningless since the successive gradients could all be pushed by the same Byzantine worker.
Furthermore, although most of the previous work~\citep{blanchard2017machine,Chen2017DistributedSM,yin2018byzantine,feng2014distributed,Su2016FaultTolerantMO,su2016defending,alistarh2018byzantine,Cao2018RobustDG} assumes a majority of honest workers, this requirement is not guaranteed to be satisfied in practice.
\end{itemize}

We propose Zeno++, a new algorithm that validates the gradients with low computation overhead on the server via lazy updates. 
Our key idea is to estimate the descent of the loss value after the candidate gradient is applied to the model parameters--for the latter we leverage the  Byzantine-tolerant synchronous SGD algorithm called  Zeno~\citep{Xie2018ZenoBS}. Intuitively, if the loss value decreases, the candidate gradient is likely to result in optimization progress. We also propose a mechanism of lazy updates to reduce the computation overhead.

To the best of our knowledge, this paper is the first to theoretically \emph{and} empirically study Byzantine-tolerant fully asynchronous SGD with anonymous workers, and potentially an unbounded number of Byzantine workers. In summary, our contributions are:
\setitemize[0]{leftmargin=*,topsep=0pt}
\begin{itemize} 
\item We propose Zeno++, a new approach for Byzantine-tolerant fully asynchronous SGD with anonymous workers, and low overhead of the additional computation for the validation on the servers.
\item We show that Zeno++ tolerates Byzantine workers without any limit on either the staleness or the number of Byzantine workers. 
\item We prove the convergence of Zeno++ for non-convex problems.
\item Experimental results validate that 1) existing algorithms may fail in practical scenarios, and 2) Zeno++ gracefully handles such cases.
\end{itemize}

\section{Related Work}
\label{sec:related_work}

Most of the existing Byzantine-tolerant SGD algorithms focus on synchronous training. \citet{Chen2017DistributedSM,Su2016FaultTolerantMO,su2016defending,yin2018byzantine,xie2018phocas} use robust statistics~\citep{huber2011robust} including the geometric median, coordinate-wise median, and trimmed mean as Byzantine-tolerant aggregation rules. \citet{blanchard2017machine,mhamdi2018hidden} propose Krum and its variants, which select the candidates with minimal local sum of Euclidean distances. \citet{alistarh2018byzantine} utilize historical information to identify harmful gradients. \citet{Chen2018DRACOBD} use coding theory and majority voting to recover correct gradients. Most of these synchronous algorithms assume that most of the workers are non-Byzantine. However, in practice, there are no guarantees that the number of Byzantine workers can be controlled.
\citet{Xie2018ZenoBS,Cao2018RobustDG} propose synchronous SGD algorithms for an unbounded number of Byzantine workers.

Recent years have witnessed an increasing number of large-scale machine learning algorithms, including asynchronous SGD~\citep{zinkevich2009slow,Lian2018AsynchronousDP,zheng2017asynchronous,Zhou2018DistributedAO}. 
\citet{damaskinos2018asynchronous} proposed Kardam, which to our knowledge is the only prior work to address Byzantine-tolerant asynchronous training. Kardam utilizes the Lipschitzness of the gradients to filter out outliers. However, Kardam assumes a threat model much weaker than ours. The major differences in the threat model are listed as follows:
\setitemize[0]{leftmargin=*,topsep=0pt}
\begin{itemize}
\item \textbf{Verification of worker identity.} Unlike Kardam, we do not require verifying the identities of the workers when the server receives gradients. Kardam uses the so-called \textit{empirical Lipschitz coefficient}, to test the benignity of the gradient sent by a specific worker. Such a mechanism keeps the record of the \textit{empirical Lipschitz coefficient} of each worker. Thus, whenever a gradient is received, the Kardam server must be able to identify the identity/index of the worker. However, since Byzantine workers can behave arbitrarily, they can fake their identities/indices when sending gradients to the servers. Thus, Kardam assumes a threat model much weaker than the traditional Byzantine failure/threat model. Note that for synchronous training, the server can partially counter the index spoofing attack by simply filtering out all the gradients with duplicated indices. However, such an approach is infeasible for asynchronous training.
\item \textbf{Bounded staleness of workers/limit of successive gradients.} Unlike Kardam, we do not require bounded staleness of the workers. Kardam requires that the number of gradients successively received from a single worker is bounded above. To be more specific, on the server, any sequence of successively received gradients of length $2q+1$ must contain at least $q+1$ gradients from honest workers. However, in real-world asynchronous training, such an assumption is very difficult to satisfy. 
\item \textbf{A majority of honest workers.} Unlike Kardam, we do not require a majority of honest workers. Kardam requires that the number of Byzantine workers is less than one-third of the total number of workers -- much stronger restriction than the standard setting that allows for the number of Byzantine workers to be up to 50\% of the total number of workers. Zeno++ further extends this guarantee to allow for not only 50\%, but also a majority of Byzantine workers.
\end{itemize}

\section{Model}

We consider the following optimization problem: $\min_{x \in \R^d} F(x)$, 
where $F(x) = \frac{1}{m} \sum_{i \in [m]} \E_{z_i \sim \mathcal{D}_i} f(x; z_i)$, for $\forall i \in [m]$, $z_i$ is sampled from the local data $\mathcal{D}_i$ on the $i$th device.

We solve this problem in a distributed manner with $m$ workers. Each worker trains the model on local data. 
In each iteration, the $i$th worker will sample $n$ independent data points from the dataset $\mathcal{D}_i$, and compute the gradient of the local empirical loss $F_i(x) = \frac{1}{n} \sum_{j=1}^n f(x; z_{i,j}), \forall i \in [m]$, where $z_{i,j} \sim \mathcal{D}_i$ is the $j$th sampled data on the $i$th worker. 
When there are no Byzantine failures, the servers update the model whenever a new gradient is received:
\begin{align*}
x_{t+1} &= x_t - \gamma_t g_\tau, \\
g_\tau &= \frac{1}{n} \sum_{j \in [n]} \nabla f(x_\tau; z_{i,j}), \tau \leq t, i \in [m].
\end{align*}

When there are Byzantine failures, $g_\tau$ can be replaced by arbitrary value~\citep{damaskinos2018asynchronous}. 
Formally, we define the threat model as follows.
\begin{definition} 
\label{def:byz}
(Threat Model). When the server receives a gradient estimator $\tilde{g}_\tau$, it is either correct or Byzantine. If sent by a Byzantine worker, $\tilde{g}_\tau$ is assigned arbitrary value. If sent by an honest worker, the correct gradient is $\frac{1}{n} \sum_{j=1}^n \nabla f(x_\tau; z_{i,j}), \tau \leq t, i \in [m]$. Thus, we have
\begin{align*}
\tilde{g}_\tau = 
\begin{cases}
\mbox{arbitrary value}, & \mbox{if the worker is Byzantine}, \\
 \frac{1}{n} \sum_{j=1}^n \nabla f(x_\tau; z_{i,j}), & \mbox{otherwise.}
\end{cases}
\end{align*}
We assume that $q$ out of $m$ workers are Byzantine, where $q \leq m$. Furthermore, the indices of Byzantine workers can change across different iterations.
\end{definition}

\begin{table}[htb]
\caption{Notation}
\label{tbl:notations}
\begin{center}
\begin{small}
\begin{tabular}{|l|l|}
\hline 
Notation  & Description \\ \hline
$m$, $[m]$    & Number of workers, set of integers $\{1, \ldots, m \}$ \\ \hline
$q$    & Number of Byzantine workers \\ \hline
$\DM_i$    & $\DM_i$ is the training dataset on the $i$th worker \\ \hline
$\SM$    & The validation dataset on \textit{Zeno++} server \\ \hline
$n$   & Mini-batch size of workers \\ \hline
$n_s$    & Mini-batch size of \texttt{Zeno++} server \\ \hline
$T$, $t$    & Number of server steps, server step index  \\ \hline
$\gamma$    & Learning rate  \\ \hline
$\rho$, $\epsilon$    & Hyperparameters of Zeno++  \\ \hline
$k$    & Maximum delay of $g_r$, i.e., ``server delay''  \\ \hline
$k_w$    & Maximum delay of workers, i.e., ``worker delay''  \\ \hline
$\| \cdot \|$    & All the norms in this paper are $l_2$-norms  \\ \hline
\end{tabular}
\end{small}
\end{center}
\end{table}

\section{Methodology}

In this section, we introduce \texttt{Zeno++}, a Byzantine-tolerant asynchronous SGD algorithm based on inner-product validation. \texttt{Zeno++} is a computationally efficient version of its prototype: \texttt{Zeno+}.

\subsection{Zeno+}
For completeness, we first introduce an auxiliary algorithm: \texttt{Zeno+}. Note that \texttt{Zeno+} is hard to be applied in practice, due to its heavy computation overhead on the server for validation. We introduce \texttt{Zeno+} only to help understand the concept of \texttt{Zeno++}.

Inspired by \texttt{Zeno}~\citep{Xie2018ZenoBS}, we compute a score for each candidate gradient estimator by using the stochastic zero-order oracle. However, in contrast to the existing synchronous SGD with majority-based aggregation methods, we need a hard threshold to decide whether a gradient is accepted, as sorting is not meaningful in asynchronous settings. This descent score is described next.
\begin{definition}(Stochastic Descent Score~\citep{Xie2018ZenoBS})
\label{def:score}
Denote $f_s(x) = \frac{1}{n_s} \sum_{j=1}^{n_s} f(x; z_j)$, where $z_j$'s are i.i.d. samples drawn from $\SM$, where $\SM \neq \DM_i, \forall i \in [m]$, and $n_s$ is the batch size of $f_s(\cdot)$. For any gradient estimator~(correct or Byzantine) $g$, model parameter $x$, learning rate $\gamma$, and a constant weight $\rho > 0$, we define its \textit{stochastic descent score} as follows:
\begin{align*}
Score_{\gamma, \rho}(g, x) = f_s(x) - f_s(x - \gamma g) - \rho \| g \|^2.
\end{align*} 
\end{definition}

\begin{remark}
Note that we assume that the dataset $\SM$ for computing $f_s(\cdot)$ is different from the training dataset, e.g., can be a separated validation dataset. In other words, $\SM \neq \mathcal{D}_1 \neq \cdots \neq \DM_m \neq \cup_{i=1}^m \DM_i$.
\end{remark}

The score defined in Definition~\ref{def:score} is composed of two parts: the estimated descent of the loss function, and the magnitude of the update. The score increases when the estimated descent of the loss function, $f_s(x) - f_s(x - \gamma g)$, gets larger. We penalize the score by $-\rho \| g \|^2$, so that the change of the model parameter will not be too large. A large descent suggests faster convergence. Observe that even when a gradient is Byzantine, a small magnitude indicates that it will be less harmful to the model. 

Using the \textit{stochastic descent score}, we can set a hard threshold parameterized by $\epsilon$ to filter out candidate gradients with relatively small scores. The detailed algorithm is outlined in Algorithm~\ref{alg:zeno_plus}.

\begin{algorithm}[hbt!]
\caption{Zeno+}
\begin{algorithmic}[]
\NewProcedure{Server}
\STATE $x_0 \leftarrow rand(), t \leftarrow 1$ 
\REPEAT
	\STATE Randomly sample $z_j \sim \SM, \; \forall j \in [n_s]$ to compute $f_s$ (Note: $\SM \neq \DM_1 \neq \cdots \neq \DM_m$)
	\STATE Receive $\tilde{g}$ from an arbitrary worker
	\STATE Normalize $g = c \tilde{g}$ such that $\|g\|^2 = \|\nabla f_s(x_{t-1})\|^2$
	\IF{$Score_{\gamma, \rho}(g, x_{t-1}) \geq -\gamma\epsilon$}
		\STATE $x_t \leftarrow x_{t-1} - \gamma g$, $t \leftarrow t + 1$
	\ENDIF
\UNTIL{Convergence}
\NewProcedure{Worker $i = 1, \ldots, m$}
\IF{The worker is honest}
	\REPEAT
	    \STATE Pull $x_\tau$ from the server
	    \STATE Draw random samples $z_{i,j} \sim \DM_i$, $\forall j \in [n]$, 
	    compute $\tilde{g} \leftarrow \frac{1}{n} \sum_{j \in [n]} \nabla f(x_\tau; z_{i,j})$
	    \STATE Push $\tilde{g}$ to the server
	\UNTIL{Convergence}
\ENDIF
\end{algorithmic}
\label{alg:zeno_plus}
\end{algorithm}

\subsection{Zeno++}

\begin{algorithm}[hbt!]
\caption{Zeno++}
\begin{algorithmic}[]
\NewProcedure{Server}
\STATE $x_0 \leftarrow rand(), t \leftarrow 1$
\REPEAT
	\REPEAT
		\STATE Receive $\tilde{g}$ from an arbitrary worker
		\STATE Read $v$ with lock ($v$ may be from an old version of $x$: $v = \nabla f_s(x_\tau), \tau \leq t-1$)
		\STATE Normalize $g = c \tilde{g}$ such that $\|g\|^2 = \|v\|^2$ \algline{zeno_plusplus:normalize} (\ref{zeno_plusplus:normalize})
	\UNTIL{$\gamma \ip{v}{g} - \rho \| g \|^2 \geq -\gamma\epsilon$ \algline{zeno_plusplus:taylor} (\ref{zeno_plusplus:taylor})}
	\STATE $x_t \leftarrow x_{t-1} - \gamma g$, \quad $t \leftarrow t + 1$
	\STATE Lazy update of $v$: Run non-blocking $ZenoUpdater(x_t)$, if idle, or after every $k$ iterations \algline{zeno_plusplus:lazy_update} (\ref{zeno_plusplus:lazy_update})
\UNTIL{Convergence}
\NewProcedure{ZenoUpdater($x$) on server}
	\STATE Randomly sample $z_j \sim \SM, \; \forall j \in [n_s]$ to compute $f_s$ (Note: $\SM \neq \DM_1 \neq \cdots \neq \DM_m$)
	\STATE Write with lock: $v \leftarrow \nabla f_s(x) = \frac{1}{n_s} \sum_{j=1}^{n_s} \nabla f(x; z_j)$
\NewProcedure{Worker $i = 1, \ldots, m$}
\IF{The worker is honest}
	\REPEAT
	    \STATE Pull $x_\tau$ from the server
	    \STATE Draw random samples $z_{i,j} \sim \DM_i$, $\forall j \in [n]$, 
	    compute $\tilde{g} \leftarrow \frac{1}{n} \sum_{j \in [n]} \nabla f(x_\tau; z_{i,j})$
	    \STATE Push $\tilde{g}$ to the server
	\UNTIL{Convergence}
\ENDIF
\end{algorithmic}
\label{alg:zeno_plusplus}
\end{algorithm}

Calculating the \textit{stochastic descent score} for every candidate gradient can be computationally expensive. To reduce the computation overhead, we approximate it by its first-order Taylor's expansion with stale validation gradients.

\begin{definition}(Approximated Stochastic Descent Score)
\label{def:score_approx}
Denote $f_s(x) = \frac{1}{n_s} \sum_{j=1}^{n_s} f(x; z_j)$, where $z_j$'s are i.i.d. samples drawn from $\SM$, where $\SM \neq \DM_i, \forall i \in [m]$, and $n_s$ is the batch size of $f_s(\cdot)$. 
%$\E[f_s(x)] \neq F(x)$. 
For any gradient estimator~(correct or Byzantine) $g$, model parameter $x$, learning rate $\gamma$, and a constant weight $\rho > 0$, we approximate its \textit{stochastic descent score} as follows:
\begin{align*}
Score_{\gamma, \rho}(g, x) \approx \gamma \ip{\nabla f_s(x)}{g} - \rho \| g \|^2,
\end{align*} 
where $x$ is a stale version of model on the server.
\end{definition}
In brief, \texttt{Zeno++} is a computation-efficient version of \texttt{Zeno+} which reduces the computation overhead via two techniques: \textit{approximated stochastic descent score}, and lazy updates of the validation gradient $v$. The detailed algorithm is shown in Algorithm~\ref{alg:zeno_plusplus}. Compared to \texttt{Zeno}, we highlight several new techniques in \texttt{Zeno++}~(Algorithm~\ref{alg:zeno_plusplus}), specially designed for asynchronous training: 1) re-scaling the candidate gradient~(Line~(\ref{zeno_plusplus:normalize})); 2) first-order Taylor's expansion~(Line~(\ref{zeno_plusplus:taylor})); 3) hard threshold instead of comparison with the others~(Line~(\ref{zeno_plusplus:taylor})); 4) lazy update for reducing the computation overhead~(Line~(\ref{zeno_plusplus:lazy_update})).

Before moving forward, we wish to highlight several practical remarks for \texttt{Zeno++}:
\setitemize[0]{leftmargin=*,topsep=0pt}
\begin{itemize}
\item \textbf{Preparing the validation dataset for Zeno++:} The dataset $\SM$ used for calculating $v$~(the validation gradient of \texttt{Zeno++}) can be collected in many ways. As a machine-learning routine, it is common to separate the entire dataset into 3 parts: training data, validation data, and testing data. Such partition naturally provides the validation dataset for \texttt{Zeno++}. It can also be a separate validation dataset provided by a trusted third party. Another reasonable choice is that, a group of trusted workers can upload local data perturbed by additional noise (to help protect the users' privacy). Typically, the validation dataset is small and different from the training dataset, thus can only be used to validate the gradients, and cannot be directly used for training, as shown in Section~\ref{sec:exp}.
%, and also makes the validation dataset different from the training datasets.
\item \textbf{Scheduling $ZenoUpdater(x)$:} $ZenoUpdater(x)$ updates $v$ in the background. It will only be triggered when the global model $x_t$ is updated and the server is idle. Another scheduling strategy is to trigger $ZenoUpdater(x)$ after every $k$ iterations. Thus, $k$ is the upper bound of the delay of $v$. A reasonable choice is $k=m$, so that ideally $v$ is updated after all the $m$ workers respond.
\item \textbf{Computational efficiency:} We can reduce the computation overhead of the \texttt{Zeno++} server by decreasing the mini-batch size $n_s$, or the frequency of the activation of $ZenoUpdater(x)$. However, doing so will potentially incur larger noise for $v$, which makes a trade-off.
\end{itemize}

\section{Theoretical Guarantees}

In this section, we analyze the error bound of \texttt{Zeno++}~(Algorithm~\ref{alg:zeno_plusplus}) under Byzantine failures.
We start with definitions used in the theoretical analysis. 

\begin{definition} (Smoothness)
\label{def:smooth}
Differentiable $f(x)$ satisfies $L$-smoothness if there exists $L > 0$ such that $\forall x, y$,
$
f(y) - f(x) \leq \ip{\nabla f(x; z)}{y-x} + \frac{L}{2} \|y-x\|^2.
$
\end{definition}

\begin{definition} (Polyak-\L{}ojasiewicz~(PL) inequality)
\label{def:pl}
Differentiable $f(x)$ satisfies the PL inequality~\citep{polyak1963gradient} if there exists $\mu > 0$, such that $\forall x$:
$
f(x) - f(x_*) \leq \frac{1}{2\mu} \| \nabla f(x) \|^2.
$
\end{definition}

\subsection{Main Results}

We analyze the error bound of Algorithm~\ref{alg:zeno_plusplus} for non-convex problems with the following assumption.

\begin{assumption} (Bounded server delay)
\label{asm:delay}
For \texttt{Zeno++}, we assume that the delay of the validation gradient $v$ is upper-bounded. Without loss of generality, suppose the current model is $x_t$, and $v = \nabla f_s(x_\tau)$, where $\tau \leq t$. We assume that $\forall t$, $t - \tau \leq k$.
\end{assumption}

\begin{remark}
Zeno++ does not require bounded delay for the workers. The bounded delay requirement in Assumption~\ref{asm:delay} is only for the validation gradient $v$ on the server.
\end{remark}

\begin{assumption}
There exists at least one global minimum $x_*$, where
$
F(x_*) \leq F(x), \forall x.
$
\label{asm:global_min}
\end{assumption}

\begin{remark}
Note that in Algorithm~\ref{alg:zeno_plusplus}, the counter of server steps $t$ increases only when a candidate gradient passes the validation. Otherwise, the algorithm will get stuck in the validation loop. In the following theoretical analysis, we study the convergence of Zeno++ after $T$ server steps, i.e., there is a sequence of $T$ candidate gradients that pass the validation. Thus, the actually number of candidate gradients received on the server is not the same as $T$. So far, we do not have the theoretical analysis for the acceptance ratio of the candidate gradients. Instead, we report the false-positive rate (the fraction of correct gradients that do not pass the validation) in the empirical results in Section~\ref{sec:exp}.
\end{remark}

\begin{remark}
In the following theorems, we show the upper-bounded errors in the worst cases, where the server is overwhelmed by the Byzantine gradients. Note that since we do not make any assumption on the delay of the workers, the Byzantine workers can keep sending successive arbitrary messages to the server, and prevent the server from receiving any correct gradients. Even in such extreme cases, the bounded errors in our theorems still hold.
\end{remark}

We first analyze the case where the functions satisfy the PL inequality.

\begin{theorem}
\label{thm:pl_zeno++}
Assume that $F(x)$ and $f_s(x)$ are $L$-smooth and satisfy the PL inequality.
Assume that $\forall x$, the correct gradients and validation gradients are upper-bounded: $\| \nabla F(x) \|^2 \leq V_1$, $\| \nabla f_s(x) \|^2 \leq V_1$, and the validation gradients are always non-zero and lower-bounded: $\| \nabla f_s(x) \|^2 \geq V_2$, where $0 < V_2 \leq V_1$. Furthermore, we assume that the validation set is close to the training set, which implies bounded variance: $\E\left[ \|\nabla f_s(x) - \nabla F(x)\|^2 \right] \leq V_3, \forall x$. Taking $\gamma < \min(1, \frac{1}{L})$ and $\rho \geq \frac{\alpha \sqrt{\gamma} V_1}{2\mu V_2}$, after $T$ global updates, Algorithm~\ref{alg:zeno_plusplus} has the error bound:
$
\E\left[ F(x_T) - F(x_*) \right] \leq (1 - \alpha \sqrt{\gamma})^T \left[ F(x_{0}) - F(x_*) \right] 
+ \frac{1}{\alpha} \left( \sqrt{\gamma} V_3 + L^2 k^2 \gamma^{5/2} V_1 + \frac{\sqrt{\gamma}}{2} V_1 + \sqrt{\gamma}\epsilon + \frac{L \gamma^{3/2}}{2} V_1 \right). 
$
\end{theorem}

\begin{remark}
The assumption of the lower bounded gradient $\| \nabla f_s(x) \|^2 \geq V_2$ is necessary. We need $\nabla f_s(x) \neq 0$, so that the normalization in Line 6 and inner product in Line 7 of Algorithm~\ref{alg:zeno_plusplus} are feasible. In practice, if we have a mini-batch with zero gradient $\nabla f_s(x) = 0$ on server, we can simply draw additional samples and add them to the mini-batch, until such gradient is non-zero. 
\end{remark}

For general smooth but non-convex functions, we have the following error bound.

\begin{theorem}
Assume that $F(x)$ and $f_s(x)$ are $L$-smooth and potentially non-convex.
Assume that $\forall x$, the true gradients and validation gradients are upper-bounded: $\| \nabla F(x) \|^2 \leq V_1$, $\| \nabla f_s(x) \|^2 \leq V_1$, and the validation gradients are always non-zero and lower-bounded: $\| \nabla f_s(x) \|^2 \geq V_2$, where $0 < V_2 \leq V_1$. Furthermore, we assume that the validation set is close to the training set, which implies bounded variance: $\E\left[ \|\nabla f_s(x) - \nabla F(x)\|^2 \right] \leq V_3, \forall x$. Taking $\gamma < \min(1, \frac{1}{L})$ and $\rho \geq \frac{\alpha \sqrt{\gamma} V_1}{V_2}$, after $T$ global updates, Algorithm~\ref{alg:zeno_plusplus} has the error bound:
$
\frac{\E\left[ \sum_{t \in [T]} \| \nabla F(x_{t-1}) \|^2 \right]}{T} \leq \frac{\E\left[ F(x_{0}) - F(x_*) \right]}{\alpha \sqrt{\gamma} T} 
 + \frac{1}{\alpha} \left( \sqrt{\gamma} V_3 + L^2 k^2 \gamma^{5/2} V_1 + \frac{\sqrt{\gamma}}{2} V_1 + \sqrt{\gamma}\epsilon + \frac{L \gamma^{3/2}}{2} V_1 \right).
$
%Furthermore, if we take $\gamma = \frac{1}{LT}$, then we have
%$
%\frac{\E\left[ \sum_{t \in [T]} \| \nabla F(x_{t-1}) \|^2 \right]}{T} \leq  \mathcal{O}\left( \frac{1}{\alpha \sqrt{T}} \right).
%$
\end{theorem}

\begin{remark}
$\rho$ controls the trade-off between the acceptance ratio and the convergence rate. Large positive $\rho$ makes the convergence faster, but fewer candidate gradients pass the test of \texttt{Zeno++}. Small positive $\rho$ increases the acceptance ratio, but may also potentially slow down the convergence or incur larger variance. We use $\alpha > 0$ to bridge $\rho$ to the convergence rate and the variance. Larger $\alpha$ makes $\rho$ larger, which improves the convergence rate, but also enlarges the variance. Using non-zero $\epsilon$ potentially results in negative thresholds, which enlarges the acceptance ratio, but also increases the false negative ratio (the ratio of Byzantine gradients that are not filtered out by \texttt{Zeno++}).
\end{remark}

\section{Experiments}
\label{sec:exp}

In this section, we evaluate the proposed algorithm, \texttt{Zeno++}. Note that we do not evaluate the auxiliary algorithm \texttt{Zeno+}, since its computation overhead is too large for practical settings. Due to the space limitation, zoomed figures and additional experiments (including evaluation on an additional types of attacks, and testing the sensitivity to hyperparameters) are presented in the appendix. 
We conduct experiments on two benchmarks: CIFAR-10 image classification dataset~\cite{krizhevsky2009learning}, and WikiText-2 language modeling dataset~\cite{merity2016pointer}. 

\subsection{Baselines}
We use the asynchronous SGD without attacks as the gold standard, referred to as \texttt{AsyncSGD without attack}. Since \texttt{Kardam} is the only previous work on Byzantine-tolerant asynchronous SGD, we use it as the baseline.

One may conjecture that \texttt{Zeno++} is analogous to training on the validation data. To explore this, we consider training only on $\SM$ -- assumed to be clean data on the server, i.e., update the model only using $v = \nabla f_s(x)$ on the server, without using any workers. We call this baseline \texttt{Server-only}. We fine-tune the learning rate and show the best results of \texttt{Server-only}.

In the experiments, we evaluate the convergence of the algorithms w.r.t. the number of epochs. To conduct a fair evaluation, each epoch is a full pass of the training data. In each epoch, the server receives the same number of candidate gradients, including both normal gradients and Byzantine gradients, for all the algorithms. Note that the number of gradients in the global epochs is different from the number of server steps $t$ in Algorithm~\ref{alg:zeno_plusplus}.

\subsection{CIFAR-10}

The CIFAR-10 image classification dataset~\citep{krizhevsky2009learning} is composed of 50k images for training and 10k images for testing. We use a convolutional neural network~(CNN) with 4 convolutional layers followed by 2 dense layers. The detailed network architecture can be found in our submitted source code (will be released upon publication). From the training set, we randomly extracted 2.5k of them as the validation set for \texttt{Zeno++}, the remaining are randomly partitioned onto all the workers.
In each experiment, we launch 10 worker processes. We repeat each experiment 10 times and take the average. Each experiment is composed of 200 epochs, where each epoch is a full pass of the training dataset. We simulate asynchrony by drawing random delay from a uniform distribution in the range of $[0, k_w]$, where $k_w$ is the maximum worker delay~(different from the maximum server delay $k$ of \texttt{Zeno++}). In all the experiments, we take the learning rate $\gamma=0.1$, mini-batch size $n = n_s = 128$, $\rho = 0.002$, $\epsilon=0.1$, $k=10$. 

We use the top-1 accuracy on the testing set and the cross-entropy loss function on the training set as the evaluation metrics. We also report the false positive rate~(FP), which is the ratio of correct gradients that are recognized as Byzantine and filtered out by \texttt{Zeno++} or the \texttt{Kardam} baseline.

\subsubsection{Empirical Results}

\begin{figure*}[htb!]
\centering
\subfigure[Top-1 accuracy on testing set, $k_w = 5$.]{\includegraphics[width=0.245\textwidth,height=3.2cm]{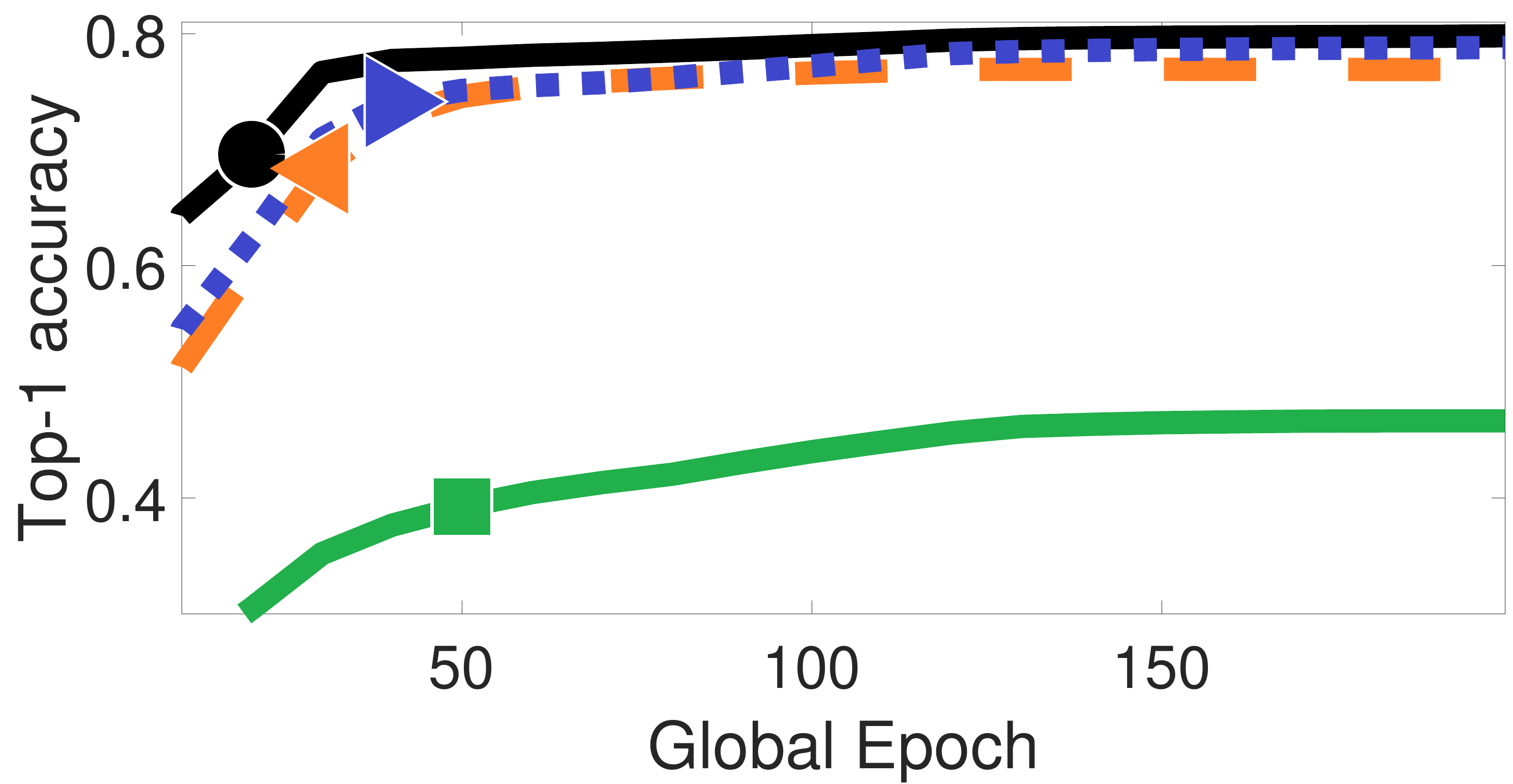}}
\subfigure[Cross entropy on training set, $k_w = 5$.]{\includegraphics[width=0.245\textwidth,height=3.2cm]{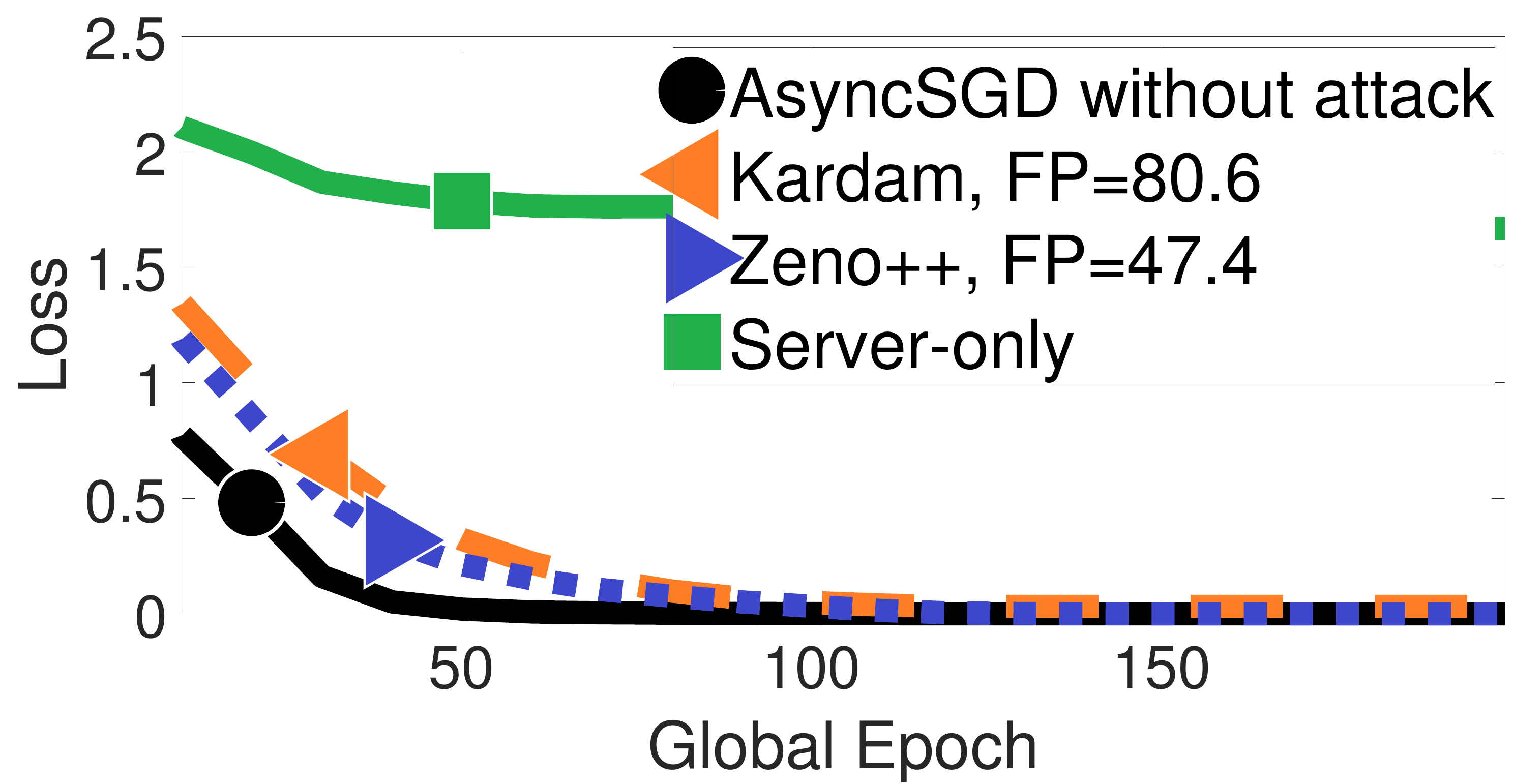}}
\subfigure[Top-1 accuracy on testing set, $k_w = 10$.]{\includegraphics[width=0.245\textwidth,height=3.2cm]{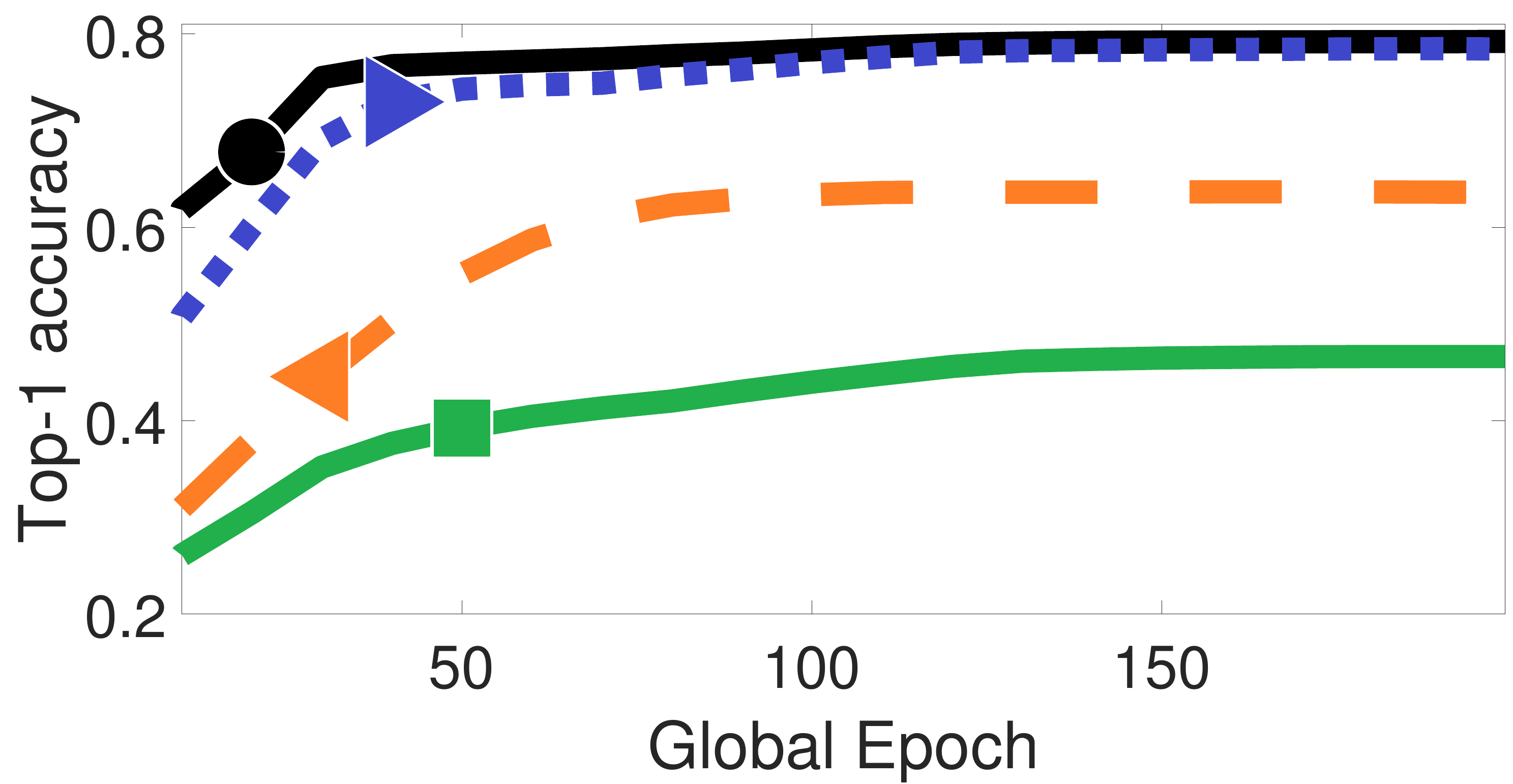}}
\subfigure[Cross entropy on training set, $k_w = 10$.]{\includegraphics[width=0.245\textwidth,height=3.2cm]{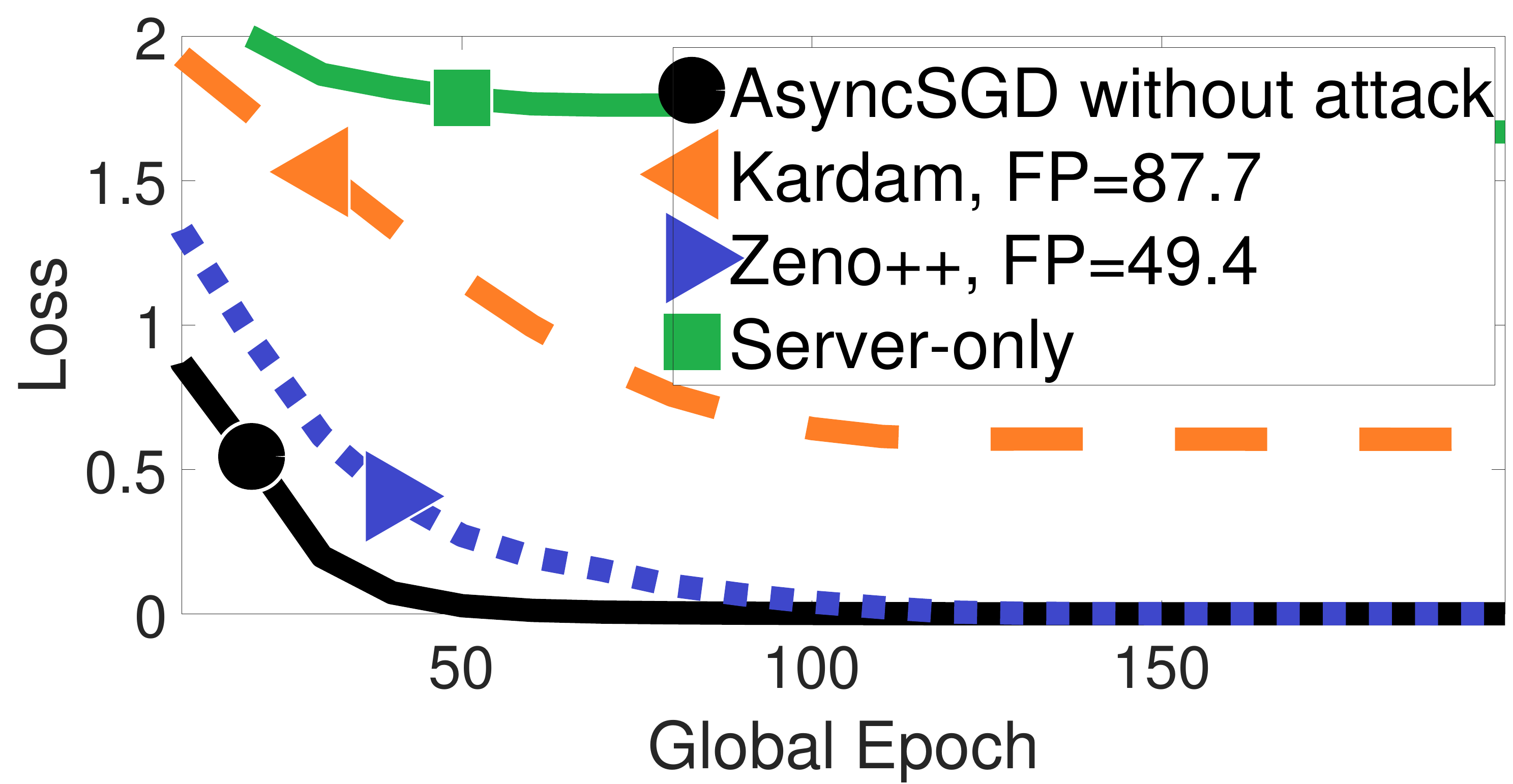}}
\caption{Results on CNN and CIFAR-10, without attacks, with different maximum worker delays $k_w$. $\rho=0.002, \epsilon=0.1, k=10$ for \texttt{Zeno++}. {\em FP} refers to the fraction of false positive detect ions i.e. incorrect prediction that a message is Byzantine.}
\label{fig:nobyz_mini}
\end{figure*}

We first test the convergence when there are no attacks. 
For \texttt{Kardam}, we take $q=2$~(i.e. here \texttt{Kardam} assumes that there are 2 Byzantine workers). The result is shown in Figure~\ref{fig:nobyz_mini}. \texttt{Zeno++} converges a little bit slower than \texttt{AsyncSGD}, but faster than \texttt{Kardam}, especially when the worker delay is large. When $k_w=10$, \texttt{Zeno++} converges much faster than \text{Kardam}. \texttt{Server-only} performs badly on both training and testing data.

\begin{figure*}[htb!]
\centering
\subfigure[Top-1 accuracy on testing set, $k_w = 5$, $q=4$.]{\includegraphics[width=0.245\textwidth,height=3.2cm]{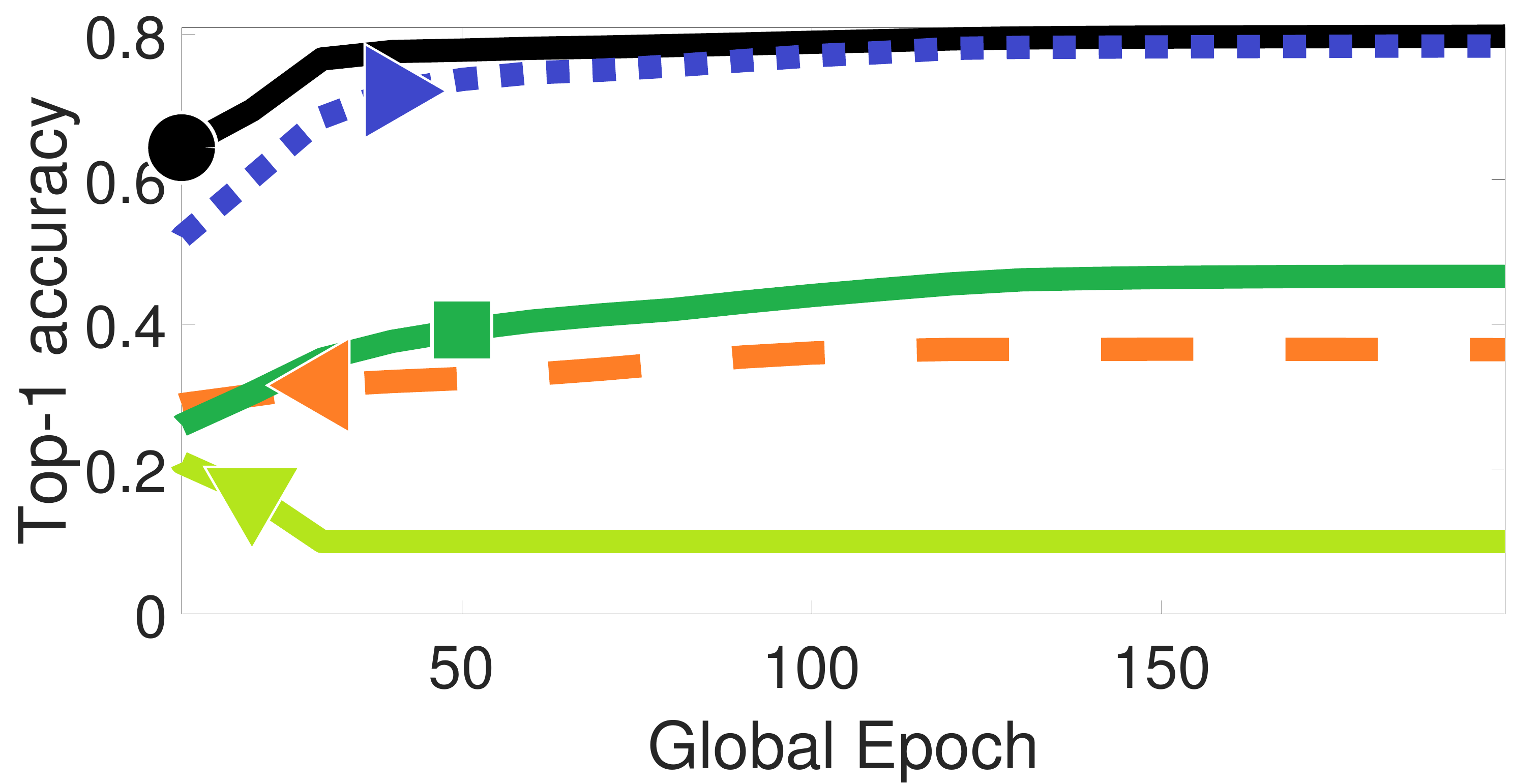}}
\subfigure[Cross entropy on training set, $k_w = 5$, $q=4$.]{\includegraphics[width=0.245\textwidth,height=3.2cm]{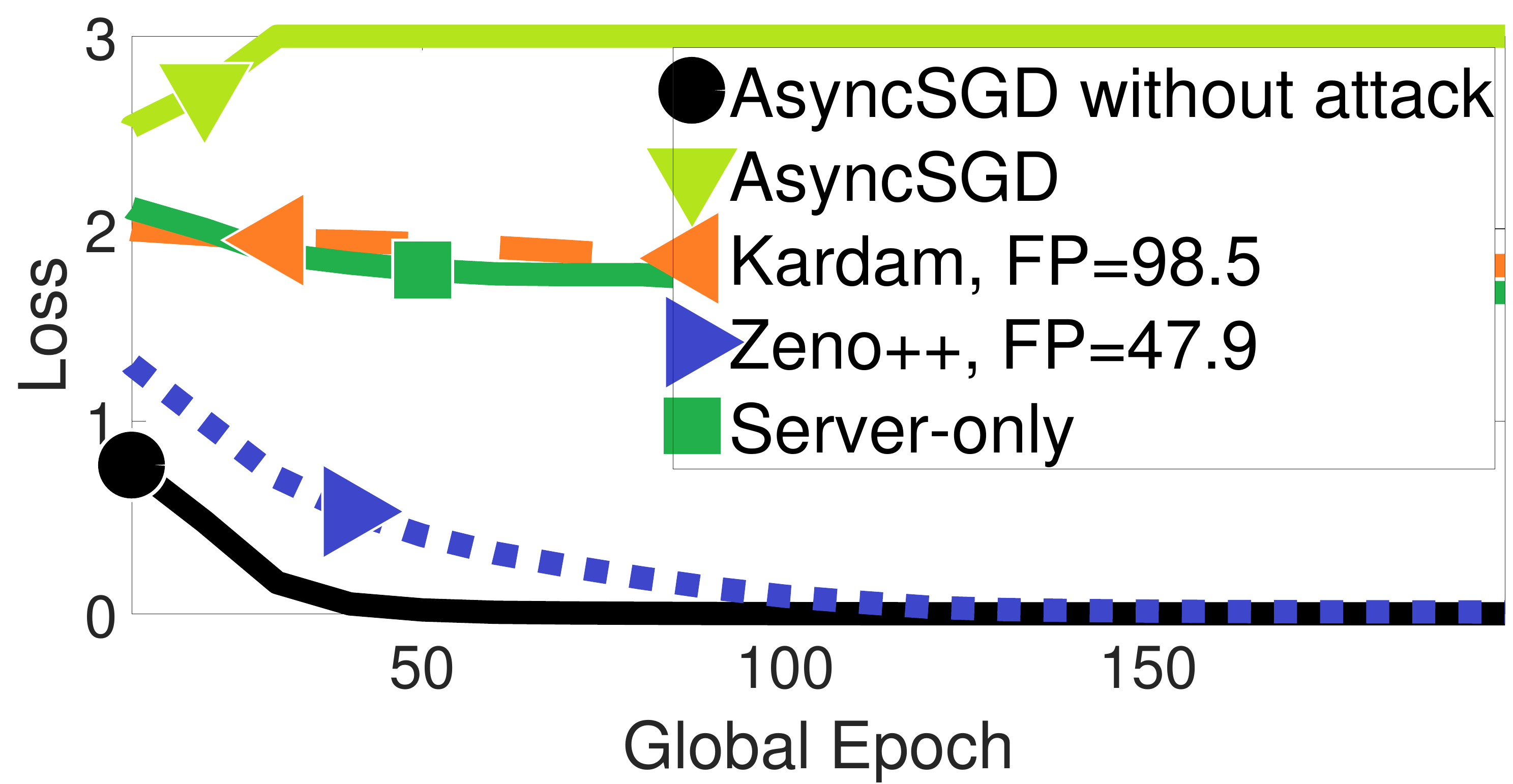}}
\subfigure[Top-1 accuracy on testing set, $k_w = 15$, $q=4$.]{\includegraphics[width=0.245\textwidth,height=3.2cm]{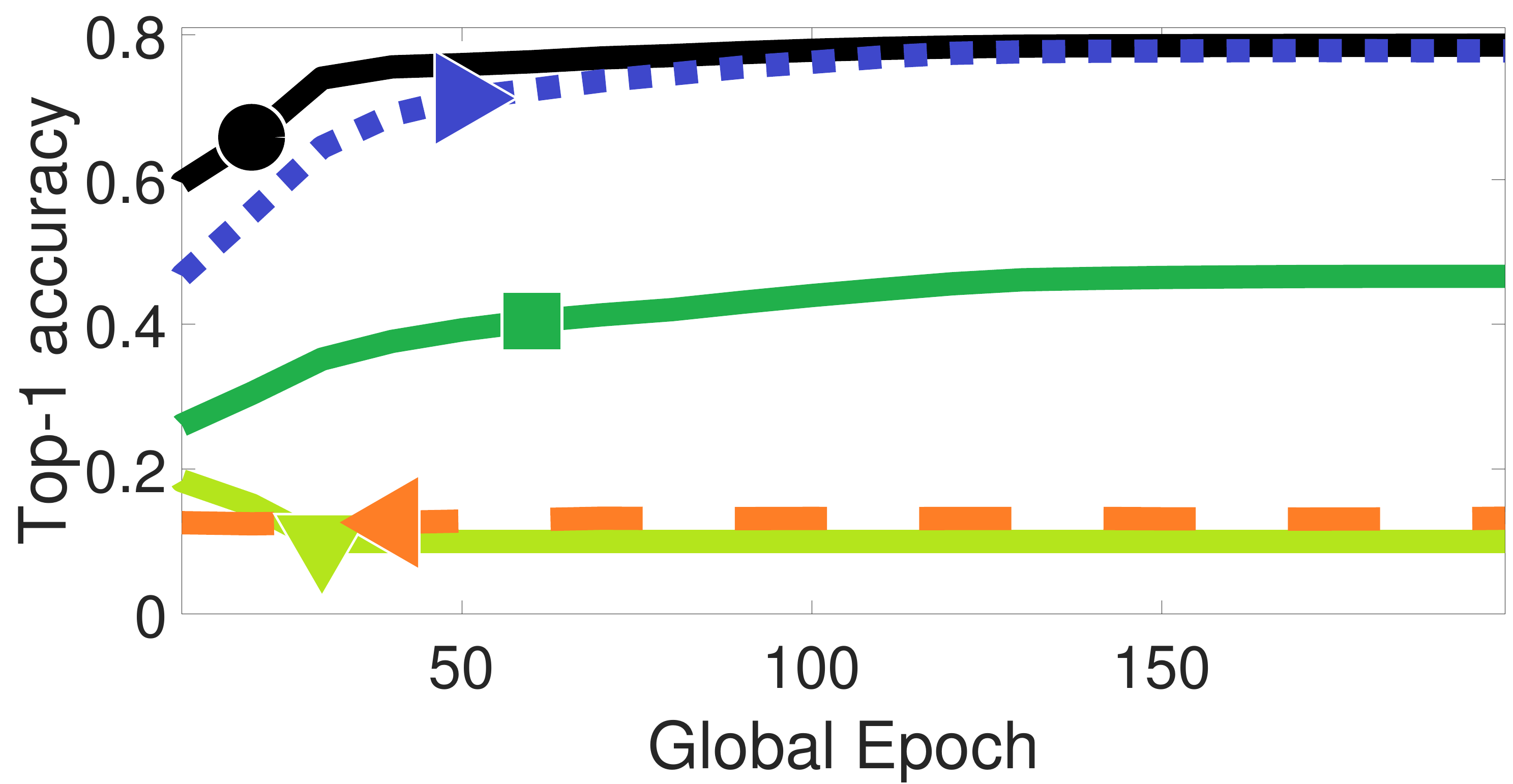}}
\subfigure[Cross entropy on training set, $k_w = 15$, $q=4$.]{\includegraphics[width=0.245\textwidth,height=3.2cm]{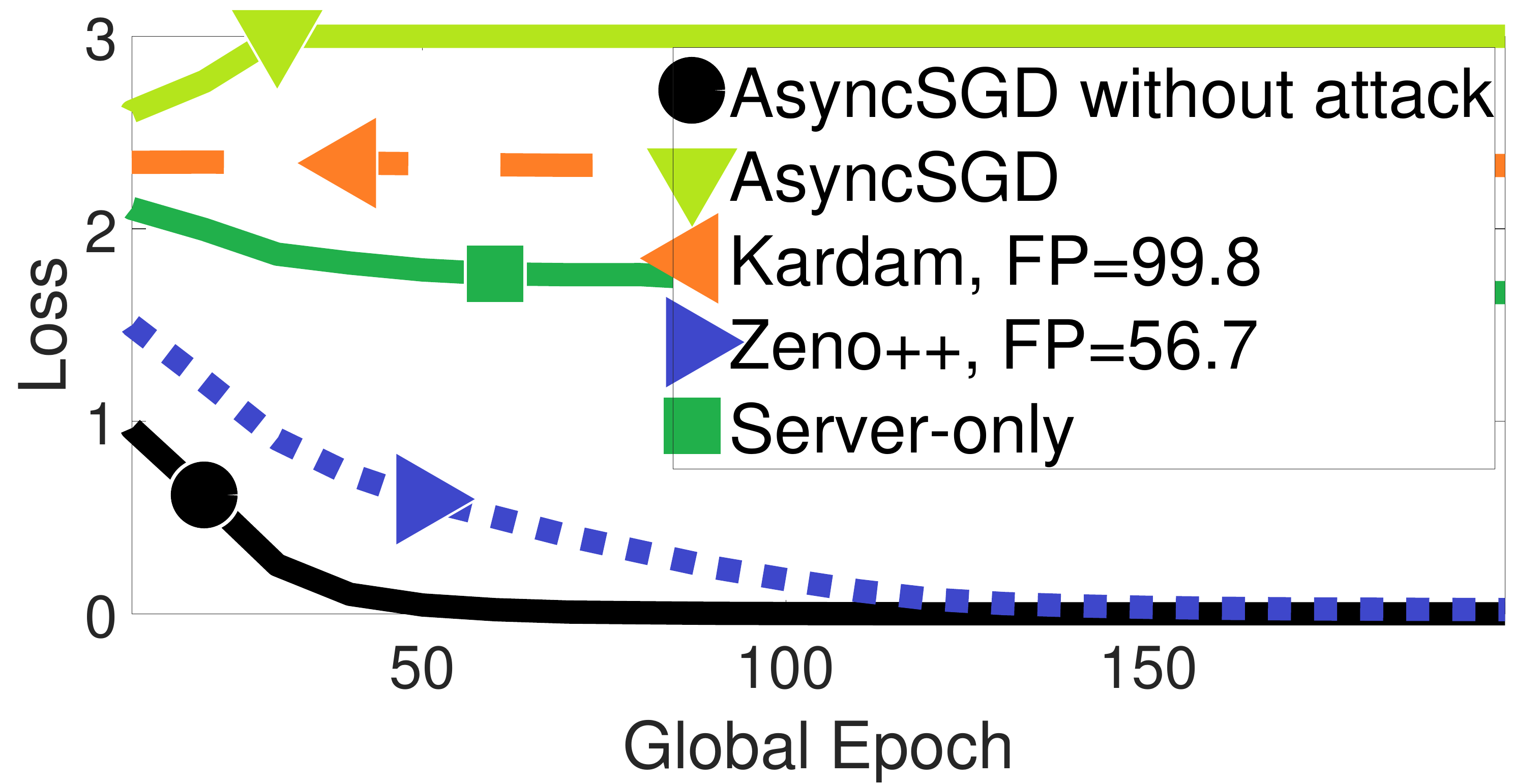}}
\subfigure[Top-1 accuracy on testing set, $k_w = 5$, $q=8$.]{\includegraphics[width=0.245\textwidth,height=3.2cm]{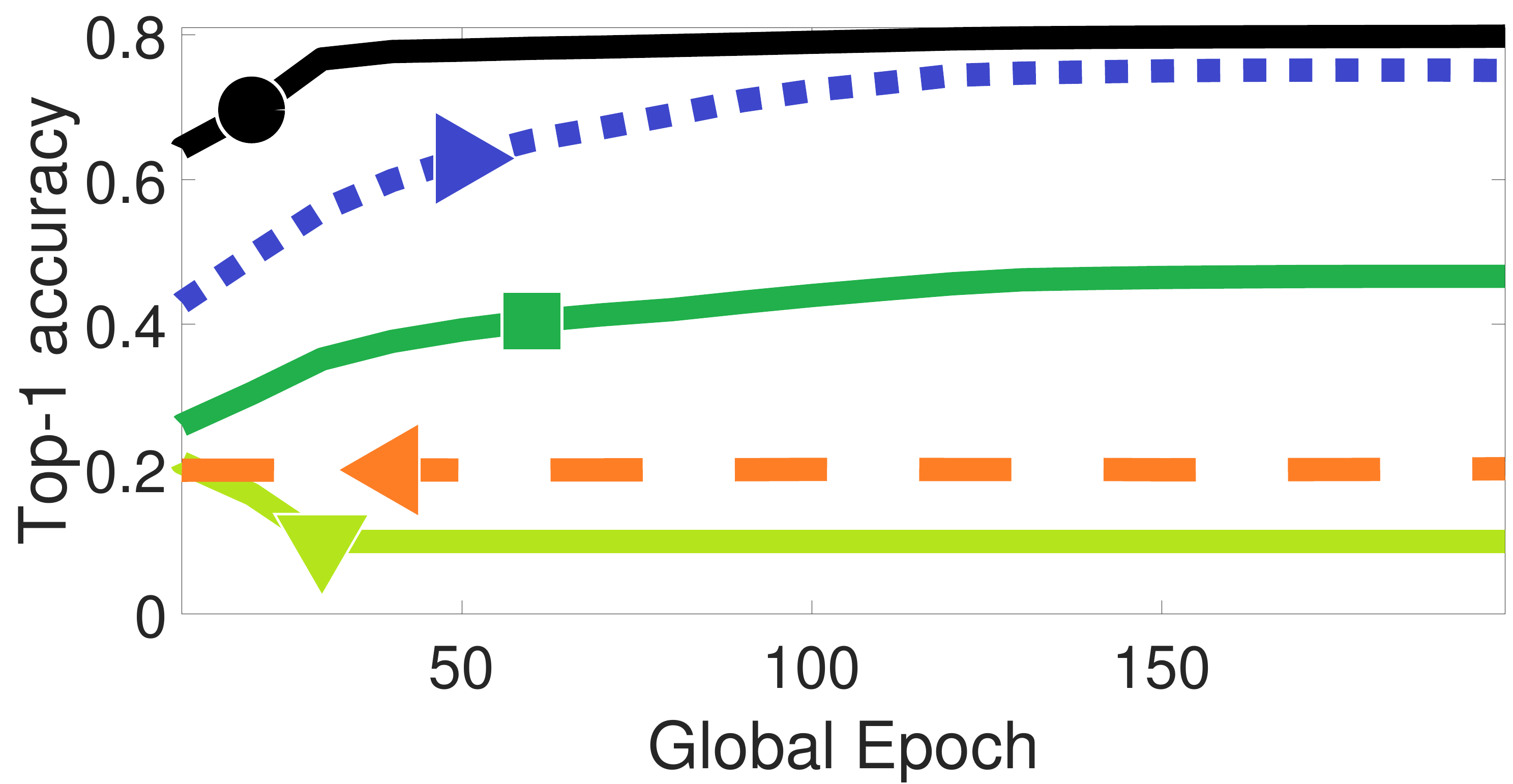}}
\subfigure[Cross entropy on training set, $k_w = 5$, $q=8$.]{\includegraphics[width=0.245\textwidth,height=3.2cm]{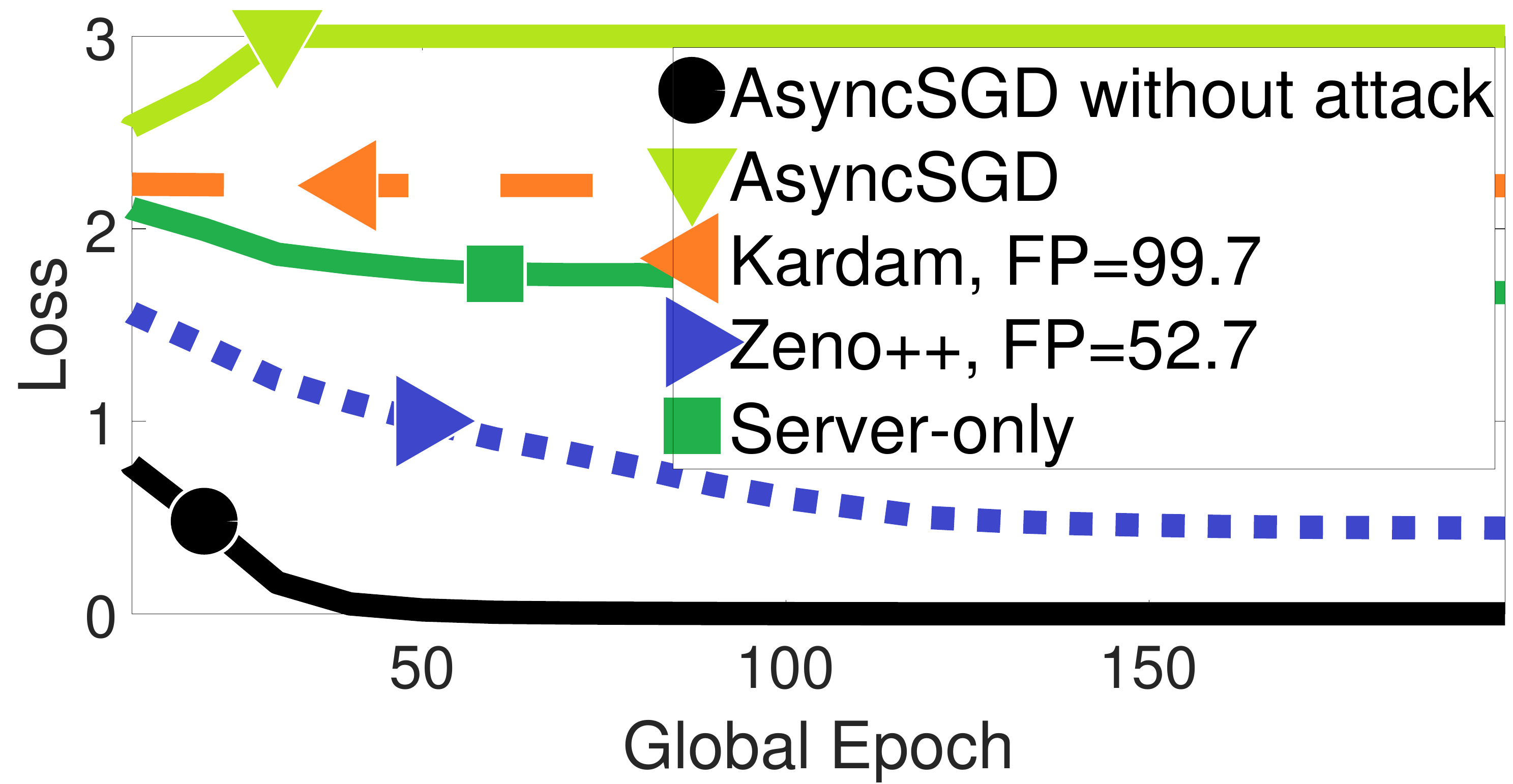}}
\subfigure[Top-1 accuracy on testing set, $k_w = 15$, $q=8$.]{\includegraphics[width=0.245\textwidth,height=3.2cm]{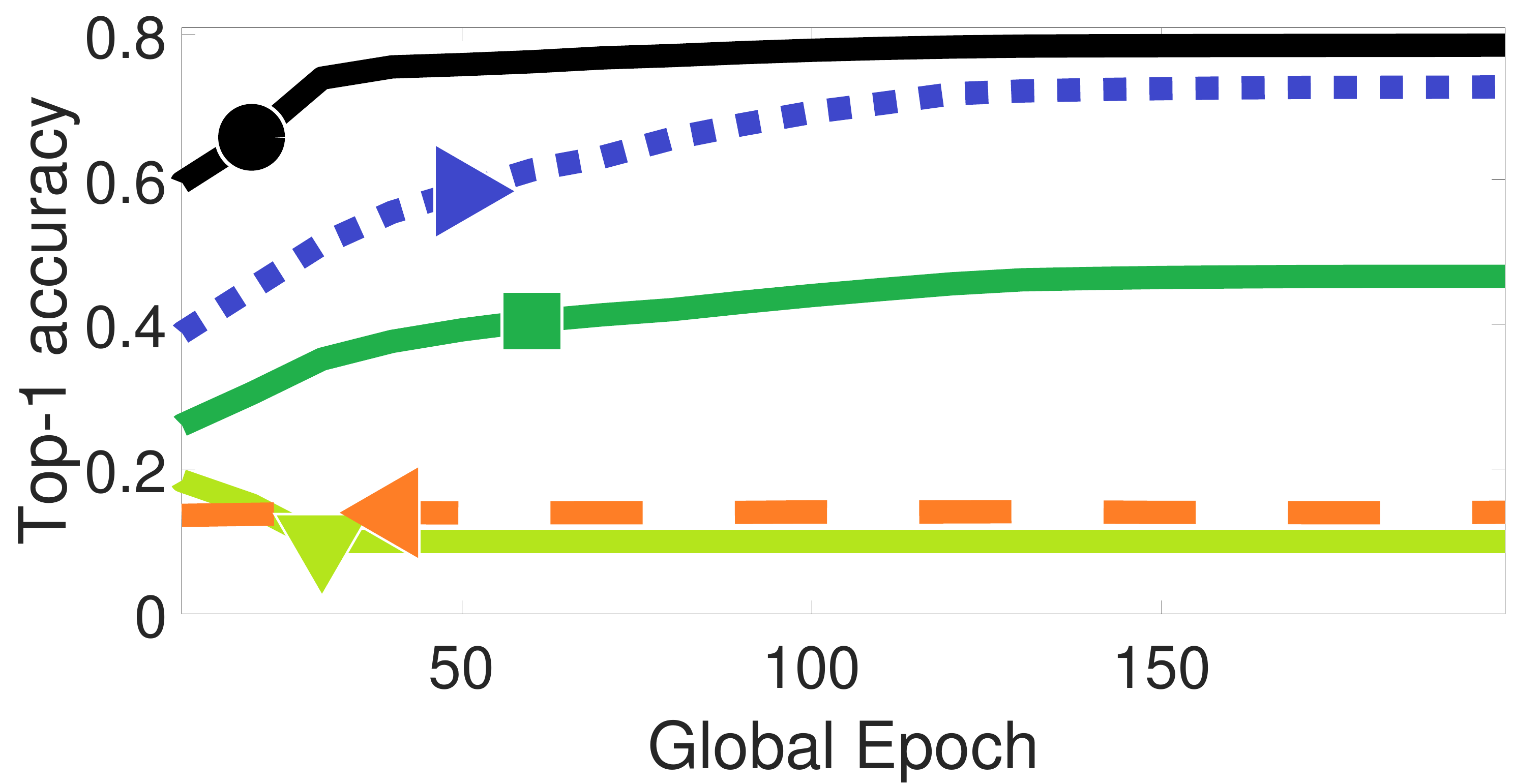}}
\subfigure[Cross entropy on training set, $k_w = 15$, $q=8$.]{\includegraphics[width=0.245\textwidth,height=3.2cm]{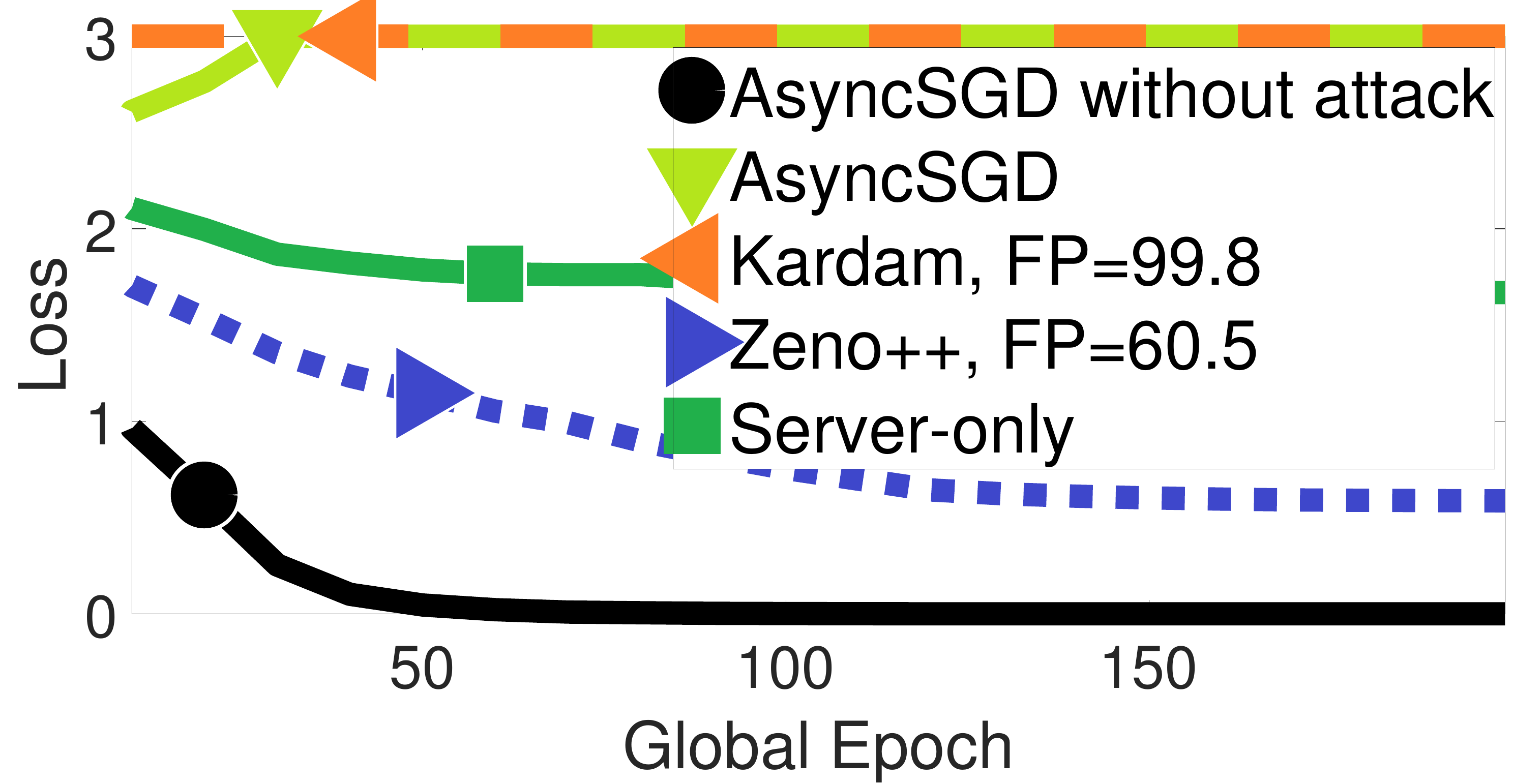}}
\caption{Results on CNN and CIFAR-10, with sign-flipping attacks, and different maximum worker delays $k_w$. For any correct gradient $g$, if selected to be Byzantine, $g$ will be replaced by $-10g$. $q \in \{4,8\}$ out of the 10 workers are Byzantine. $\rho=0.002, \epsilon=0.1, k=10$ for \texttt{Zeno++}. {\em FP} refers to the fraction of false positive detect ions i.e. incorrect prediction that a message is Byzantine.}
\label{fig:signflip_mini}
\end{figure*}

We test the Byzantine-tolerance to the ``sign-flipping'' attack, which was proposed in \cite{damaskinos2018asynchronous}. In such attacks, the Byzantine workers send $-10 \nabla f(x)$ instead of the correct gradient $\nabla f(x)$ to the server. 
% In all the experiments, we take the learning rate $\gamma=0.1$, mini-batch size $n = n_s = 128$, $\rho = 0.002$, $\epsilon=0.1$, $k=10$. 
The result is shown in Figure~\ref{fig:signflip_mini}, with different number of Byzantine workers $q$. It is shown that when $q=4$, \texttt{Zeno++} converges slightly slower than \texttt{AsyncSGD} without attacks, and much faster than \texttt{Kardam}. Actually, we observe that \texttt{Kardam} fails to make progress when the worker delay is large. When the number of Byzantine workers gets larger~($q=8$), the convergence of \texttt{Zeno++} gets slower, but it still makes reasonable progress, while \texttt{AsyncSGD} and \texttt{Kardam} fail.
Note that \texttt{Kardam} performs even worse than \texttt{Server-only}, which means that \texttt{Kardam} is not even as good as training on a single honest worker. Thus, when there are Byzantine workers, distributed training with \texttt{Kardam} is meaningless.

\begin{figure*}[htb!]
\centering
\subfigure[Sensitivity to $\rho$ and $\epsilon$, $k=10$, $q=4$.]{\includegraphics[width=0.495\textwidth,height=4cm]{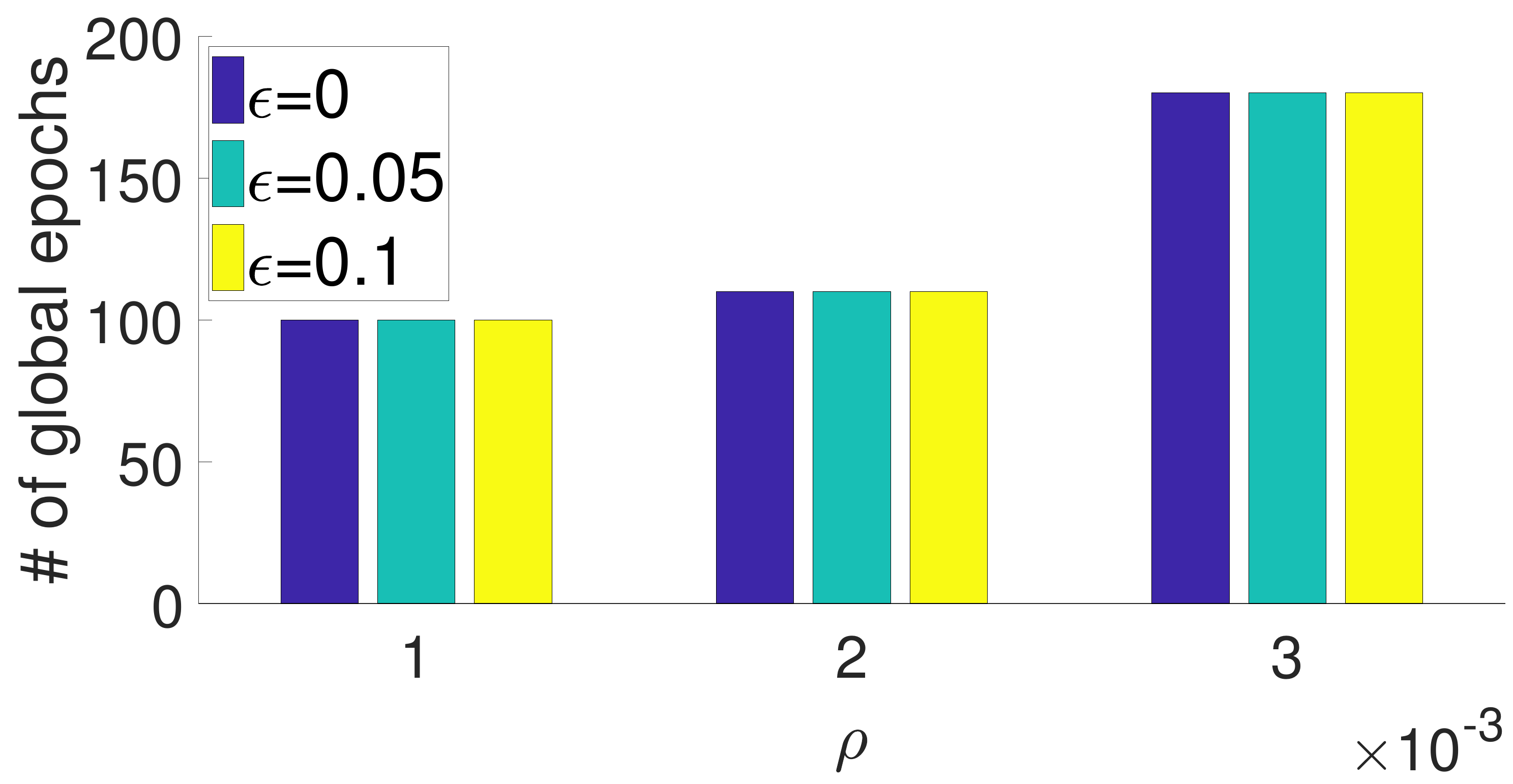}}
\subfigure[Sensitivity to $\rho$ and $k$, $\epsilon=0$, $q=6$.]{\includegraphics[width=0.495\textwidth,height=4cm]{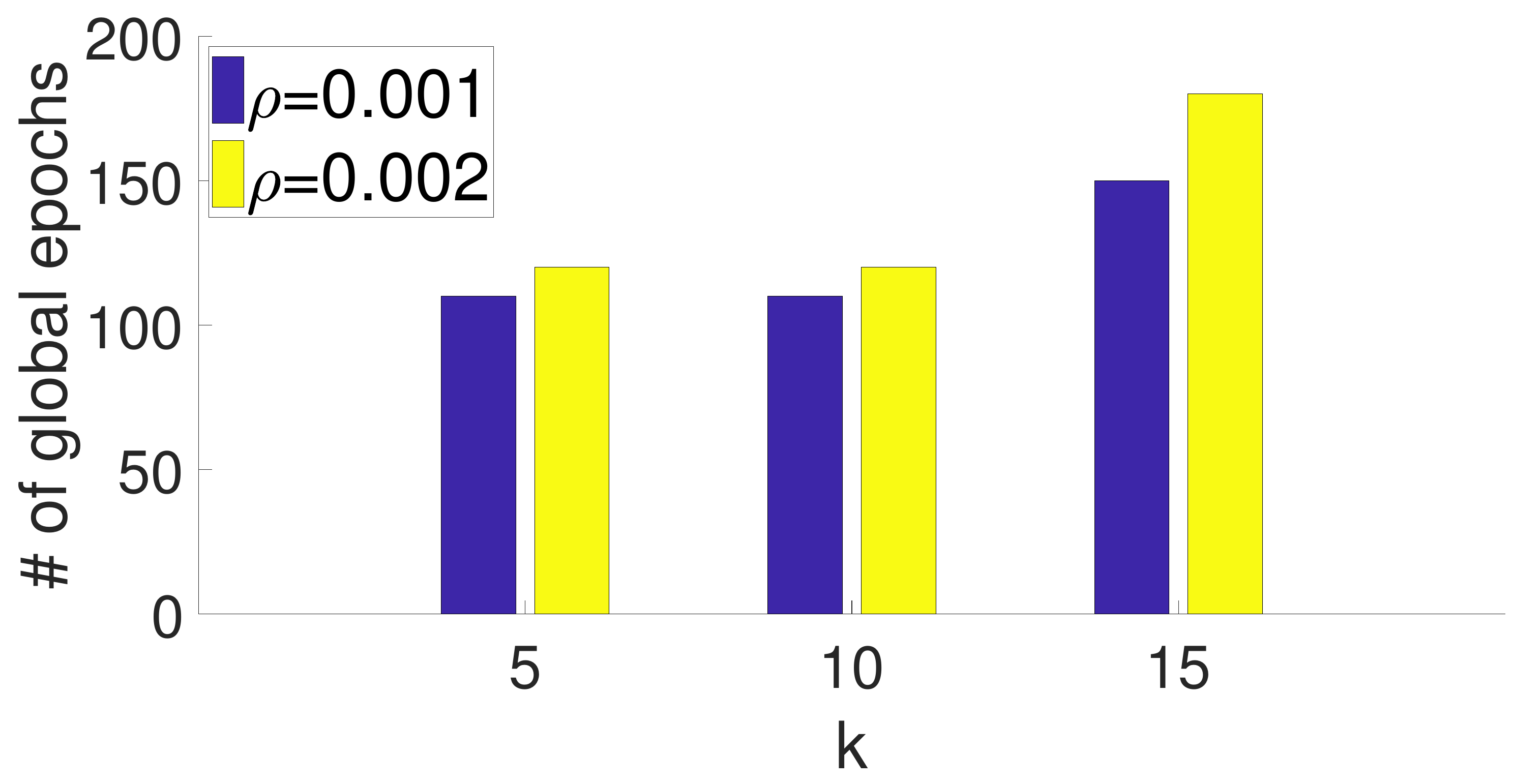}}
\caption{Sensitivity to the hyperparameters on CNN and CIFAR-10. We test the number of global epochs to reach training loss value $0.2$, with sign-flipping attacks. $k_w = 15$.}
\label{fig:signflip_sensitivity}
\end{figure*}

In Figure~\ref{fig:signflip_sensitivity}, we show how the hyperparameters $\rho$, $\epsilon$, and $k$ affect the convergence. In general, \texttt{Zeno++} is insensitive to $\epsilon$. Larger $\rho$ and $k$ slow down the convergence.

\begin{figure*}[htb!]
\centering
\subfigure[Perplexity on testing set, $q=0$.]{\includegraphics[width=0.33\textwidth,height=3.4cm]{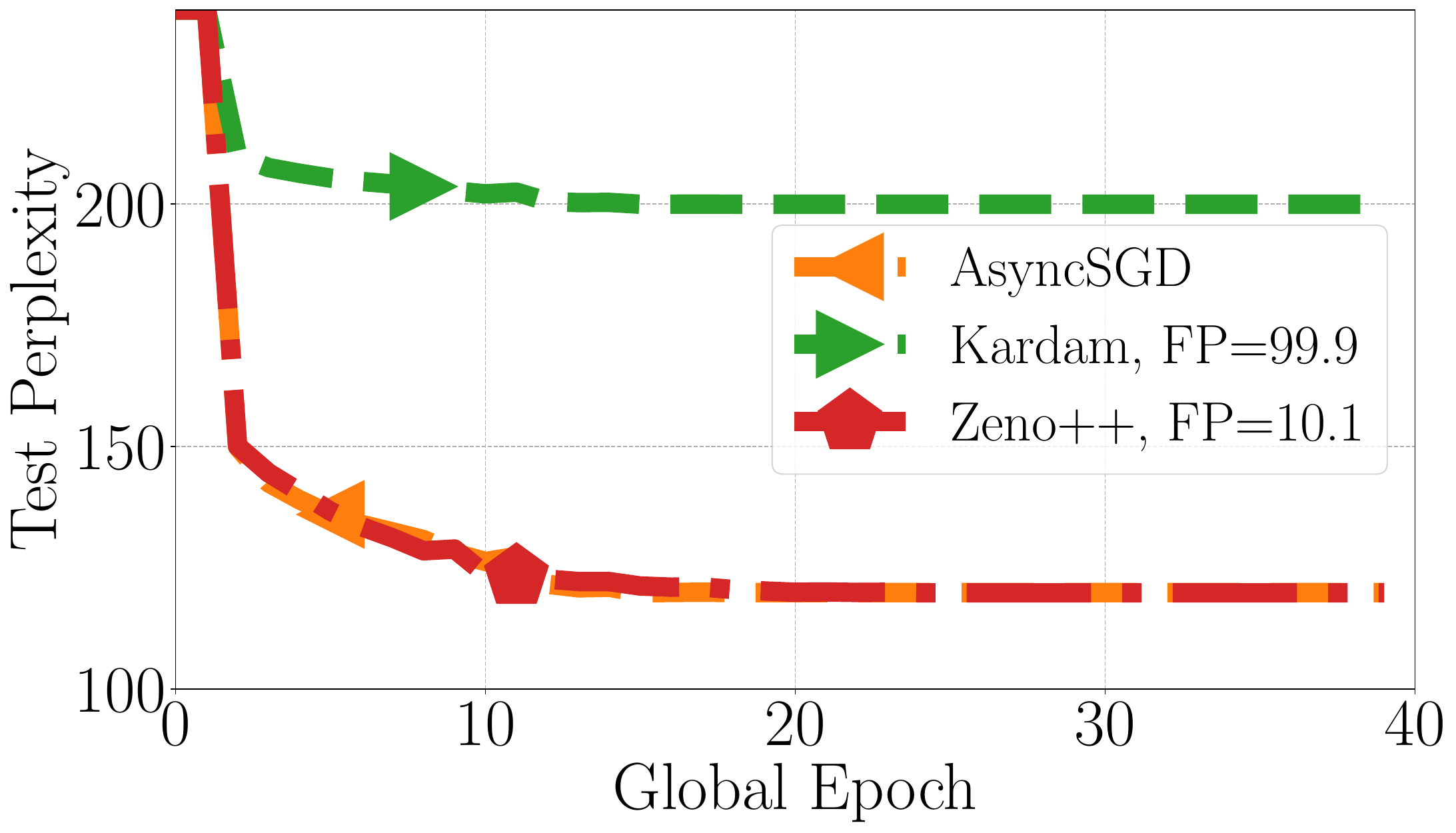}}
\subfigure[Perplexity on testing set, $q=4$.]{\includegraphics[width=0.33\textwidth,height=3.4cm]{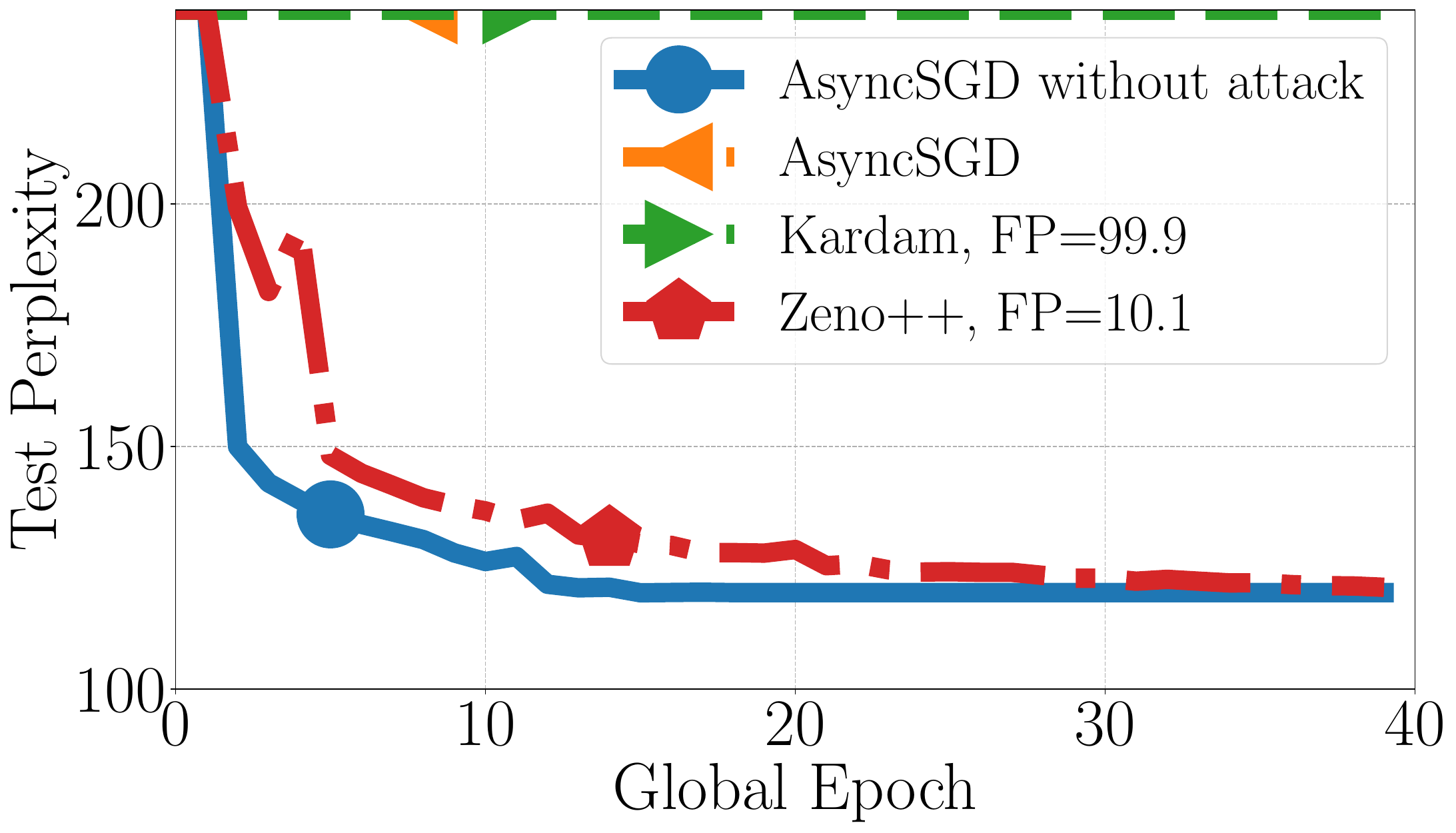}}
\subfigure[Perplexity on testing set, $q=6$.]{\includegraphics[width=0.33\textwidth,height=3.4cm]{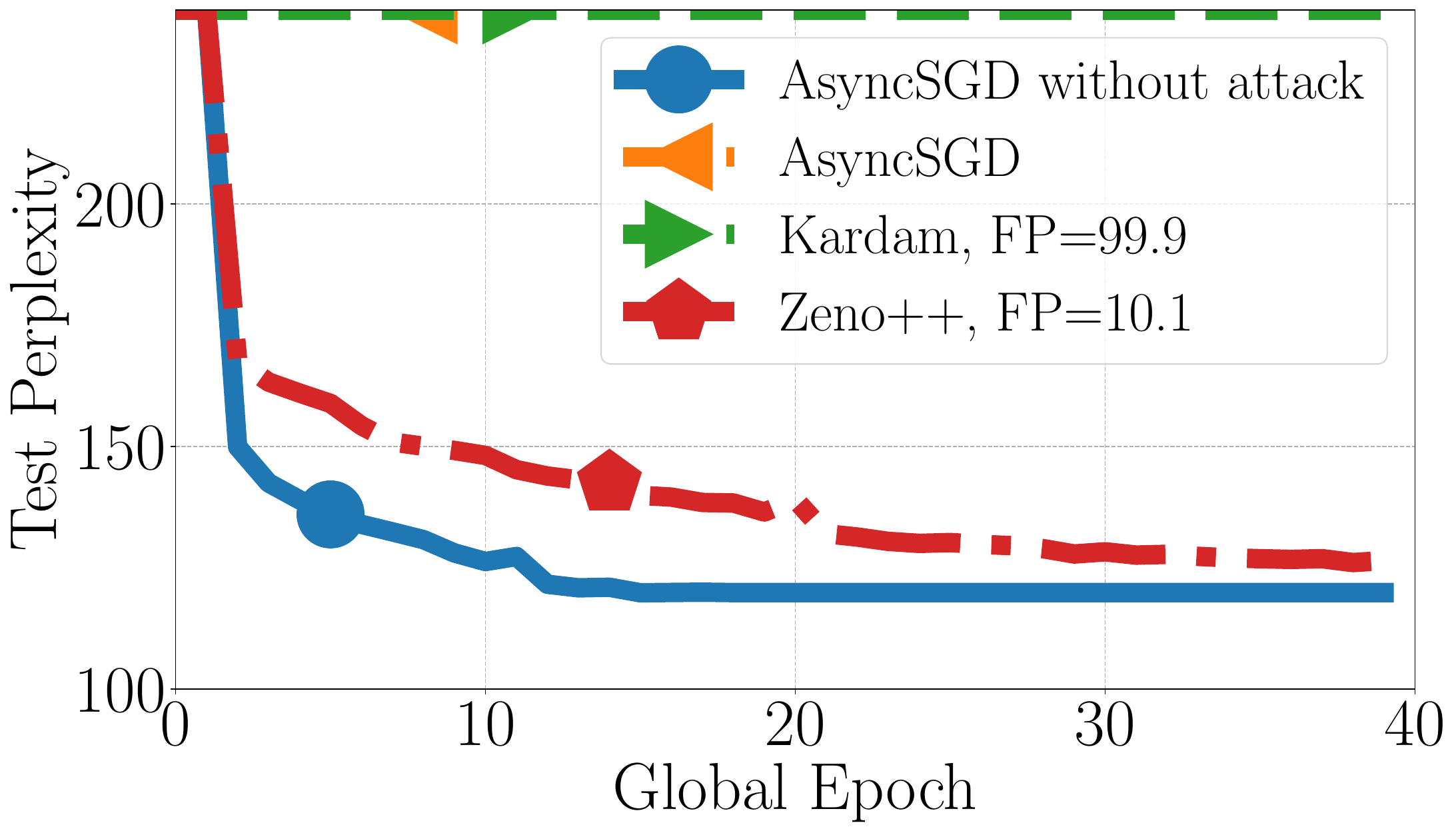}}
\caption{Perplexity~(the lower the better) on LSTM-based language model and WikiText-2 dataset. We show the convergence under sign-flipping attacks with maximum worker delays $k_w = 10$. For any correct gradient $g$, if selected to be Byzantine, $g$ will be replaced by $-10g$. $q \in \{0,4,6\}$ out of the 10 workers are Byzantine. $\gamma=20$ for all the algorithms. $\rho=10, \epsilon=2.0, k=10$ for \texttt{Zeno++}. {\em FP} refers to the fraction of false positive detect ions i.e. incorrect prediction that a message is Byzantine. Note that when $q=4$ or $6$, the results of Kardam are off the charts.}
\label{fig:signflip_wikitext_mini}
\end{figure*}

\subsubsection{Discussion}
\texttt{Kardam} performs surprisingly badly in our experiments. The experiments in \cite{damaskinos2018asynchronous} focus on dampening staleness when there are no Byzantine failures. For Byzantine tolerance, \citet{damaskinos2018asynchronous} only reports that \texttt{Kardam} filters out 100\% of the Byzantine gradients, which matches the results in our experiments. However, we observe that in addition to filtering out 100\% of the Byzantine gradients, \texttt{Kardam} also filters nearly 100\% of the correct gradients. In Figure~\ref{fig:signflip_mini}, we report that the false positive rate of \texttt{Kardam} is nearly 99\%, which makes the convergence extremely slow. To make things worse, \texttt{Kardam} does not even perform as good as \texttt{Server-only}. For these reasons, we discourage the use of \texttt{Kardam} for distributed training.
One reason why \texttt{Kardam} performs badly is that we use a more general threat model in this paper, which does not guarantee an important assumption of \texttt{Kardam}, namely ``any sequence of successively received gradients of length $2q+1$ must contain at least $q+1$ gradients from honest workers''. It is clear that this assumption is quite strong, as in an asynchronous setting, Byzantine workers can easily send long sequences of erroneous responses. Our approach does not depend on such a strong assumption.

In all the experiments, \texttt{Zeno++} converges faster than the baselines when there are Byzantine failures. Although the convergence of \texttt{Zeno++} is slower than \texttt{AsyncSGD} when there are no attacks, we find that it provides a reasonable trade-off between security and convergence speed. In general, larger worker delay $k_w$ and more Byzantine workers $q$ add more error and noise to the gradients, which slows down the convergence, because there are fewer valid gradients for the server to use. \texttt{Zeno++} can filter out most of the harmful gradients at the cost of $\text{FP} \approx 50\%$.

Note that \texttt{Server-only} is an extreme case that only uses the server and the validation dataset to train the model in a non-distributed manner, which will not be affected by Byzantine workers. However, only using the validation data is not enough for training, as shown in Figure~\ref{fig:nobyz_mini}. Similarly in practice, we can use a small dataset separated from the training data for cross-validation, but will never directly train the model only on such validation dataset. Furthermore, as shown in Figure~\ref{fig:signflip_mini}, \texttt{Zeno++} performs much better than \texttt{Server-only}. Thus, we can draw the conclusion that \texttt{Zeno++} is efficiently training the model on the honest workers in a distributed manner, which is not equivalent to training on the validation dataset only.

On average, the server computes $\frac{n_s}{k} = 12.8$ gradients in each iteration, since the validation gradient $v$ of \texttt{Zeno++} is updated after every $k=10$ iterations. Thus, the workload on the server is much smaller than a worker. Furthermore, since we can parallelize the workload on the server and workers, the computation overhead of $v$ can be hidden, so that \texttt{Zeno++} can benefit from distributed training.

\subsection{WikiText-2}

The WikiText-2 dataset contains over 2 million tokens from Wikipedia articles, and annotations from Wikidata.
We train a LSTM-based language model with one LSTM layer, and one dense layer. The dimension of the hidden state is 200. The back propagation through time~(BPTT) is 35.
In all the experiments, we take the learning rate $\gamma=20$, mini-batch size $n=n_s=20$, $k = k_w = 10$, $\rho=10$, $\epsilon=2$. Each experiment runs 40 epochs, where each epoch is a full pass of the training data. The other basic settings are the same as the experiments on CIFAR-10.

We use the perplexity~(the exponential of the loss value) on the testing set as the evaluation metric. We also report the false positive rate~(FP), which is the ratio of correct gradients that are recognized as Byzantine and filtered out by \texttt{Zeno++} or the \texttt{Kardam} baseline.

\subsubsection{Empirical Results}

In Figure~\ref{fig:signflip_wikitext_mini}, we test the Byzantine-tolerance to the ``sign-flipping'' attack, on LSTM-based language models and WikiText-2 dataset. It is shown that when there are no Byzantine workers, \texttt{Zeno++} converges as fast as vanilla asynchronous SGD. When there are Byzantine workers, \texttt{Zeno++} converges slightly slower than \texttt{AsyncSGD} without attacks, and much faster than \texttt{Kardam}. When the number of Byzantine workers gets larger, the convergence of \texttt{Zeno++} gets slower. In overall, on the language models, \texttt{Zeno++} achieves performance similar to what we observe for CNN.

\subsubsection{Discussion}

We show that \texttt{Zeno++} can protect asynchronous SGD from Byzantine failures on the workers in multiple applications, including image classification and language models. Compared to CIFAR-10, with smaller $k_w$ and $q$ for WikiText-2, the FP of \texttt{Zeno++} can be as low as $10\%$. Note that the choices of $\rho$ and $\epsilon$ depends on the learning rate $\gamma$. Large $\gamma$ requires large $\rho$ and $\epsilon$. Kardam still performs badly due to the large false positive rates.

\section{Conclusion}

We propose a novel Byzantine-tolerant fully asynchronous SGD algorithm: Zeno++. Zeno++ provably converges. Our empirical results show good performance compared to previous work. In future work, we will explore variations of our approach for other settings such as federated learning.

\section*{Acknowledgements}
This work was funded in part by the following grants: NSF IIS 1909577, NSF CNS 1908888, and a JP Morgan Chase Fellowship, along with computational resources donated by Intel, AWS, and Microsoft Azure.

\bibliography{zenoasync_icml2020}

\begin{thebibliography}{25}
\providecommand{\natexlab}[1]{#1}
\providecommand{\url}[1]{\texttt{#1}}
\expandafter\ifx\csname urlstyle\endcsname\relax
  \providecommand{\doi}[1]{doi: #1}\else
  \providecommand{\doi}{doi: \begingroup \urlstyle{rm}\Url}\fi

\bibitem[Alistarh et~al.(2018)Alistarh, Allen-Zhu, and
  Li]{alistarh2018byzantine}
Alistarh, D., Allen-Zhu, Z., and Li, J.
\newblock {Byzantine Stochastic Gradient Descent}.
\newblock In \emph{NeurIPS}, 2018.

\bibitem[Blanchard et~al.(2017)Blanchard, Mhamdi, Guerraoui, and
  Stainer]{blanchard2017machine}
Blanchard, P., Mhamdi, E. M.~E., Guerraoui, R., and Stainer, J.
\newblock {Machine Learning with Adversaries: Byzantine Tolerant Gradient
  Descent}.
\newblock In \emph{NeurIPS}, 2017.

\bibitem[Cao \& Lai(2018)Cao and Lai]{Cao2018RobustDG}
Cao, X. and Lai, L.
\newblock {Robust Distributed Gradient Descent with Arbitrary Number of
  Byzantine Attackers}.
\newblock In \emph{ICASSP}. IEEE, 2018.

\bibitem[Chen et~al.(2018)Chen, Wang, Charles, and
  Papailiopoulos]{Chen2018DRACOBD}
Chen, L., Wang, H., Charles, Z.~B., and Papailiopoulos, D.~S.
\newblock {DRACO: Byzantine-resilient Distributed Training via Redundant
  Gradients}.
\newblock In \emph{ICML}, 2018.

\bibitem[Chen et~al.(2017)Chen, Su, and Xu]{Chen2017DistributedSM}
Chen, Y., Su, L., and Xu, J.
\newblock {Distributed Statistical Machine Learning in Adversarial Settings:
  Byzantine Gradient Descent}.
\newblock \emph{POMACS}, 1:\penalty0 44:1--44:25, 2017.

\bibitem[Damaskinos et~al.(2018)Damaskinos, Mhamdi, Guerraoui, Patra, and
  Taziki]{damaskinos2018asynchronous}
Damaskinos, G., Mhamdi, E. M.~E., Guerraoui, R., Patra, R., and Taziki, M.
\newblock {Asynchronous Byzantine Machine Learning (the case of SGD)}.
\newblock In \emph{ICML}, 2018.

\bibitem[Dutta et~al.(2018)Dutta, Joshi, Ghosh, Dube, and
  Nagpurkar]{Dutta2018SlowAS}
Dutta, S., Joshi, G., Ghosh, S., Dube, P., and Nagpurkar, P.
\newblock {Slow and Stale Gradients Can Win the Race: Error-Runtime Trade-offs
  in Distributed SGD}.
\newblock In \emph{AISTATS}, 2018.

\bibitem[Feng et~al.(2014)Feng, Xu, and Mannor]{feng2014distributed}
Feng, J., Xu, H., and Mannor, S.
\newblock {Distributed Robust Learning}.
\newblock \emph{arXiv preprint arXiv:1409.5937}, 2014.

\bibitem[Huber(2004)]{huber2011robust}
Huber, P.~J.
\newblock \emph{{Robust Statistics}}, volume 523.
\newblock John Wiley \& Sons, 2004.

\bibitem[Krizhevsky(2009)]{krizhevsky2009learning}
Krizhevsky, A.
\newblock {Learning Multiple Layers of Features from Tiny Images}.
\newblock 2009.

\bibitem[Lamport et~al.(1982)Lamport, Shostak, and Pease]{lamport1982byzantine}
Lamport, L., Shostak, R., and Pease, M.
\newblock {The Byzantine Generals Problem}.
\newblock \emph{ACM Transactions on Programming Languages and Systems
  (TOPLAS)}, 4\penalty0 (3):\penalty0 382--401, 1982.

\bibitem[Lian et~al.(2018)Lian, Zhang, Zhang, and Liu]{Lian2018AsynchronousDP}
Lian, X., Zhang, W., Zhang, C., and Liu, J.
\newblock {Asynchronous Decentralized Parallel Stochastic Gradient Descent}.
\newblock In \emph{ICML}, 2018.

\bibitem[Merity et~al.(2017)Merity, Xiong, Bradbury, and
  Socher]{merity2016pointer}
Merity, S., Xiong, C., Bradbury, J., and Socher, R.
\newblock {Pointer Sentinel Mixture Models}.
\newblock In \emph{ICLR}, 2017.

\bibitem[Mhamdi et~al.(2018)Mhamdi, Guerraoui, and Rouault]{mhamdi2018hidden}
Mhamdi, E. M.~E., Guerraoui, R., and Rouault, S.
\newblock {The Hidden Vulnerability of Distributed Learning in Byzantium}.
\newblock In \emph{ICML}, 2018.

\bibitem[Polyak(1963)]{polyak1963gradient}
Polyak, B.~T.
\newblock Gradient methods for minimizing functionals.
\newblock \emph{Zhurnal Vychislitel'noi Matematiki i Matematicheskoi Fiziki},
  3\penalty0 (4):\penalty0 643--653, 1963.

\bibitem[Su \& Vaidya(2016)Su and Vaidya]{Su2016FaultTolerantMO}
Su, L. and Vaidya, N.~H.
\newblock {Fault-Tolerant Multi-Agent Optimization: Optimal Iterative
  Distributed Algorithms}.
\newblock In \emph{PODC}, 2016.

\bibitem[Su \& Vaidya(2018)Su and Vaidya]{su2016defending}
Su, L. and Vaidya, N.~H.
\newblock {Defending non-Bayesian learning against adversarial attacks}.
\newblock \emph{Distributed Computing}, pp.\  1--13, 2018.

\bibitem[Xie et~al.(2018)Xie, Koyejo, and Gupta]{xie2018phocas}
Xie, C., Koyejo, O., and Gupta, I.
\newblock {Phocas: Dimensional Byzantine-resilient Stochastic Gradient
  Descent}.
\newblock \emph{arXiv preprint arXiv:1805.09682}, 2018.

\bibitem[Xie et~al.(2019{\natexlab{a}})Xie, Koyejo, and Gupta]{xie2019slsgd}
Xie, C., Koyejo, O., and Gupta, I.
\newblock {SLSGD: Secure and Efficient Distributed On-device Machine Learning}.
\newblock In \emph{Joint European Conference on Machine Learning and Knowledge
  Discovery in Databases}, pp.\  213--228. Springer, 2019{\natexlab{a}}.

\bibitem[Xie et~al.(2019{\natexlab{b}})Xie, Koyejo, and Gupta]{Xie2018ZenoBS}
Xie, C., Koyejo, S., and Gupta, I.
\newblock {Zeno: Distributed Stochastic Gradient Descent with Suspicion-based
  Fault-tolerance}.
\newblock In \emph{ICML}, 2019{\natexlab{b}}.

\bibitem[Xie et~al.(2019{\natexlab{c}})Xie, Koyejo, and Gupta]{Xie2019FallOE}
Xie, C., Koyejo, S., and Gupta, I.
\newblock {Fall of Empires: Breaking Byzantine-tolerant SGD by Inner Product
  Manipulation}.
\newblock In \emph{UAI}, 2019{\natexlab{c}}.

\bibitem[Yin et~al.(2018)Yin, Chen, Ramchandran, and
  Bartlett]{yin2018byzantine}
Yin, D., Chen, Y., Ramchandran, K., and Bartlett, P.
\newblock {Byzantine-Robust Distributed Learning: Towards Optimal Statistical
  Rates}.
\newblock In \emph{ICML}, 2018.

\bibitem[Zheng et~al.(2017)Zheng, Meng, Wang, Chen, Yu, Ma, and
  Liu]{zheng2017asynchronous}
Zheng, S., Meng, Q., Wang, T., Chen, W., Yu, N., Ma, Z., and Liu, T.-Y.
\newblock {Asynchronous Stochastic Gradient Descent with Delay Compensation}.
\newblock In \emph{ICML}, 2017.

\bibitem[Zhou et~al.(2018)Zhou, Mertikopoulos, Bambos, Glynn, Ye, Li, and
  Fei-Fei]{Zhou2018DistributedAO}
Zhou, Z., Mertikopoulos, P., Bambos, N., Glynn, P.~W., Ye, Y., Li, L.-J., and
  Fei-Fei, L.
\newblock {Distributed Asynchronous Optimization with Unbounded Delays: How
  Slow Can You Go?}
\newblock In \emph{ICML}, 2018.

\bibitem[Zinkevich et~al.(2009)Zinkevich, Langford, and
  Smola]{zinkevich2009slow}
Zinkevich, M., Langford, J., and Smola, A.~J.
\newblock {Slow Learners are Fast}.
\newblock In \emph{NeurIPS}, 2009.

\end{thebibliography}
\bibliographystyle{icml2020}

\appendix
\newpage
\onecolumn

\vspace*{0.1cm}
\begin{center}
	\Large\textbf{Appendix}
\end{center}
\vspace*{0.1cm}

\setcounter{theorem}{0}

\section{Proofs}

\subsection{Zeno++}

We first analyze the convergence of the functions whose gradients grow as a quadratic function of sub-optimality.

\begin{theorem}
Assume that $F(x)$ and $f_s(x)$ have $L$-smoothness and PL inequality (potentially non-convex). 
Assume that $\forall x$, the true gradients and stochastic gradients are upper-bounded: $\| \nabla F(x) \|^2 \leq V_1$, $\| \nabla f_s(x) \|^2 \leq V_1$, and the stochastic gradients for Zeno testing are always non-zero and lower-bounded: $\| \nabla f_s(x) \|^2 \geq V_2$, where $0 < V_2 \leq V_1$. Furthermore, we assume that the validation set is close to the training set, which implies bounded variance: $\E\left[ \|\nabla f_s(x) - \nabla F(x)\|^2 \right] \leq V_3, \forall x$. Taking $\gamma < \min(1, \frac{1}{L})$ and $\rho \geq \frac{\alpha \sqrt{\gamma} V_1}{2\mu V_2}$, after $T$ global updates, Algorithm~\ref{alg:zeno_plusplus} has the error bound:
\begin{align*}
\E\left[ F(x_T) - F(x_*) \right] \leq (1 - \alpha \sqrt{\gamma})^T \left[ F(x_{0}) - F(x_*) \right] + \frac{1}{\alpha} \left( \sqrt{\gamma} V_3 + L^2 k^2 \gamma^{5/2} V_1 + \frac{\sqrt{\gamma}}{2} V_1 + \sqrt{\gamma}\epsilon + \frac{L \gamma^{3/2}}{2} V_1 \right).
\end{align*}
\end{theorem}

\begin{proof}
If any gradient estimator $g$ passes the test of \texttt{Zeno++}, then we have
\begin{align*}
\ip{\nabla f_s(x_\tau)}{-\gamma g} \leq -\rho \|g\|^2 + \gamma\epsilon,
\end{align*}
where $\tau \leq t-1$.

Thus, we have
\begin{align*}
&\ip{\nabla F(x_{t-1})}{-\gamma g} \\
&\leq \ip{\nabla F(x_{t-1}) - \nabla f_s(x_\tau)}{-\gamma g} - \rho \| \nabla f_s(x_\tau) \|^2 + \gamma\epsilon \\
&\leq -\rho \frac{V_2}{V_1} \| \nabla F(x_{t-1}) \|^2 + \frac{\gamma}{2} \| \nabla F(x_{t-1}) - \nabla f_s(x_\tau) \|^2 + \frac{\gamma}{2} \| \nabla f_s(x_\tau) \|^2 + \gamma\epsilon \\
&\leq -\rho \frac{V_2}{V_1} \| \nabla F(x_{t-1}) \|^2 + \gamma \| \nabla F(x_\tau) - \nabla f_s(x_\tau) \|^2 + \gamma \| \nabla F(x_{t-1}) - \nabla F(x_\tau) \|^2 + \frac{\gamma}{2} V_1 + \gamma\epsilon \\
&\leq -\rho \frac{V_2}{V_1} \| \nabla F(x_{t-1}) \|^2 + \gamma V_3 + \gamma \| \nabla F(x_{t-1}) - \nabla F(x_\tau) \|^2 + \frac{\gamma}{2} V_1 + \gamma\epsilon.
\end{align*}

$\| \nabla F(x_{t-1}) - \nabla F(x_\tau) \|^2$ can be upper-bounded using $L$-smoothness and the bounded delay:
\begin{align*}
\| \nabla F(x_{t-1}) - \nabla F(x_\tau) \|^2 
\leq L^2\| x_{t-1} - x_\tau \|^2
\leq L^2 k^2 \gamma^2 V_1. 
\end{align*}

Again, using smoothness, taking $\rho \geq \frac{\alpha \sqrt{\gamma} V_1}{2\mu V_2}$, we have
\begin{align*}
&\E\left[ F(x_t) - F(x_*) \right] \\
&\leq F(x_{t-1}) - F(x_*) + \ip{\nabla F(x_{t-1})}{-\gamma \E[g]} + \frac{L \gamma^2}{2} \E\|g\|^2 \\
&\leq F(x_{t-1}) - F(x_*) -\rho \frac{V_2}{V_1} \| \nabla F(x_{t-1}) \|^2 + \gamma V_3 + L^2 k^2 \gamma^3 V_1 + \frac{\gamma}{2} V_1 + \gamma\epsilon + \frac{L \gamma^2}{2} V_1 \\
%&\leq F(x_{t-1}) - F(x_*) -\rho \frac{V_2}{V_1} \| \nabla F(x_{t-1}) \|^2 + \gamma \mathcal{O}(k^2 V_1 + V_3 + \epsilon) \\
&\leq \left[ 1- 2\mu \rho \frac{V_2}{V_1} \right] \left[ F(x_{t-1}) - F(x_*) \right] + \gamma V_3 + L^2 k^2 \gamma^3 V_1 + \frac{\gamma}{2} V_1 + \gamma\epsilon + \frac{L \gamma^2}{2} V_1 \\
&\leq (1 - \alpha \sqrt{\gamma}) \left[ F(x_{t-1}) - F(x_*) \right] + \gamma V_3 + L^2 k^2 \gamma^3 V_1 + \frac{\gamma}{2} V_1 + \gamma\epsilon + \frac{L \gamma^2}{2} V_1.
\end{align*}

By telescoping and taking total expectation, after $T$ global updates, we have
\begin{align*}
\E\left[ F(x_T) - F(x_*) \right] \leq (1 - \alpha \sqrt{\gamma})^T \left[ F(x_{0}) - F(x_*) \right] + \frac{1}{\alpha} \left( \sqrt{\gamma} V_3 + L^2 k^2 \gamma^{5/2} V_1 + \frac{\sqrt{\gamma}}{2} V_1 + \sqrt{\gamma}\epsilon + \frac{L \gamma^{3/2}}{2} V_1 \right).
\end{align*}

\end{proof}

For general smooth but non-convex functions, we have the following convergence guarantee.

\begin{theorem}
Assume that $F(x)$ and $f_s(x)$ are $L$-smooth and potentially non-convex.
Assume that $\forall x$, the true gradients and stochastic gradients are upper-bounded: $\| \nabla F(x) \|^2 \leq V_1$, $\| \nabla f_s(x) \|^2 \leq V_1$, and the stochastic gradients for Zeno testing are always non-zero and lower-bounded: $\| \nabla f_s(x) \|^2 \geq V_2$, where $0 < V_2 \leq V_1$. Furthermore, we assume that the validation set is close to the training set, which implies bounded variance: $\E\left[ \|\nabla f_s(x) - \nabla F(x)\|^2 \right] \leq V_3, \forall x$. Taking $\gamma < \min(1, \frac{1}{L})$ and $\rho \geq \frac{\alpha \sqrt{\gamma} V_1}{V_2}$, after $T$ global updates, Algorithm~\ref{alg:zeno_plusplus} has the error bound:
\begin{align*}
\frac{\E\left[ \sum_{t \in [T]} \| \nabla F(x_{t-1}) \|^2 \right]}{T} \leq \frac{\E\left[ F(x_{0}) - F(x_*) \right]}{\alpha \sqrt{\gamma} T} + \frac{1}{\alpha} \left( \sqrt{\gamma} V_3 + L^2 k^2 \gamma^{5/2} V_1 + \frac{\sqrt{\gamma}}{2} V_1 + \sqrt{\gamma}\epsilon + \frac{L \gamma^{3/2}}{2} V_1 \right).
\end{align*}
%Furthermore, if we take $\gamma = \frac{1}{LT}$, then we have
%\begin{align*}
%\frac{\E\left[ \sum_{t \in [T]} \| \nabla F(x_{t-1}) \|^2 \right]}{T} \leq  \mathcal{O}\left( \frac{1}{\alpha \sqrt{T}} \right).
%\end{align*}
\end{theorem}

\begin{proof}
Similar to Theorem~\ref{thm:pl_zeno++}, we have
\begin{align*}
\ip{\nabla F(x_{t-1})}{-\gamma g} \leq -\rho \frac{V_2}{V_1} \| \nabla F(x_{t-1}) \|^2 + \gamma V_3 + L^2 k^2 \gamma^3 V_1 + \frac{\gamma}{2} V_1 + \gamma\epsilon + \frac{L \gamma^2}{2} V_1.
\end{align*}

Using smoothness, taking $\rho \geq \frac{\alpha \sqrt{\gamma} V_1}{V_2}$, we have
\begin{align*}
&\E\left[ F(x_t) \right] \\
&\leq F(x_{t-1})  + \ip{\nabla F(x_{t-1})}{-\gamma \E[g]} + \frac{L \gamma^2}{2} \E\|g\|^2 \\
&\leq F(x_{t-1}) -\rho \frac{V_2}{V_1} \| \nabla F(x_{t-1}) \|^2 + \gamma V_3 + L^2 k^2 \gamma^3 V_1 + \frac{\gamma}{2} V_1 + \gamma\epsilon + \frac{L \gamma^2}{2} V_1 \\
&\leq F(x_{t-1}) -\alpha \sqrt{\gamma} \| \nabla F(x_{t-1}) \|^2 + \gamma V_3 + L^2 k^2 \gamma^3 V_1 + \frac{\gamma}{2} V_1 + \gamma\epsilon + \frac{L \gamma^2}{2} V_1.
\end{align*}

Thus, we have
\begin{align*}
\| \nabla F(x_{t-1}) \|^2 \leq \frac{\E\left[ F(x_{t-1}) - F(x_t) \right]}{\alpha \sqrt{\gamma}} + \frac{1}{\alpha} \left( \sqrt{\gamma} V_3 + L^2 k^2 \gamma^{5/2} V_1 + \frac{\sqrt{\gamma}}{2} V_1 + \sqrt{\gamma}\epsilon + \frac{L \gamma^{3/2}}{2} V_1 \right).
\end{align*}

By telescoping and taking total expectation, after $T$ global updates, we have
\begin{align*}
\frac{\E\left[ \sum_{t \in [T]} \| \nabla F(x_{t-1}) \|^2 \right]}{T} \leq \frac{\E\left[ F(x_{0}) - F(x_*) \right]}{\alpha \sqrt{\gamma} T} + \frac{1}{\alpha} \left( \sqrt{\gamma} V_3 + L^2 k^2 \gamma^{5/2} V_1 + \frac{\sqrt{\gamma}}{2} V_1 + \sqrt{\gamma}\epsilon + \frac{L \gamma^{3/2}}{2} V_1 \right).
\end{align*}

%Furthermore, if we take $\gamma = \frac{1}{LT}$, then we have
%\begin{align*}
%\frac{\E\left[ \sum_{t \in [T]} \| \nabla F(x_{t-1}) \|^2 \right]}{T} \leq  \mathcal{O}\left( \frac{1}{\alpha \sqrt{T}} \right).
%\end{align*}

\end{proof}

\section{Additional experiments}

\subsection{No attack}

\begin{figure*}[htb!]
\centering
\subfigure[Top-1 Accuracy on Testing Set, max delay is 5]{\includegraphics[width=0.49\textwidth,height=4cm]{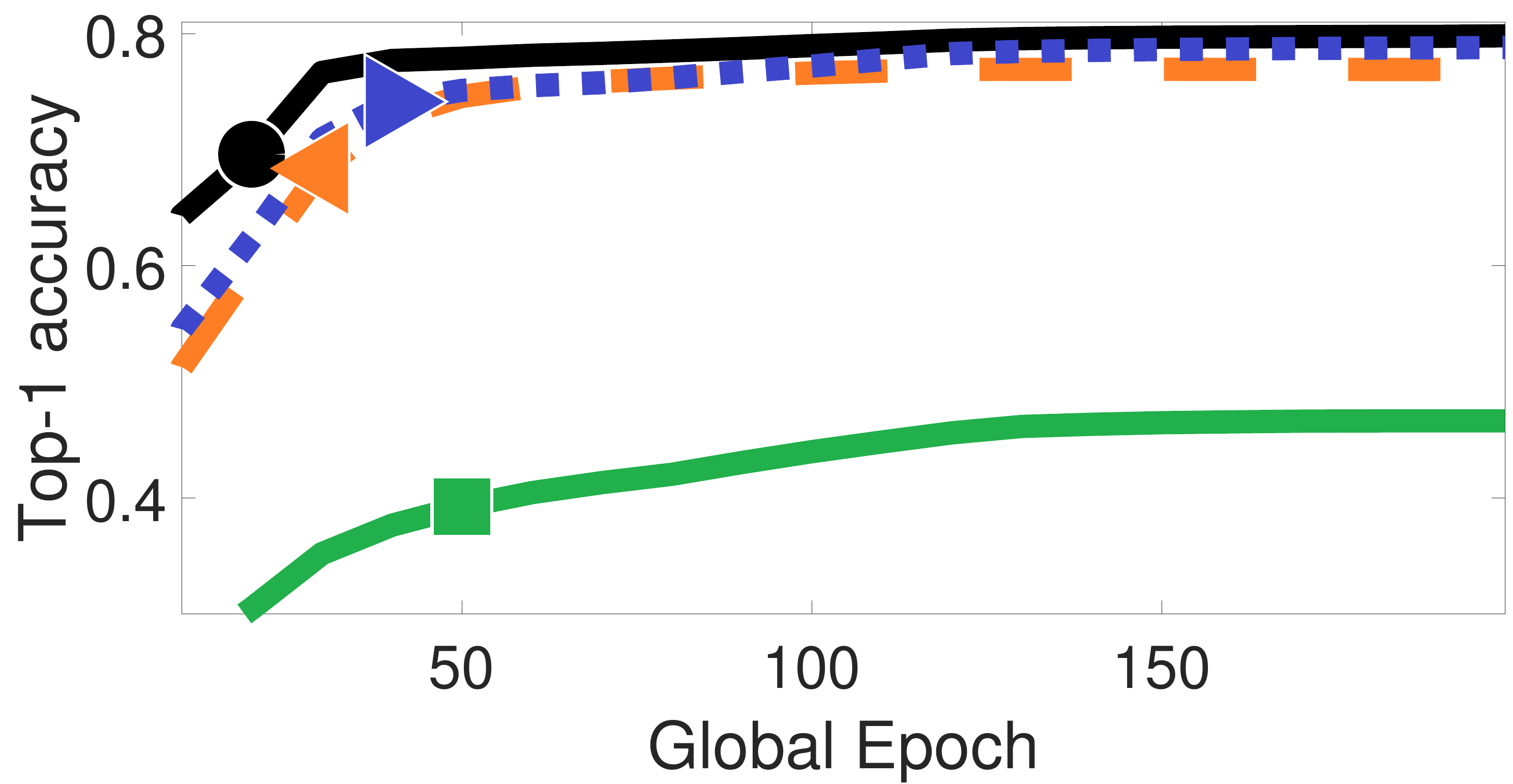}}
\subfigure[Cross Entropy on Training Set, max delay is 5]{\includegraphics[width=0.49\textwidth,height=4cm]{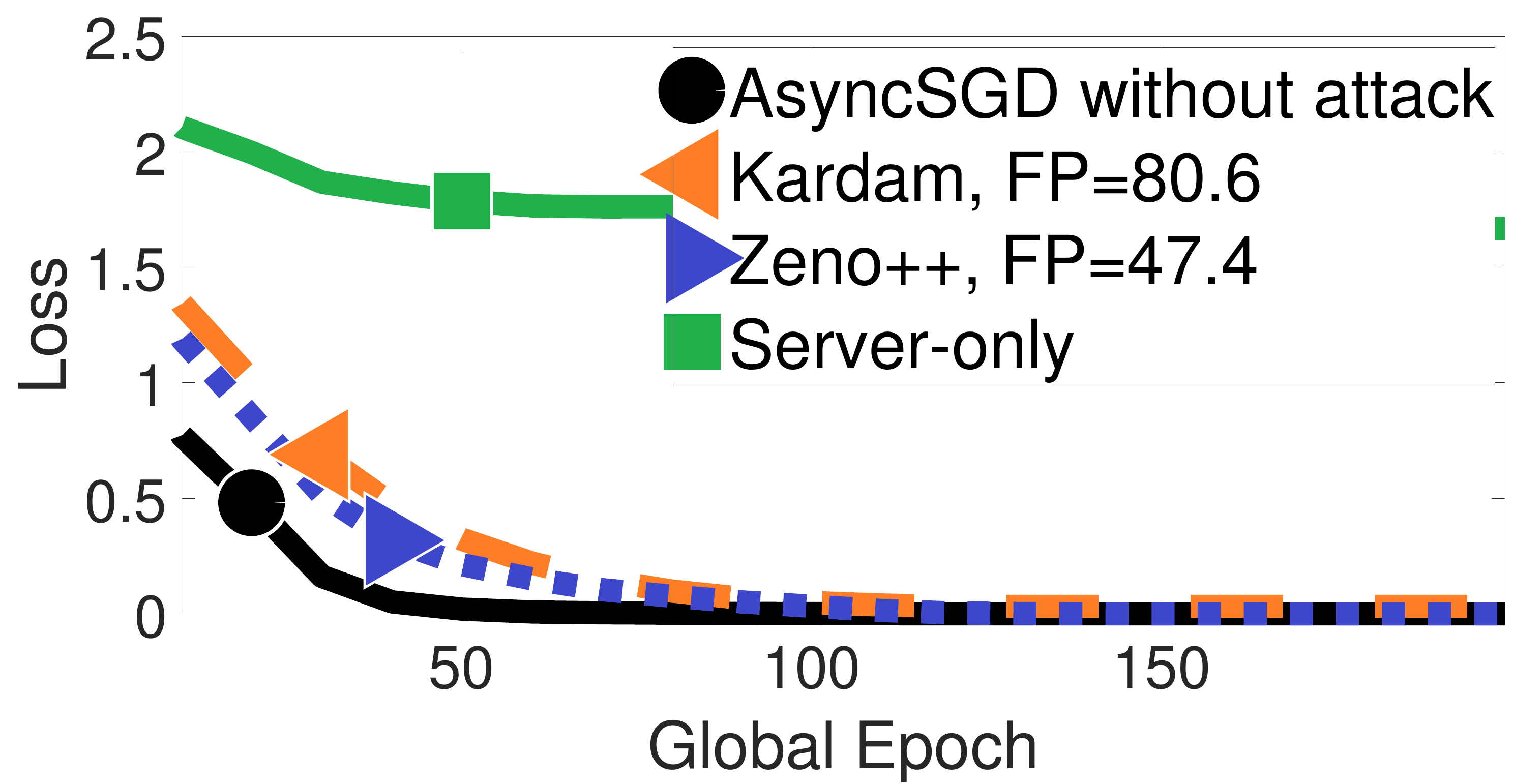}}
\subfigure[Top-1 Accuracy on Testing Set, max delay is 10]{\includegraphics[width=0.49\textwidth,height=4cm]{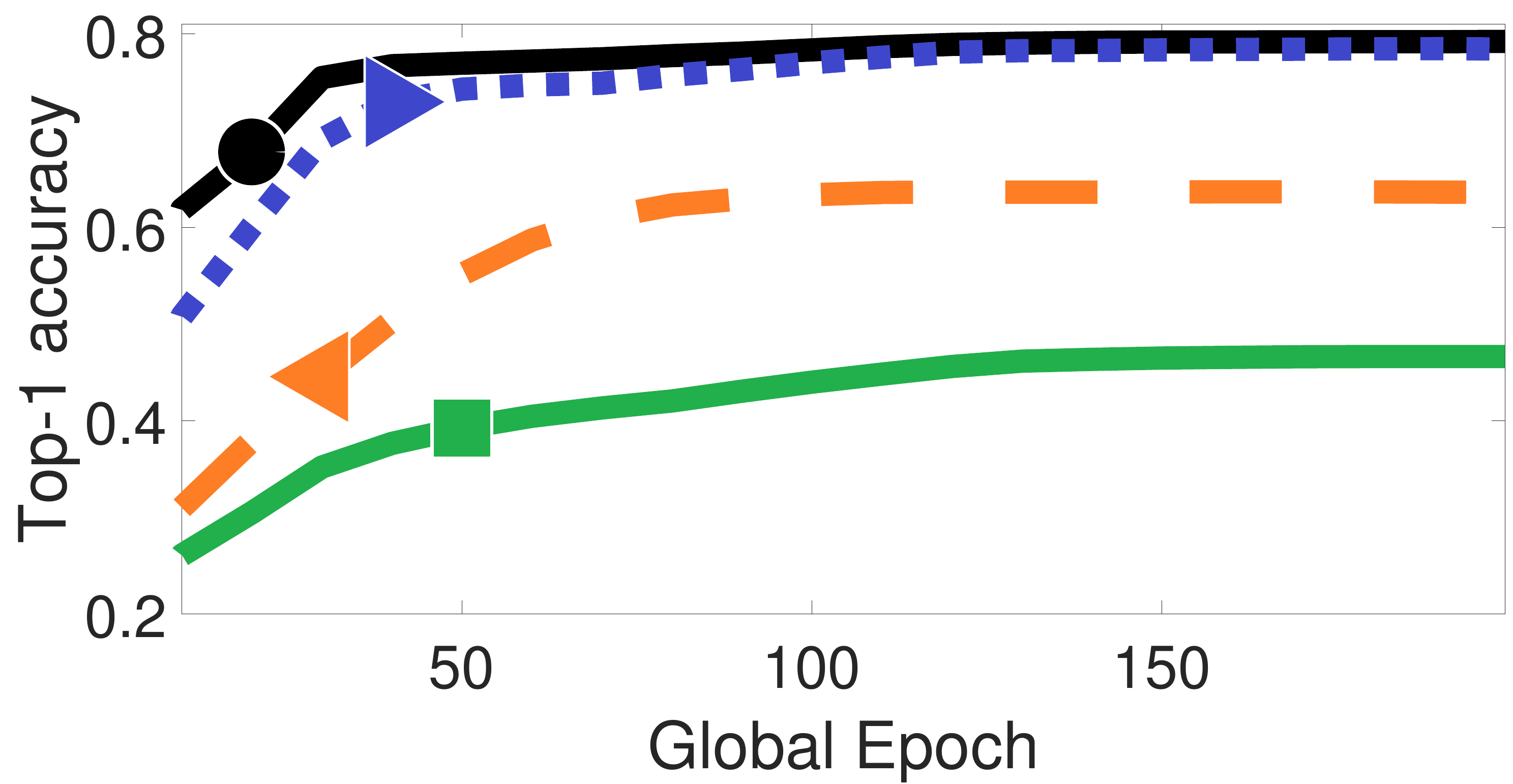}}
\subfigure[Cross Entropy on Training Set, max delay is 10]{\includegraphics[width=0.49\textwidth,height=4cm]{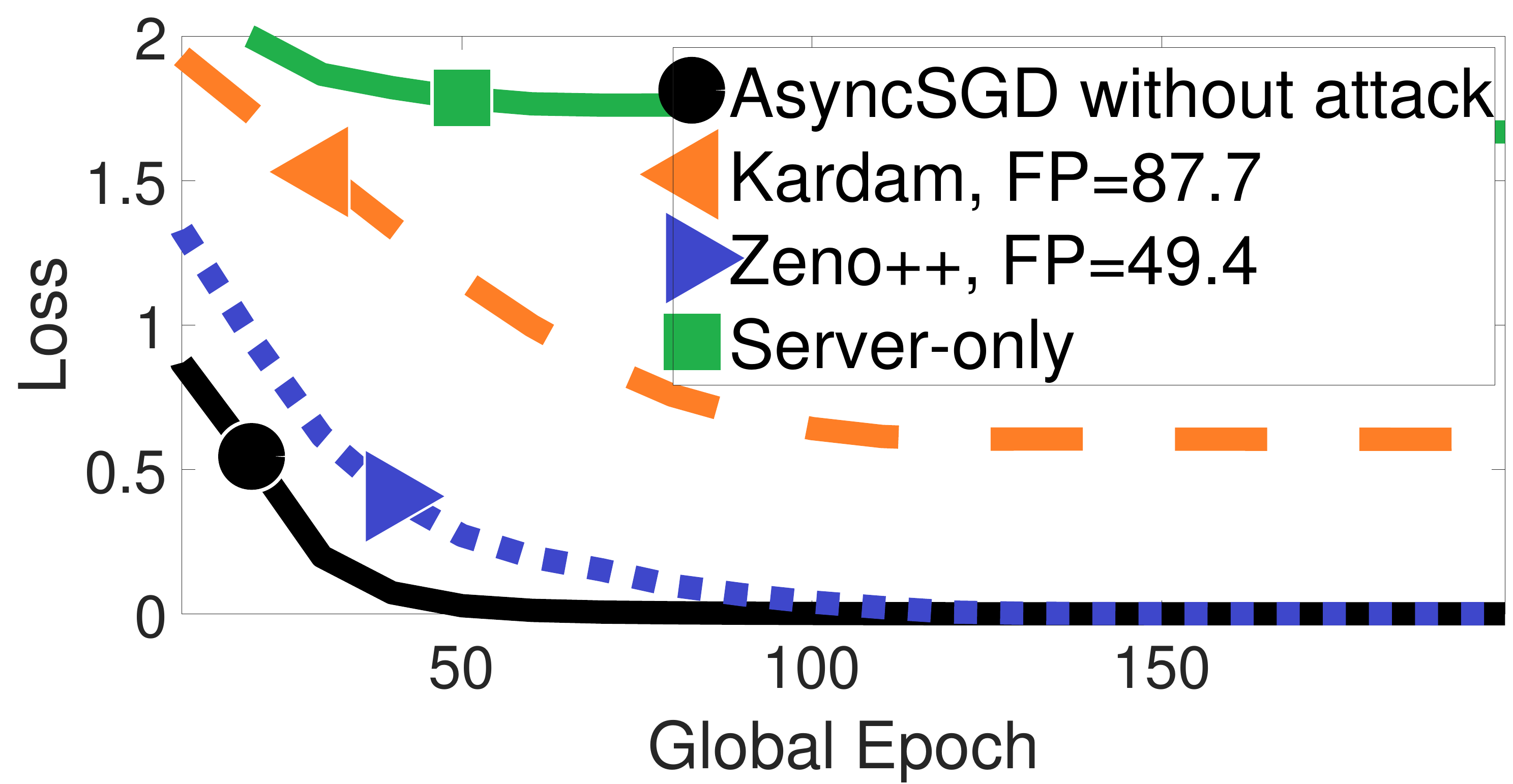}}
\caption{Convergence without attacks, with different maximum delays of workers. $\rho=0.002, \epsilon=0.1$ for \texttt{Zeno++}.}
\label{fig:nobyz}
\end{figure*}

We first test the convergence when there are no attacks. In all the experiments, we take the learning rate $\gamma=0.1$, mini-batch size $n = n_s = 128$, $\rho = 0.002$, $\epsilon=0.1$, $k=10$. For \texttt{Kardam}, we take $q=2$~(\texttt{Kardam} pretends that there are 2 Byzantine workers). The result is shown in Figure~\ref{fig:nobyz}. We can see that \texttt{Zeno++} converges a little bit slower than \texttt{AsyncSGD}, but faster than \texttt{Kardam}, especially when the worker delay is large. When $k_w=10$, \texttt{Zeno++} converges much faster than \text{Kardam}.

\subsection{Sign-flipping attack}

\begin{figure*}[htb!]
\centering
\subfigure[Top-1 Accuracy on Testing Set, $k_w = 5$.]{\includegraphics[width=0.49\textwidth,height=4cm]{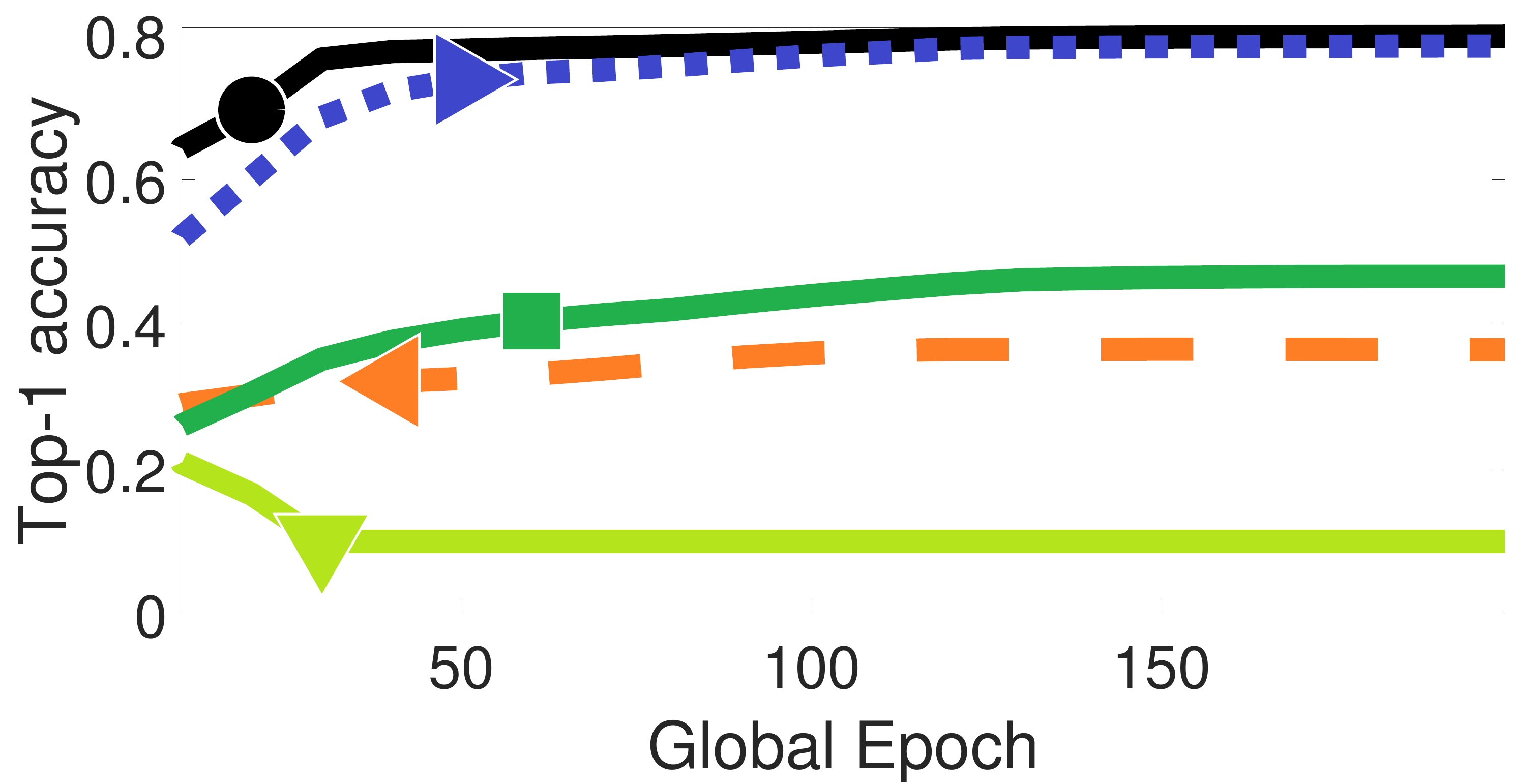}}
\subfigure[Cross Entropy on Training Set, $k_w = 5$.]{\includegraphics[width=0.49\textwidth,height=4cm]{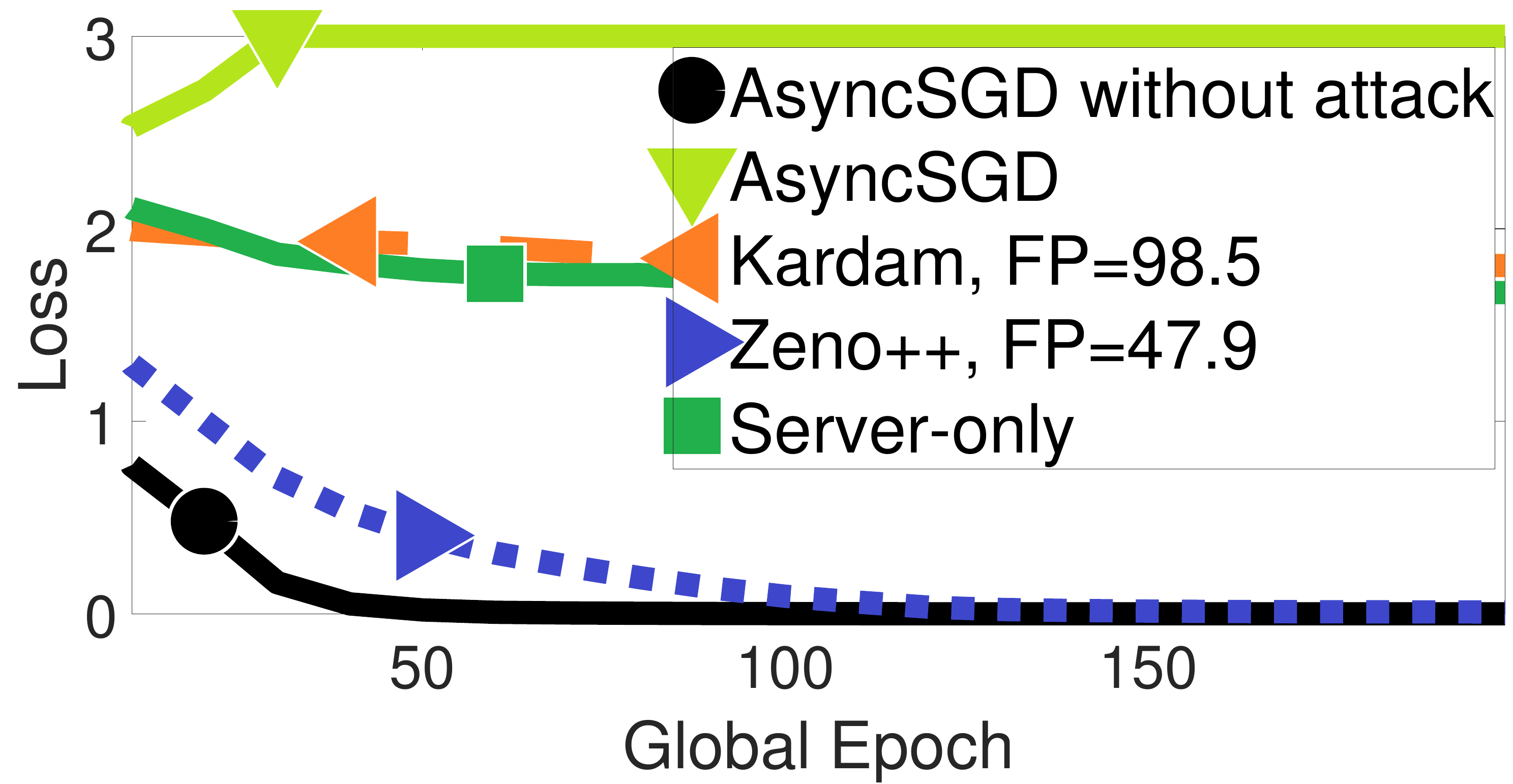}}
\subfigure[Top-1 Accuracy on Testing Set, $k_w = 15$.]{\includegraphics[width=0.49\textwidth,height=4cm]{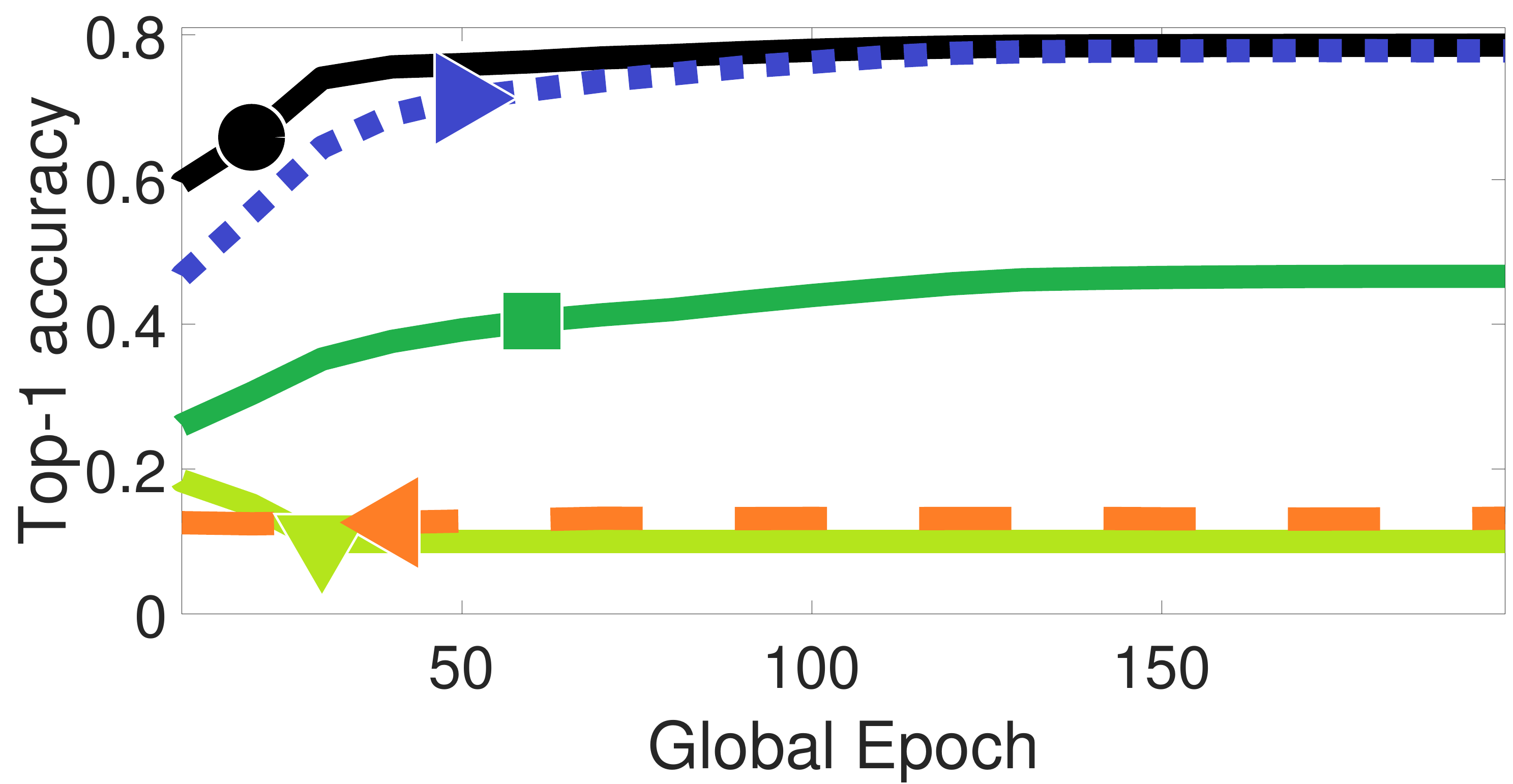}}
\subfigure[Cross Entropy on Training Set, $k_w = 15$.]{\includegraphics[width=0.49\textwidth,height=4cm]{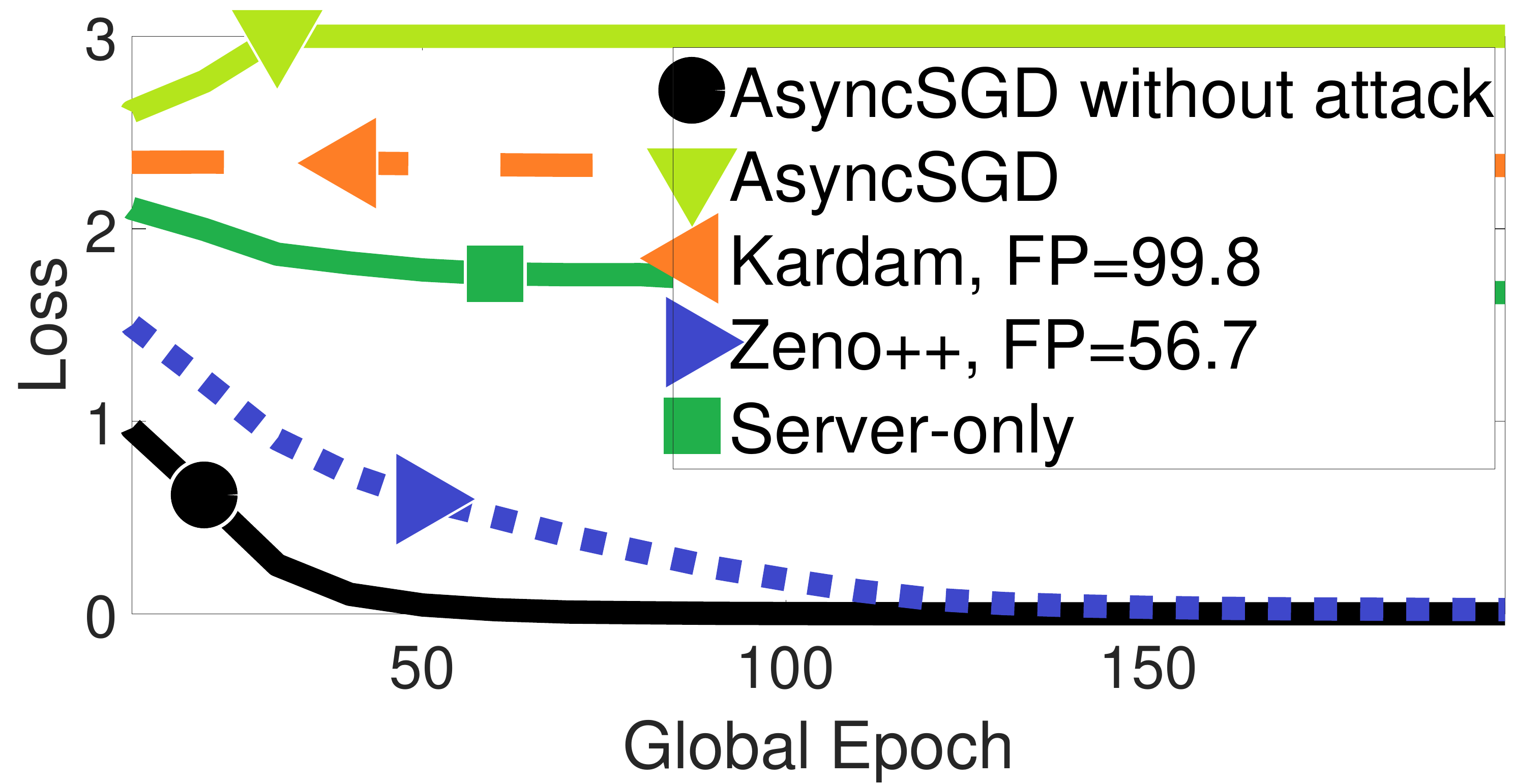}}
\caption{Convergence with sign-flipping attacks, with different maximum worker delays $k_w$. For any correct gradient $g$, if selected to be Byzantine, $g$ will be replaced by $-10g$. $q=4$ out of the 10 workers are Byzantine. $\rho=0.002, \epsilon=0.1, k=10$ for \texttt{Zeno++}.}
\label{fig:signflip_4}
\end{figure*}

\begin{figure*}[htb!]
\centering
\subfigure[Top-1 Accuracy on Testing Set, $k_w = 5$.]{\includegraphics[width=0.49\textwidth,height=4cm]{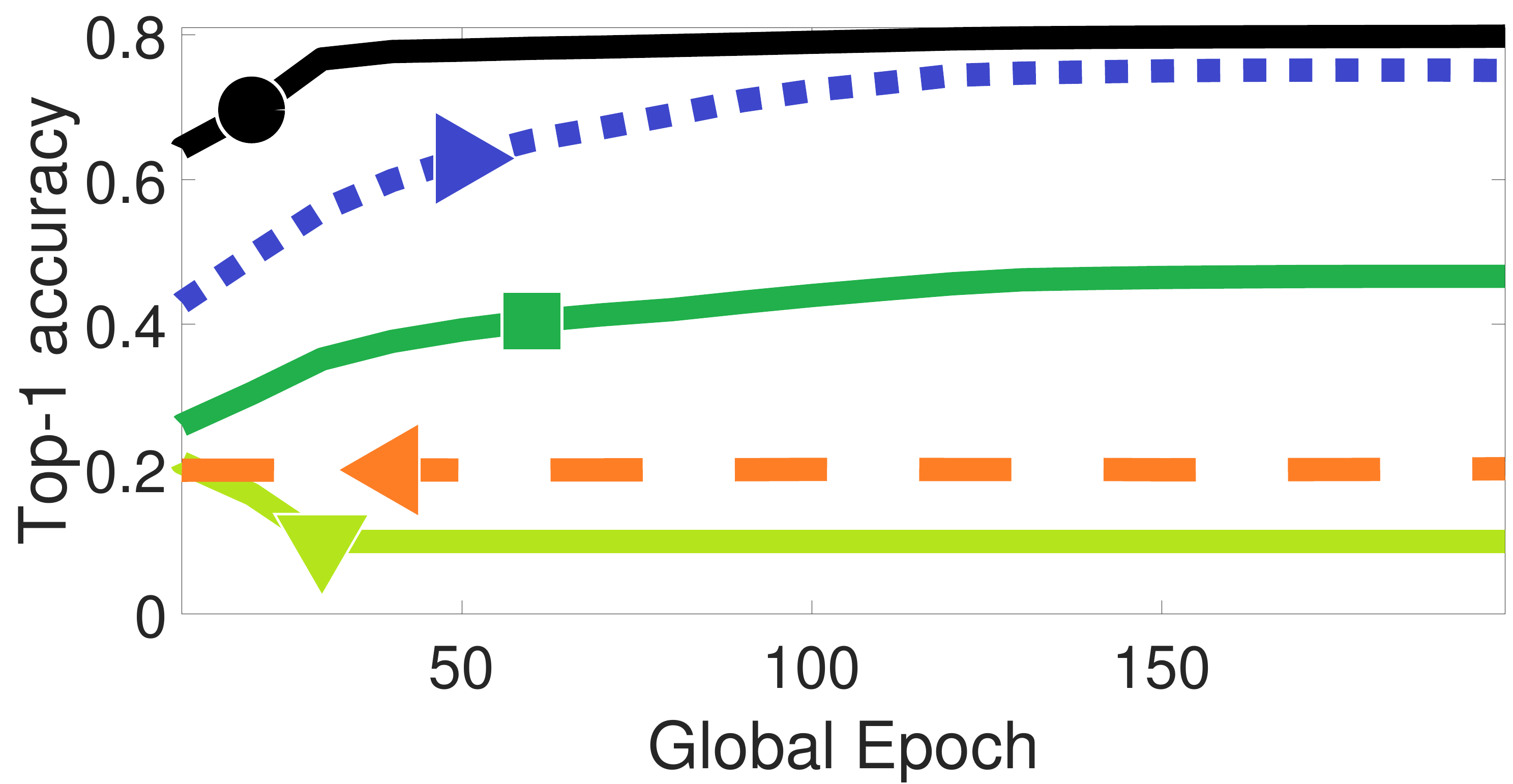}}
\subfigure[Cross Entropy on Training Set, $k_w = 5$.]{\includegraphics[width=0.49\textwidth,height=4cm]{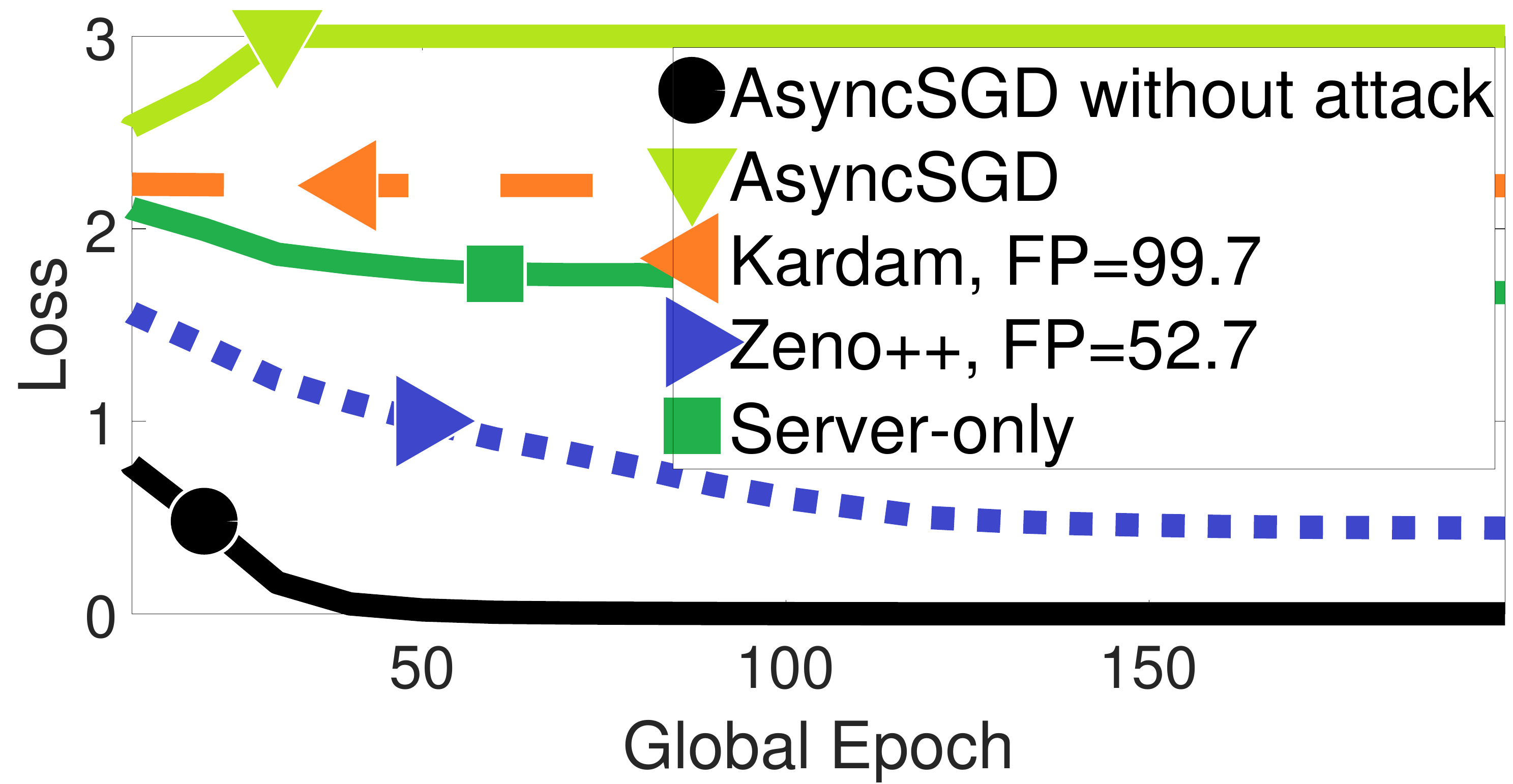}}
\subfigure[Top-1 Accuracy on Testing Set, $k_w = 15$.]{\includegraphics[width=0.49\textwidth,height=4cm]{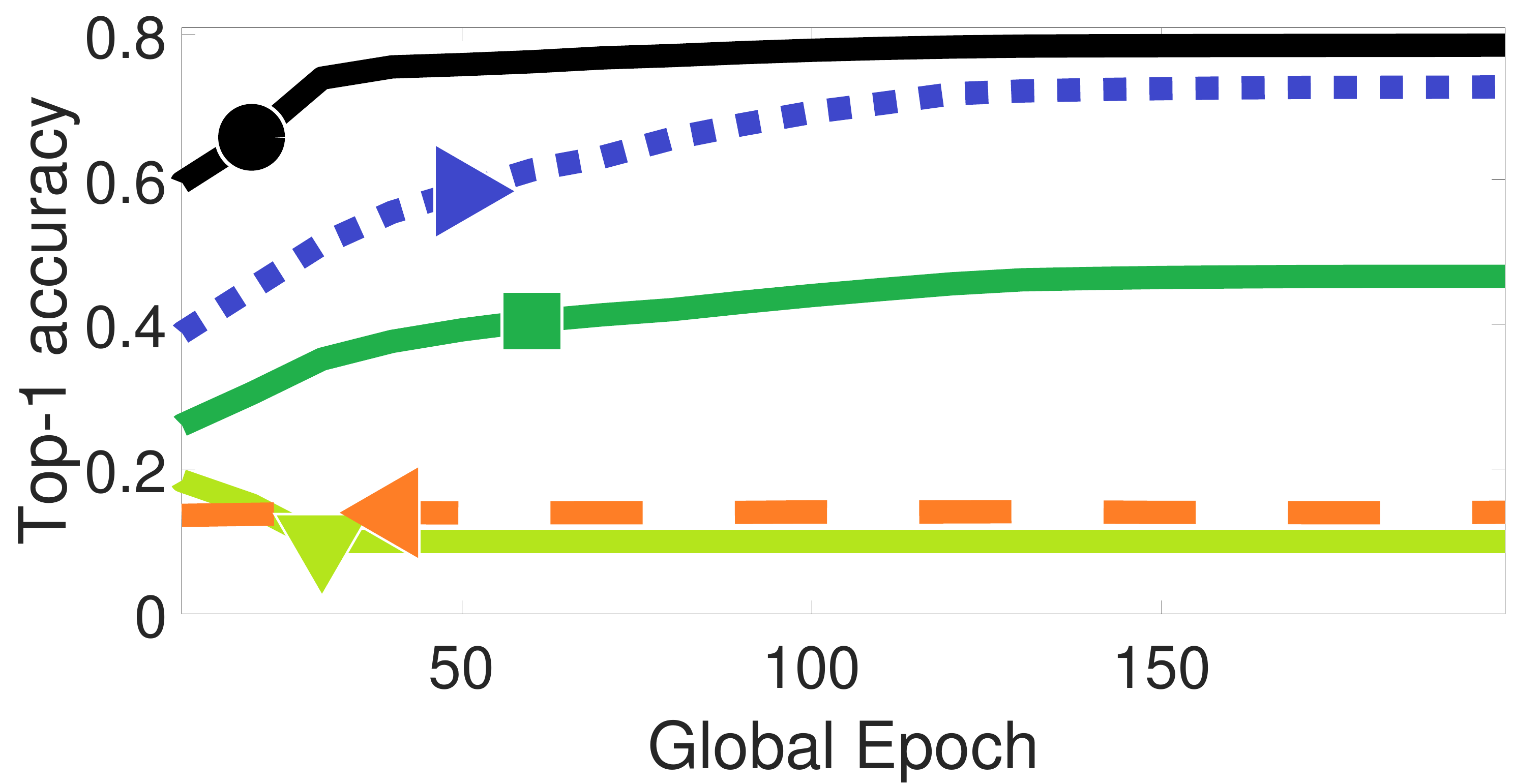}}
\subfigure[Cross Entropy on Training Set, $k_w = 15$.]{\includegraphics[width=0.49\textwidth,height=4cm]{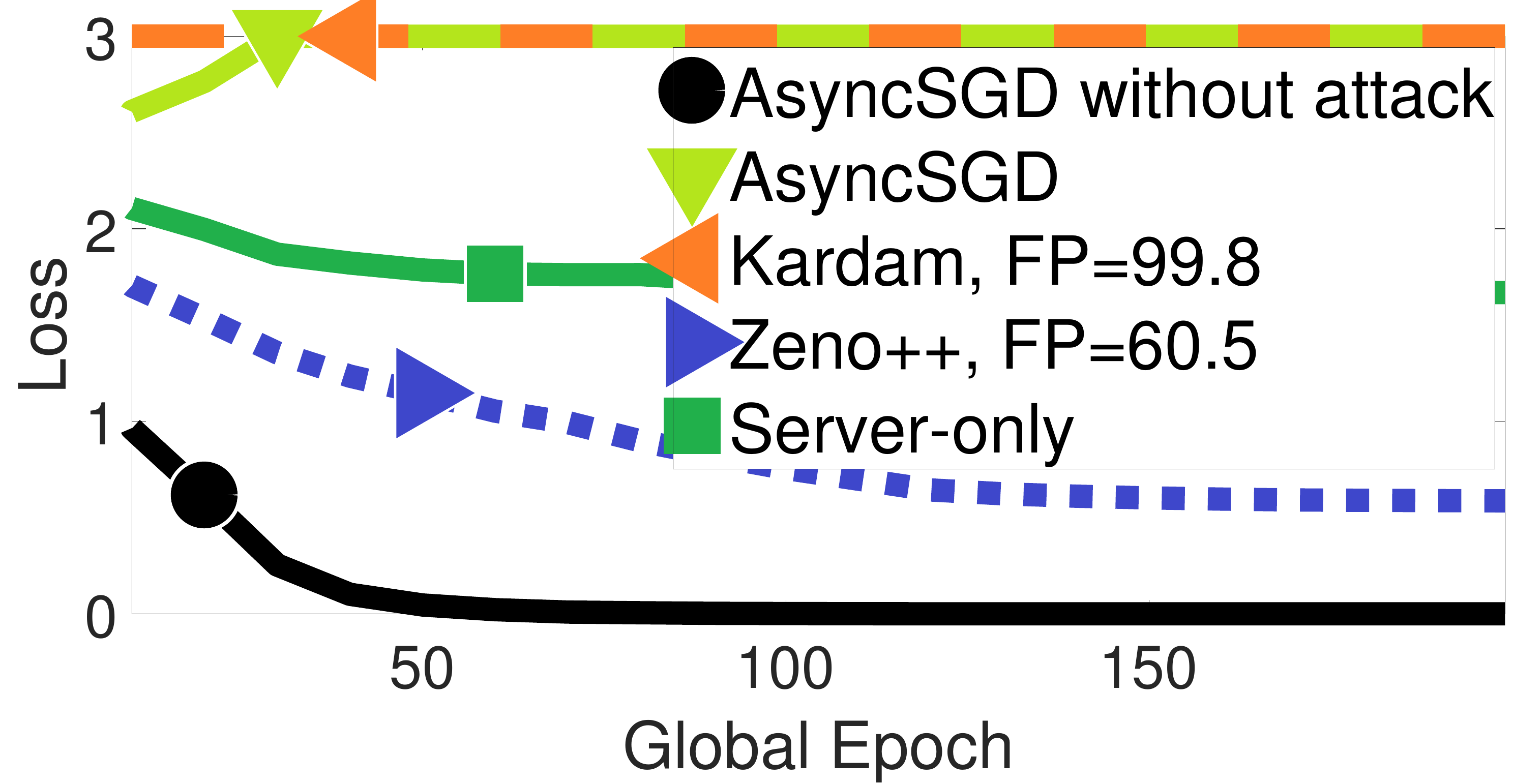}}
\caption{Convergence with sign-flipping attacks, with different maximum worker delays $k_w$. For any correct gradient $g$, if selected to be Byzantine, $g$ will be replaced by $-10g$. $q=8$ out of the 10 workers are Byzantine. $\rho=0.002, \epsilon=0.1, k=10$ for \texttt{Zeno++}.}
\label{fig:signflip_8}
\end{figure*}

We test the Byzantine-tolerance to ``sign-flipping'' attack, which is proposed in \cite{damaskinos2018asynchronous}. In such attacks, the Byzantine workers send $-10 \nabla f(x)$ instead of the correct gradient $\nabla f(x)$ to the server. In all the experiments, we take the learning rate $\gamma=0.1$, mini-batch size $n = n_s = 128$, $\rho = 0.002$, $\epsilon=0.1$, $k=10$. The result is shown in Figure~\ref{fig:signflip_4} and \ref{fig:signflip_8}, with different number of Byzantine workers $q$.

\subsection{Label-flipping attack}

\begin{figure*}[htb!]
\centering
\subfigure[Top-1 Accuracy on Testing Set, $k_w = 5$.]{\includegraphics[width=0.49\textwidth,height=4cm]{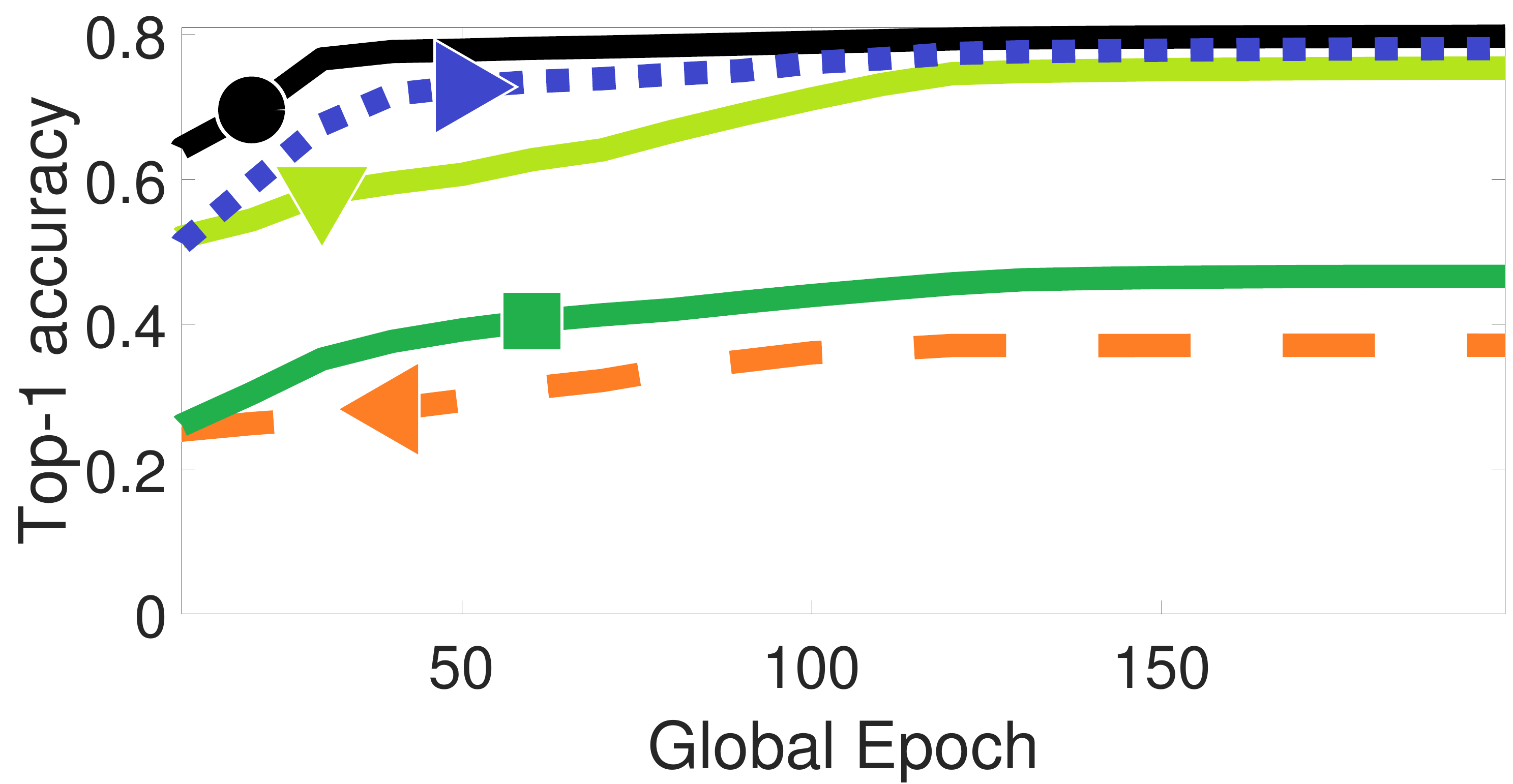}}
\subfigure[Cross Entropy on Training Set, $k_w = 5$.]{\includegraphics[width=0.49\textwidth,height=4cm]{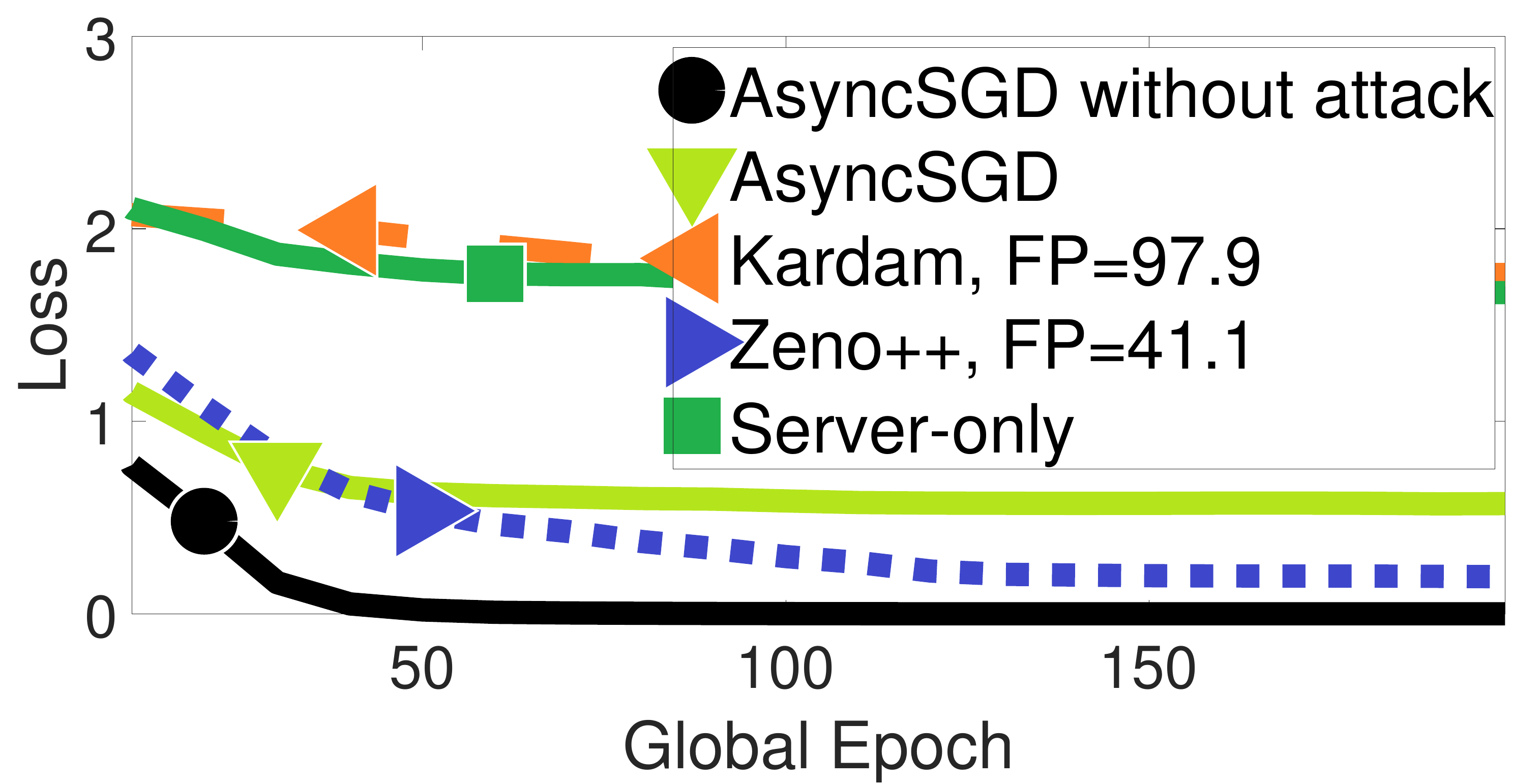}}
\subfigure[Top-1 Accuracy on Testing Set, $k_w = 15$.]{\includegraphics[width=0.49\textwidth,height=4cm]{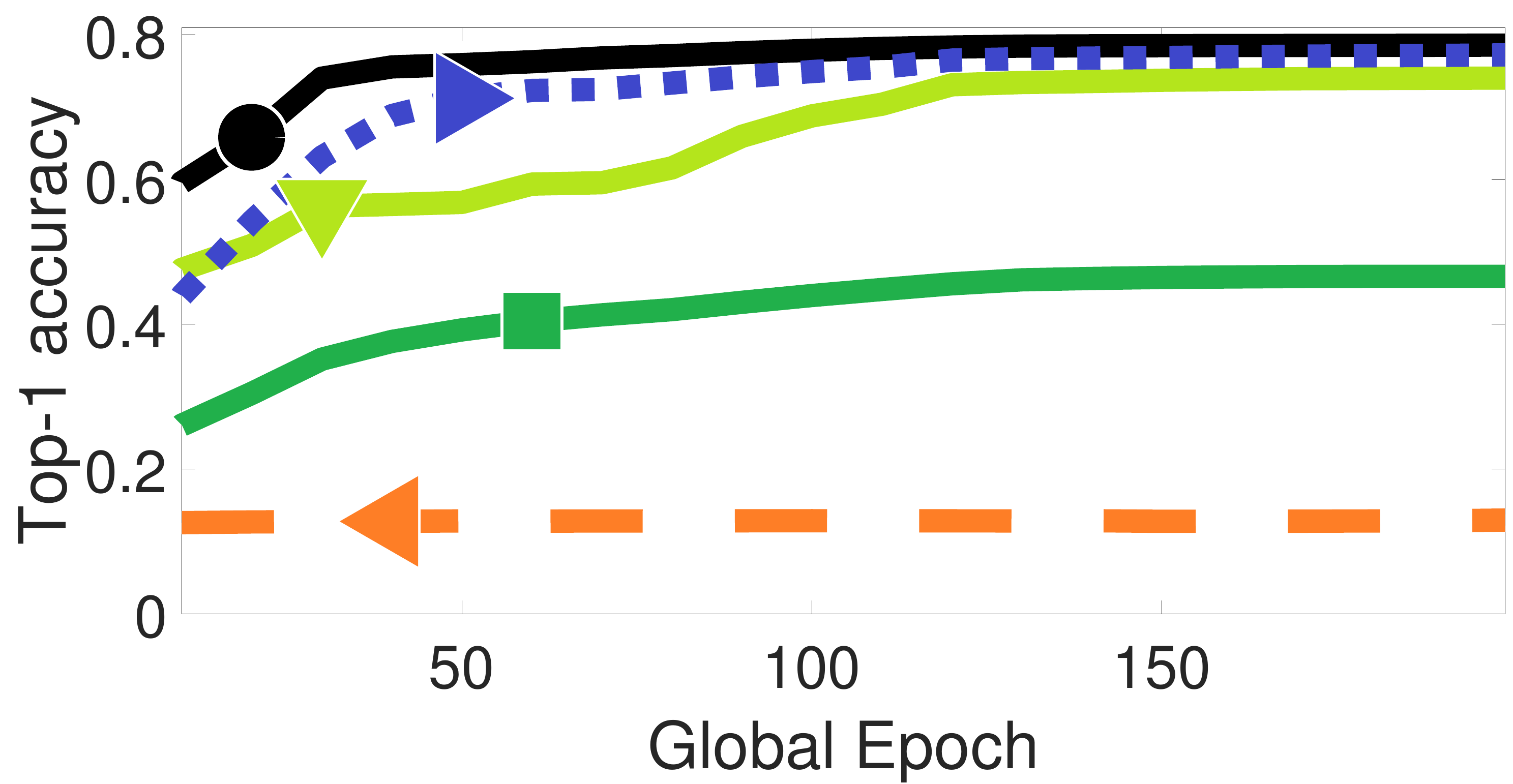}}
\subfigure[Cross Entropy on Training Set, $k_w = 15$.]{\includegraphics[width=0.49\textwidth,height=4cm]{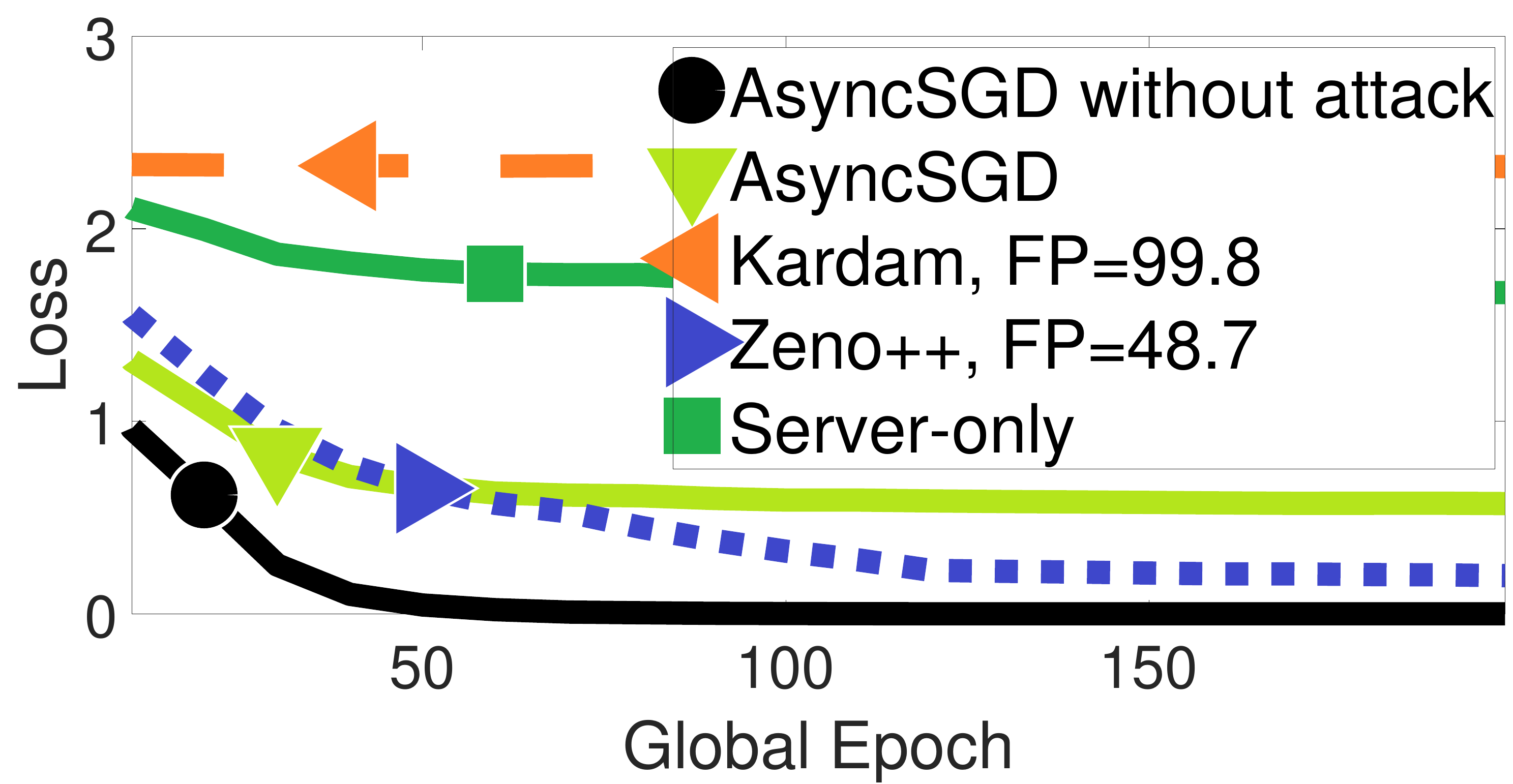}}
\caption{Convergence with label-flipping attacks, with different maximum worker delays $k_w$. $q=4$ out of the 10 workers are Byzantine. $\rho=0.002, \epsilon=0.1, k=10$ for \texttt{Zeno++}.}
\label{fig:labelflip_4}
\end{figure*}

\begin{figure*}[htb!]
\centering
\subfigure[Top-1 Accuracy on Testing Set, $k_w = 5$.]{\includegraphics[width=0.49\textwidth,height=4cm]{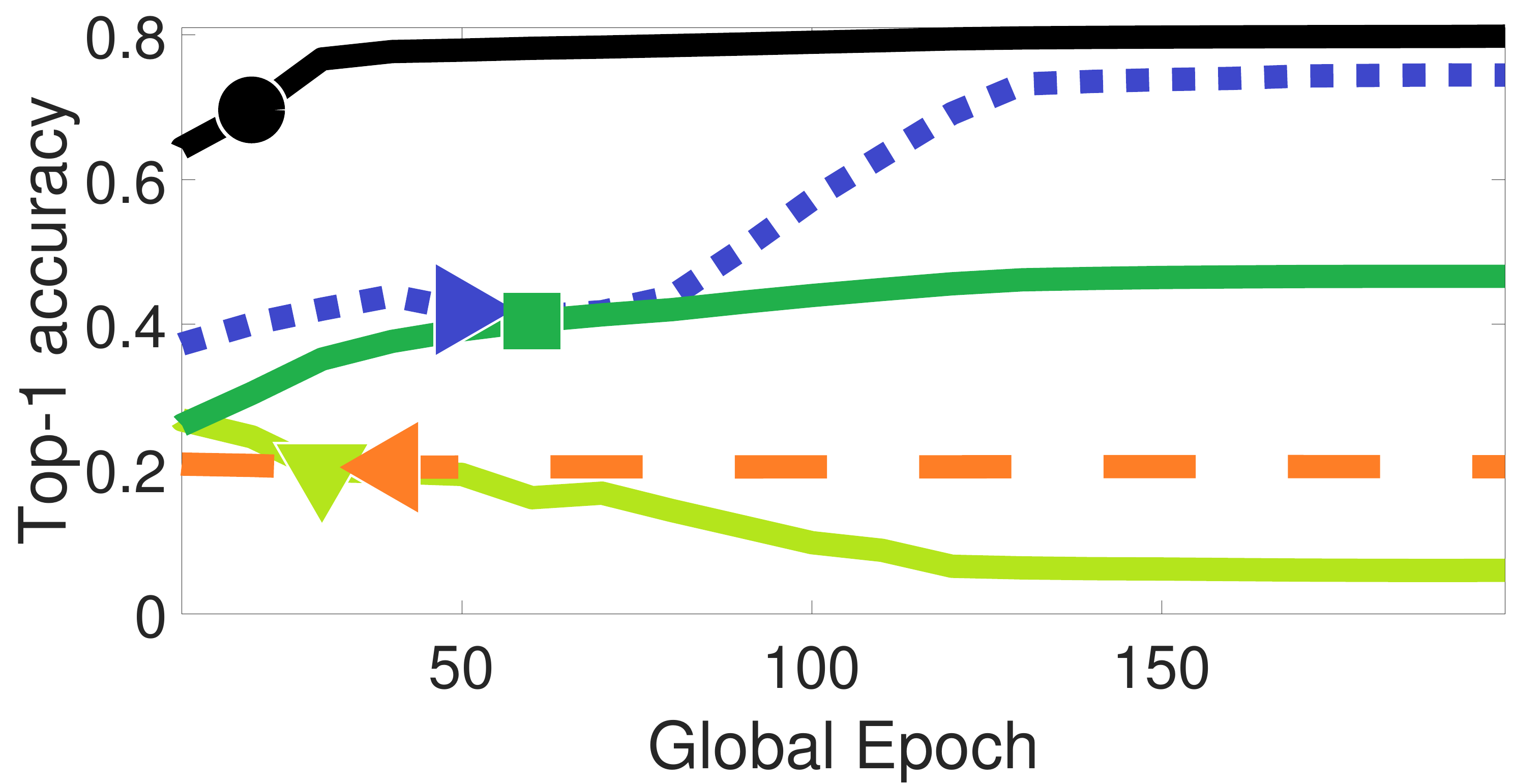}}
\subfigure[Cross Entropy on Training Set, $k_w = 5$.]{\includegraphics[width=0.49\textwidth,height=4cm]{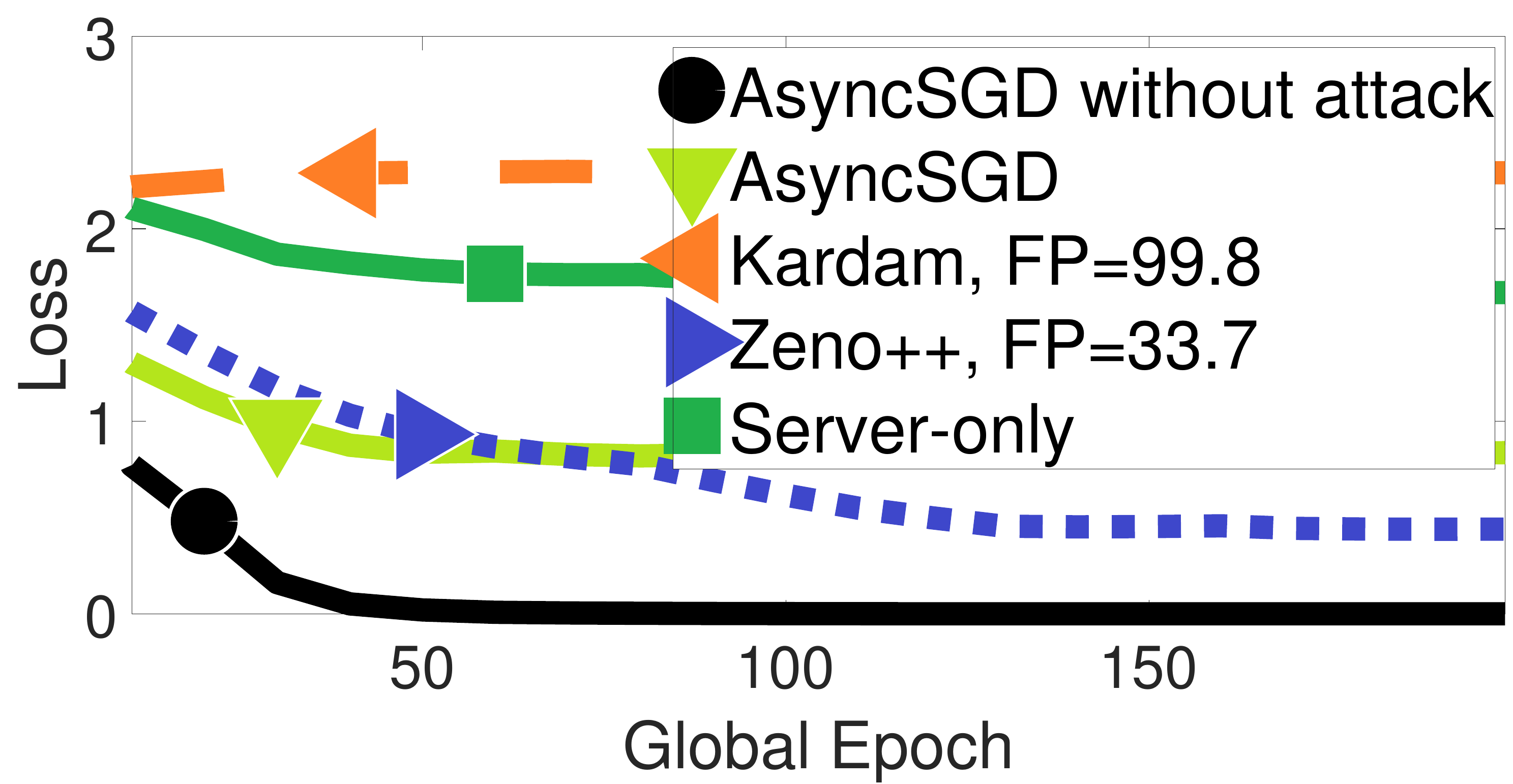}}
\subfigure[Top-1 Accuracy on Testing Set, $k_w = 15$.]{\includegraphics[width=0.49\textwidth,height=4cm]{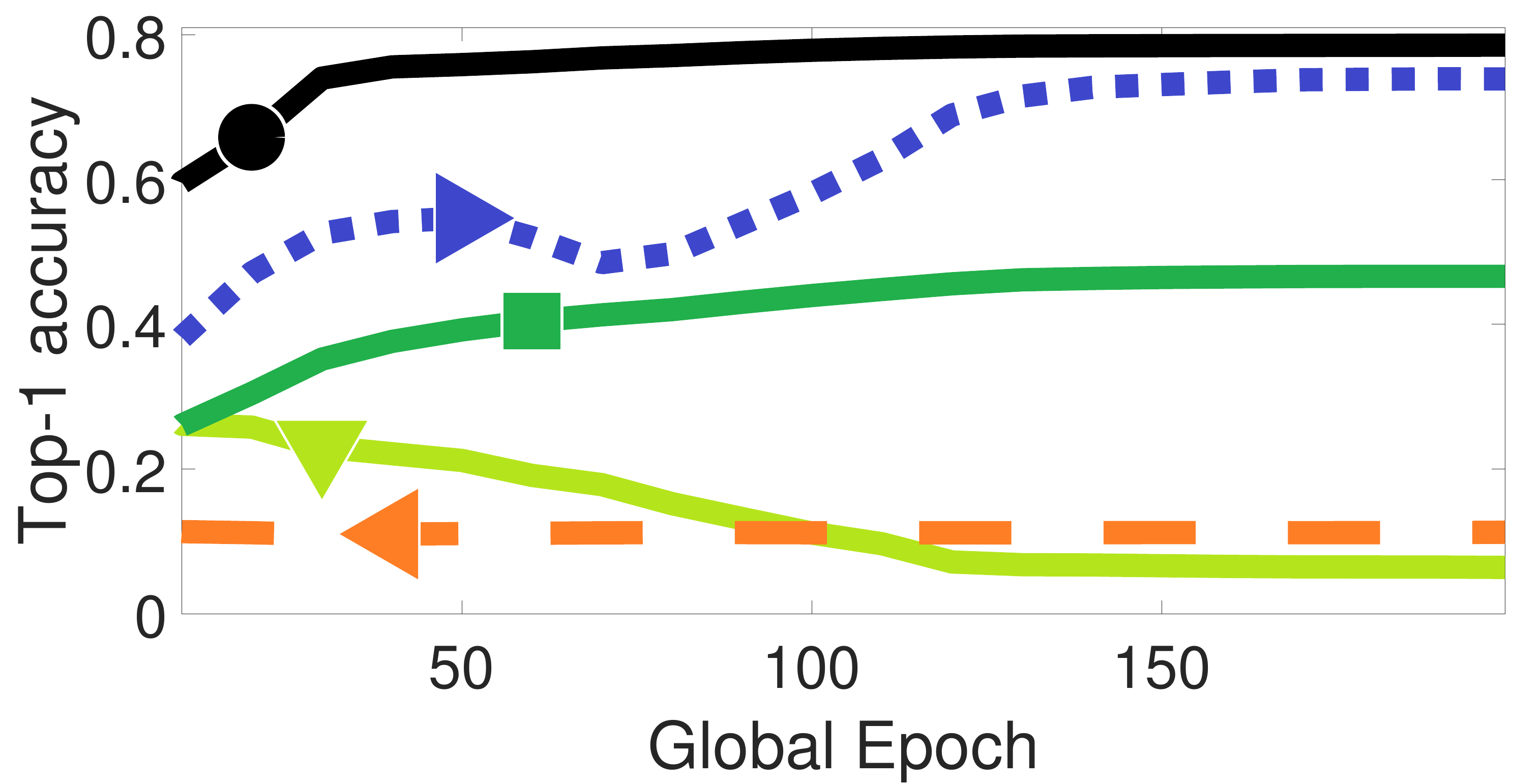}}
\subfigure[Cross Entropy on Training Set, $k_w = 15$.]{\includegraphics[width=0.49\textwidth,height=4cm]{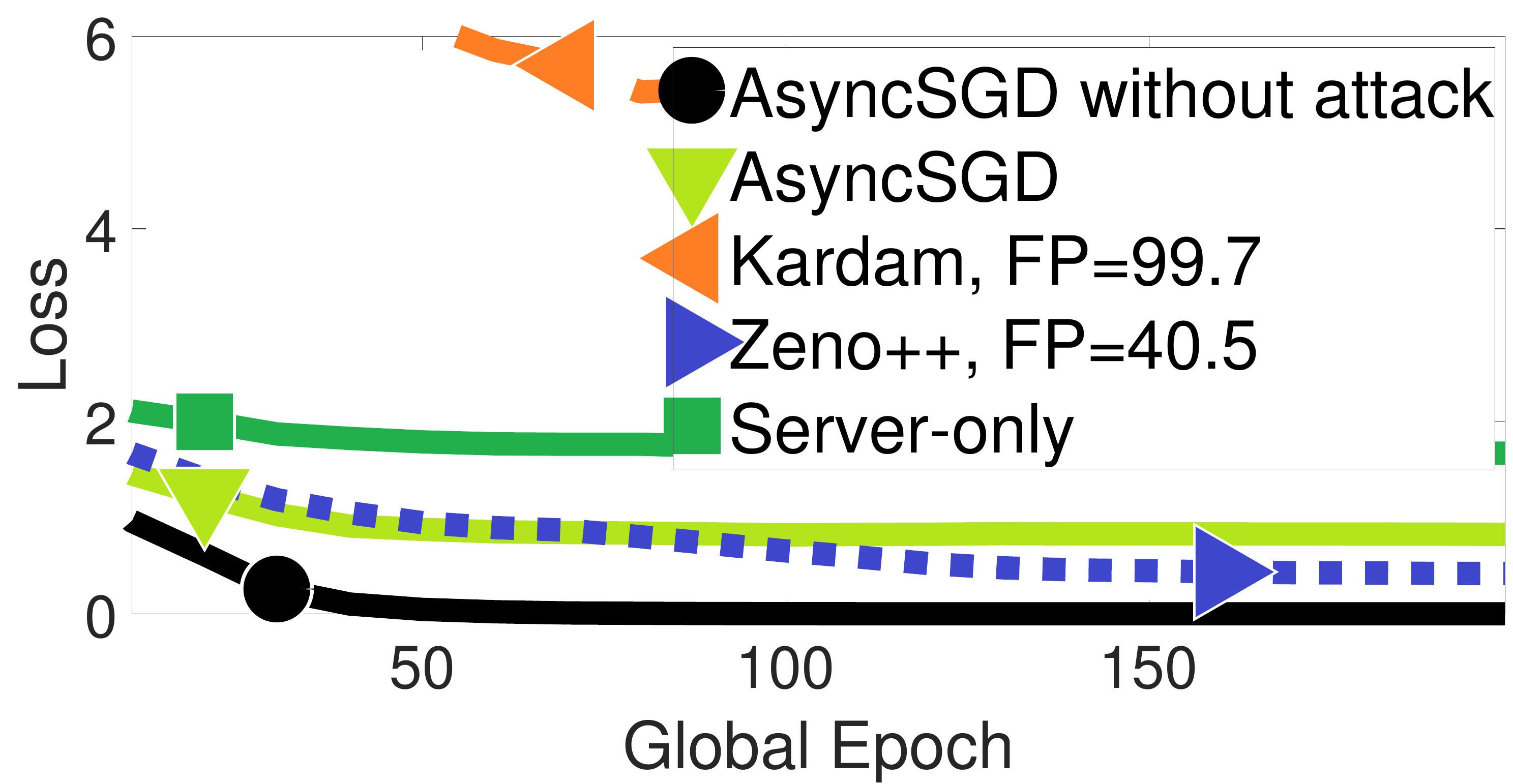}}
\caption{Convergence with label-flipping attacks, with different maximum worker delays $k_w$. $q=6$ out of the 10 workers are Byzantine. $\rho=0.002, \epsilon=0.1, k=10$ for \texttt{Zeno++}.}
\label{fig:labelflip_6}
\end{figure*}

We test the Byzantine tolerance to the label-flipping attacks. When such kind of attacks happen, the workers compute the gradients based on the training data with ``flipped" labels, i.e., any $label \in \{0, \ldots, 9\}$, is replaced by $9 - label$. Such kind of attacks can be caused by data poisoning or software failures. In all the experiments, we take the learning rate $\gamma=0.1$, mini-batch size $n = n_s = 128$, $\rho = 0.002$, $\epsilon=0.1$, $k=10$. The result is shown in Figure~\ref{fig:labelflip_4} and \ref{fig:labelflip_6}, with different number of Byzantine workers $q$.

\subsection{Sensitivity to hyperparameters}

\begin{figure*}[htb!]
\centering
\includegraphics[width=0.9\textwidth]{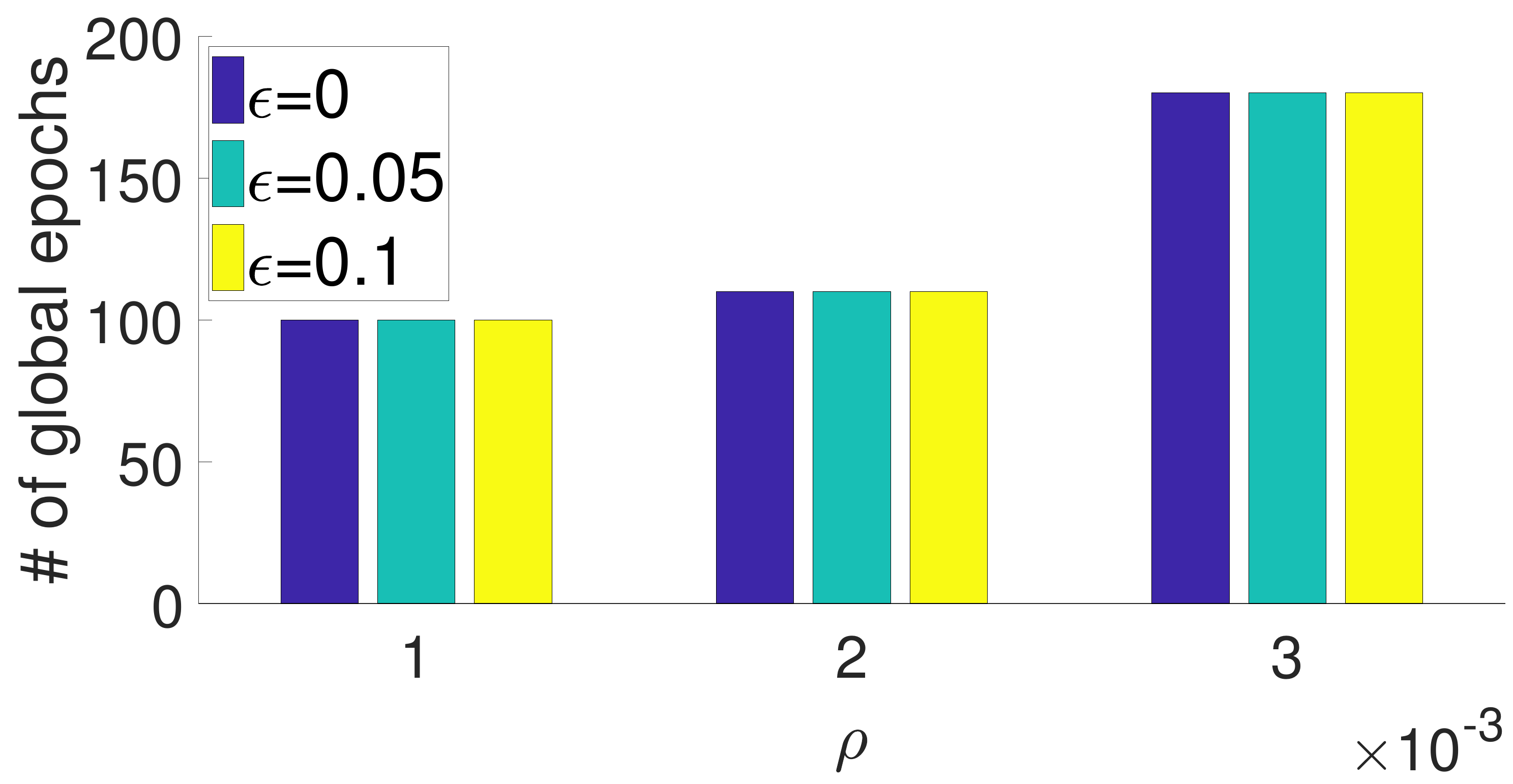}
\caption{Number of global epochs to reach training loss value $0.2$, with sign-flipping attacks and $q=4$ Byzantine workers. $k=10$ for \texttt{Zeno++}. $\rho$ and $\epsilon$ varies.}
\label{fig:rho_epsilon}
\end{figure*}

\begin{figure*}[htb!]
\centering
\includegraphics[width=0.9\textwidth]{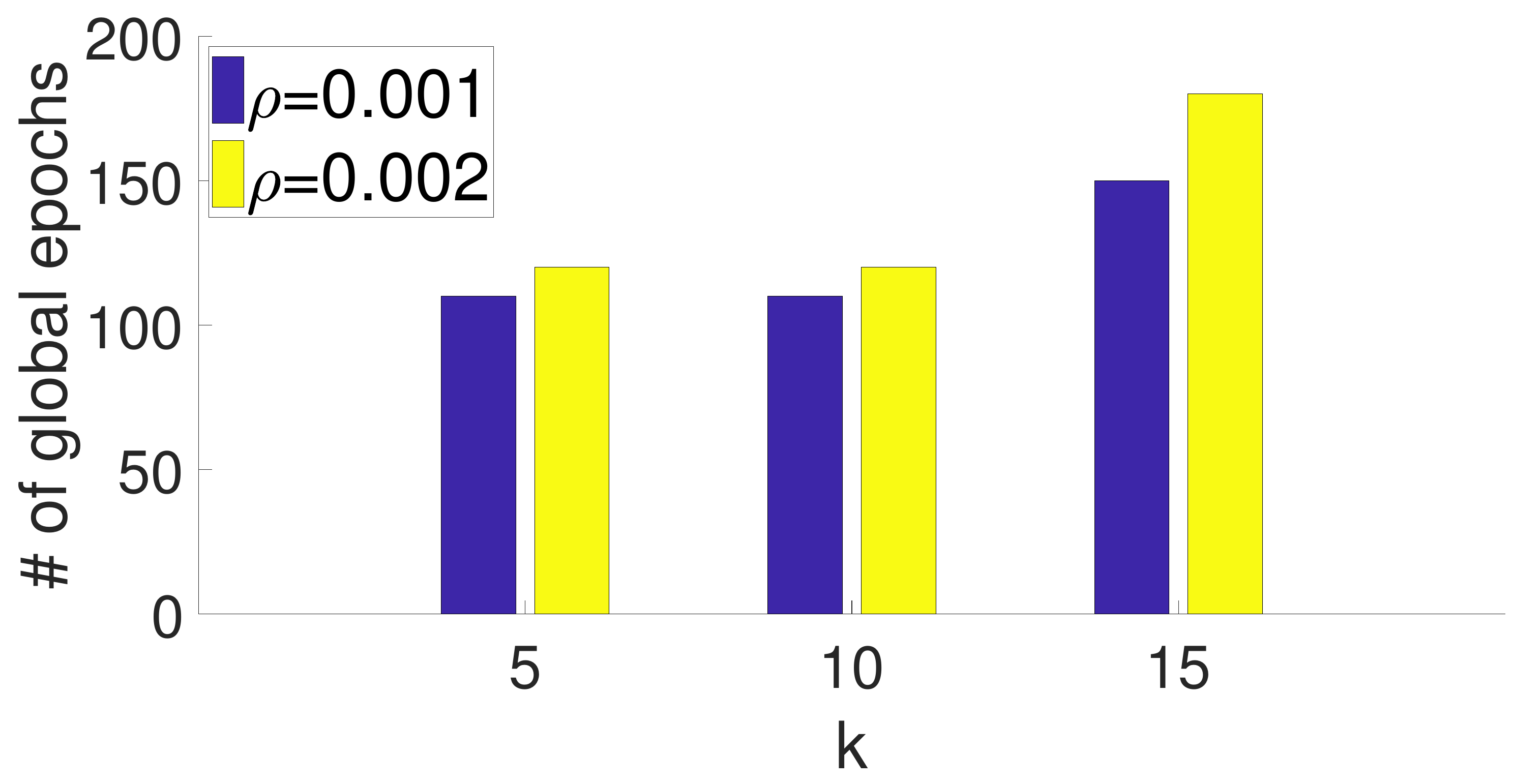}
\caption{Number of global epochs to reach training loss value $0.2$, with sign-flipping attacks and $q=6$ Byzantine workers. $\epsilon=0$ for \texttt{Zeno++}. $\rho$ and $k$ varies.}
\label{fig:zeno_delay}
\end{figure*}

In Figure~\ref{fig:rho_epsilon} and \ref{fig:zeno_delay}, we show how the hyperparameters $\rho$, $\epsilon$, and $k$ affect the convergence. In general, \texttt{Zeno++} is insensitive to $\epsilon$. Larger $\rho$ and $k$ slow down the convergence.

\begin{figure*}[htb!]
\centering
\includegraphics[width=0.9\textwidth]{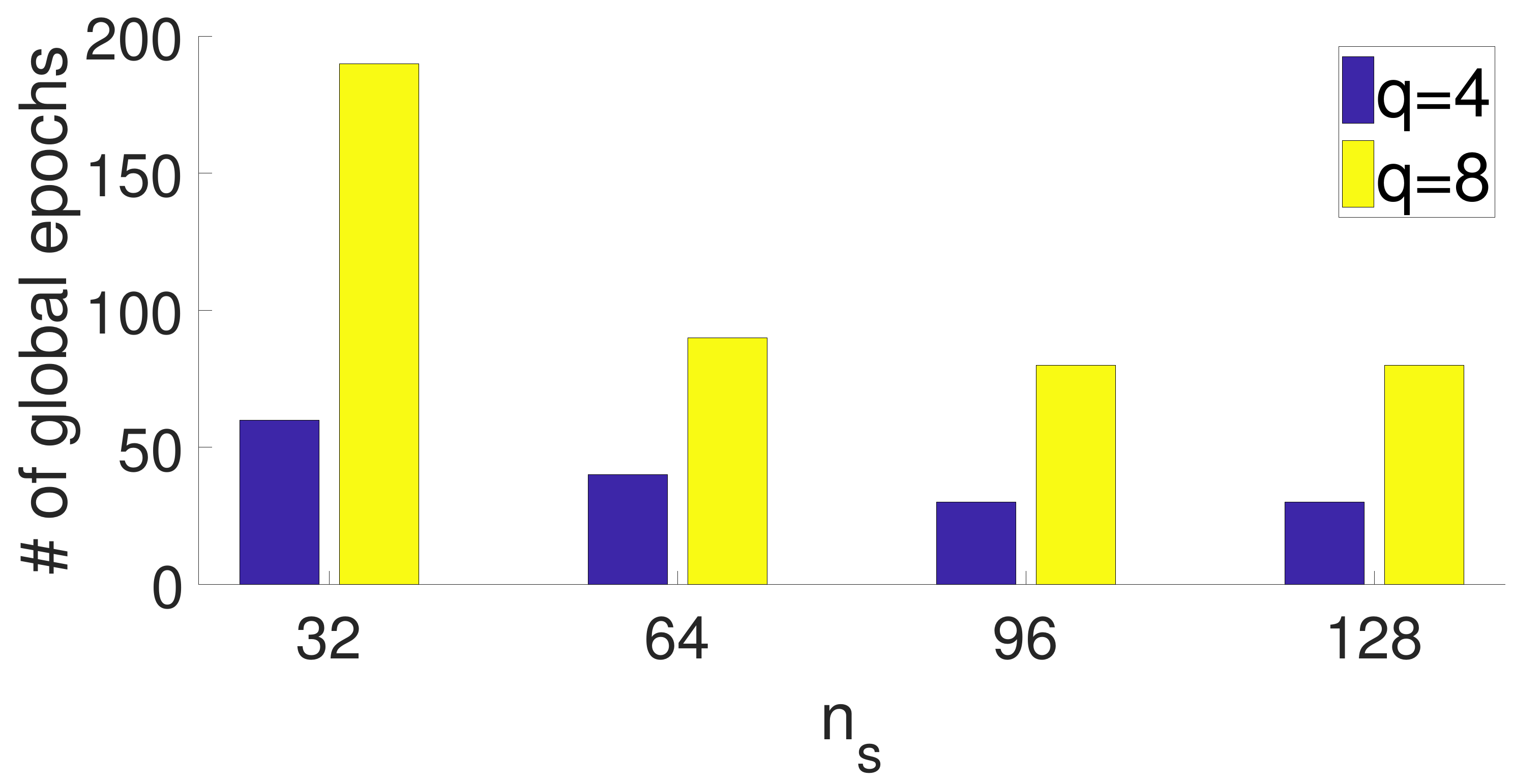}
\caption{Number of global epochs to reach training loss value $0.7$, with sign-flipping attacks and $q \in \{4, 8\}$ Byzantine workers. $\rho=0.001$, $\epsilon=0$, $k=15$ for \texttt{Zeno++}. The batch size of \texttt{Zeno++}, $n_s$, varies.}
\label{fig:zeno_ns}
\end{figure*}

\clearpage

\subsection{WikiText-2}

\begin{figure*}[htb!]
\centering
\subfigure[Perplexity on testing set, $q=0$.]{\includegraphics[width=0.58\textwidth]{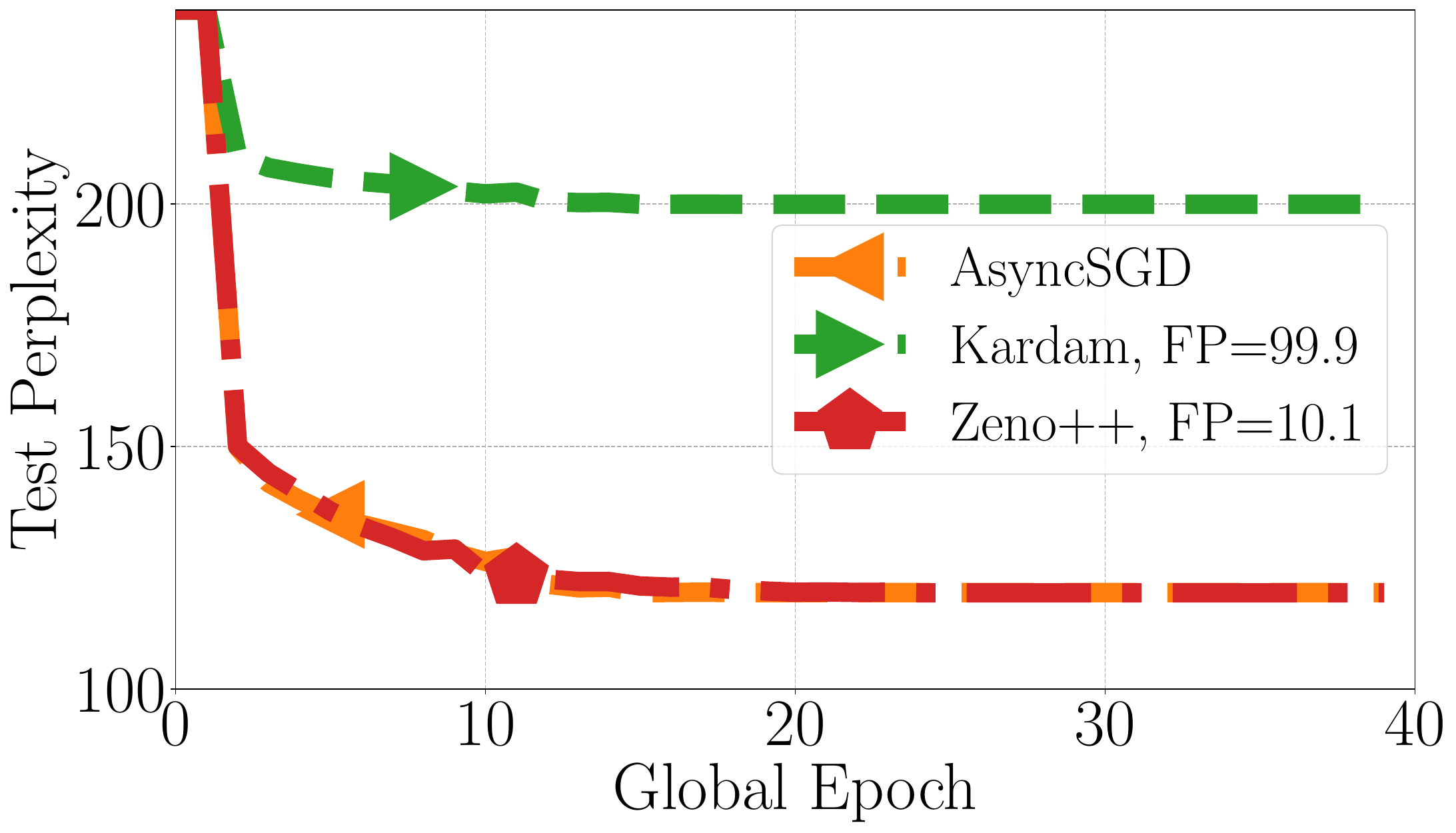}}
\subfigure[Perplexity on testing set, $q=4$.]{\includegraphics[width=0.58\textwidth]{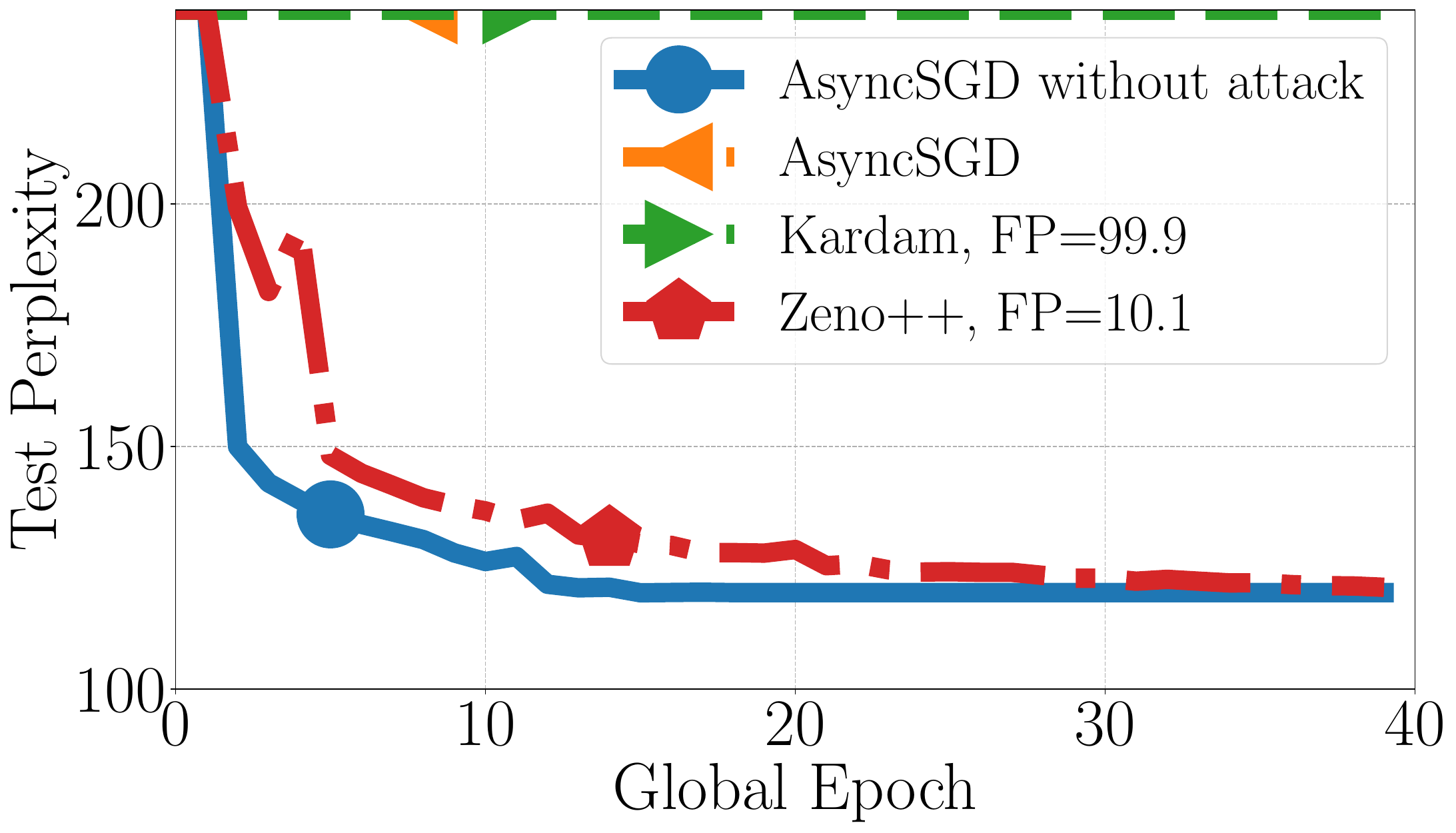}}
\subfigure[Perplexity on testing set, $q=6$.]{\includegraphics[width=0.58\textwidth]{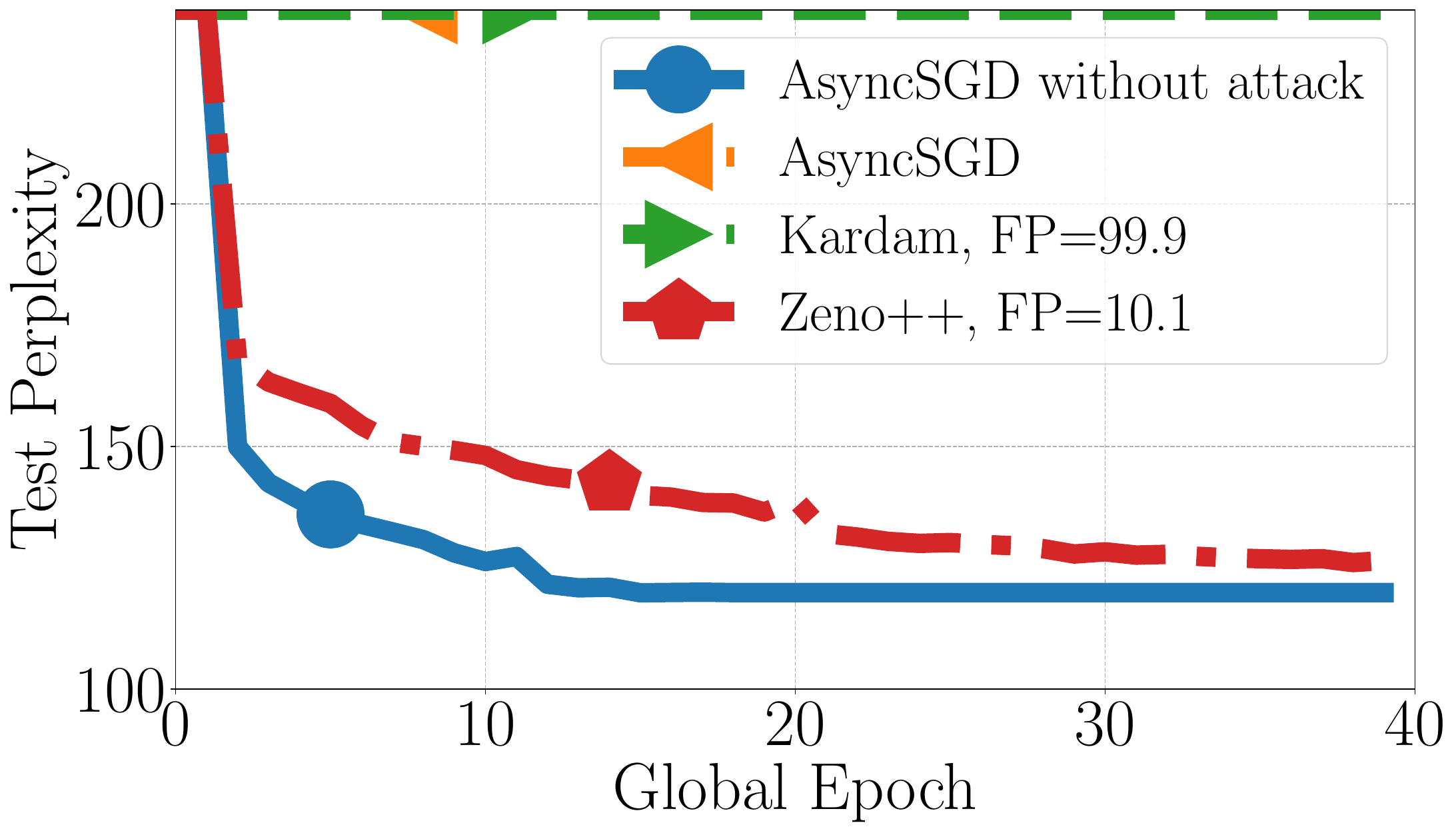}}
\caption{Perplexity~(the lower the better) on LSTM-based language model and WikiText-2 dataset. We show the convergence with sign-flipping attacks with maximum worker delays $k_w = 10$. For any correct gradient $g$, if selected to be Byzantine, $g$ will be replaced by $-10g$. $q \in \{0,4,6\}$ out of the 10 workers are Byzantine. $\gamma=20$ for all the algorithms. $\rho=10, \epsilon=2.0, k=10$ for \texttt{Zeno++}. {\em FP} refers to the fraction of false positive detect ions i.e. incorrect prediction that a message is Byzantine. Note that when $q=4$ or $6$, the results of Kardam are off the charts.}
\label{fig:signflip_wikitext}
\end{figure*}

\end{document}